\documentclass[mnsc,nonblindrev]{informs3_noheading}

\OneAndAHalfSpacedXI

\usepackage{natbib}
 \bibpunct[, ]{(}{)}{,}{a}{}{,}%
 \def\bibfont{\small}%

\usepackage{amsmath,bbm}
\usepackage[normalem]{ulem}
\usepackage[dvipsnames]{xcolor}
\usepackage[hypertexnames=false,hyperfootnotes=false]{hyperref}
\usepackage[capitalise]{cleveref}
\usepackage{enumerate}
\usepackage{xparse}
\usepackage{booktabs}
\usepackage{multirow}
\usepackage{subcaption} 
\usepackage{relsize}  

\TheoremsNumberedThrough     

\EquationsNumberedThrough    

\MANUSCRIPTNO{} 

\newtheorem{fact}[theorem]{Fact}

\newcommand{\bI}{\mathbf{1}}

\newcommand{\bR}{\mathbb{R}}
\newcommand{\bE}{\mathbb{E}}

\newcommand{\pmin}{p_{\min}}

\newcommand{\fexplore}{\textsf{fexplore}}
\newcommand{\texplore}{\textsf{texplore}}
\newcommand{\bexploit}{\textsf{bexploit}}

\newcommand{\Nf}{\textsf{Nf}}
\newcommand{\Nt}{\textsf{Nt}}
\newcommand{\Nb}{\textsf{Nb}}

\newcommand{\cI}{\mathcal{I}}
\newcommand{\cC}{\mathcal{C}}

\newcommand{\OPT}{\mathsf{OPT}}

\newcommand{\ssf}{f^*}

\newcommand{\ssA}{A^*}

\newcommand{\zerovec}{\mathbf{0}}

\newcommand{\sumt}{\sum_{t=1}^T}
\newcommand{\sumk}{\sum_{k=1}^K}

\newcommand{\ii}{\textit}

\newcommand{\cA}{\mathcal{A}}
\newcommand{\cAg}{\mathcal{A}^g}
\newcommand{\eps}{\varepsilon}

\newcommand{\cD}{\mathcal{D}}
\newcommand{\cB}{\mathcal{B}}
\newcommand{\cT}{\mathcal{T}}

\newcommand{\gmin}{\Gamma}

\newcommand{\hq}{\hat{q}}

\newcommand{\htheta}{\hat{\theta}}
\newcommand{\hmu}{\hat{\theta}}
\newcommand{\hmut}{\hat{\theta}_t}

\newcommand{\hthetat}{\hat{\theta}_t}
\newcommand{\hthetan}{\hat{\theta}^n}

\newcommand{\vgg}{\; \bigg| \;}
\newcommand{\vg}{\; \big| \;}
\newcommand{\vv}{\;| \;}

\newcommand{\KL}{\text{KL}}
\newcommand{\thetaA}{\theta^A}

\newcommand{\thetaB}{\theta^B}

\newcommand{\Prgamma}{\Pr\nolimits_{\gamma}}
\newcommand{\Prtheta}{\Pr\nolimits_{\theta}}

\newcommand{\NFUCB}{\text{PF-UCB}}
\newcommand{\PFUCB}{\text{PF-UCB}}
\newcommand{\KLUCB}{\text{KL-UCB}}
\newcommand{\pof}{\mathrm{PoF}}

\newcommand{\Reg}{\mathrm{Regret}}
\newcommand{\IReg}{\mathrm{IReg}}

\newcommand{\cG}{\mathcal{G}}

\newcommand{\Ag}{A^{\text{greedy}}}
\newcommand{\AU}{A^{\text{UCB}}}
\newcommand{\ER}{R} 

\newcommand{\Ngroupa}{J^g(a)}
\newcommand{\Nopta}{J(a)}
\newcommand{\hNgroupa}{\hat{J}^g(a)}
\newcommand{\hNopta}{\hat{J}(a)}
\newcommand{\hDelta}{\hat{\Delta}}
\newcommand{\tDelta}{\tilde{\Delta}}

\newcommand{\conspolicies}{\Psi}

\newcommand{\Asub}{\mathcal{A}_{\mathrm{sub}}}
\newcommand{\Asubg}{\mathcal{A}_{\mathrm{sub}}^g}
\newcommand{\hAsub}{\hat{\cA}_{\mathrm{sub}}}
\newcommand{\Aoptg}{\mathcal{A}_{\mathrm{opt}}^g}
\newcommand{\AKL}{\cA^{\mathrm{UCB}}_t}
\newcommand{\arr}{\mathrm{Arr}^g_t}

\newcommand{\fairopt}{\bar{L}}

\newcommand{\pulls}{\mathrm{Pull}_t}
\newcommand{\pull}{\mathrm{Pull}}
\newcommand{\gpulls}{\pulls^g}

\newcommand{\ssY}{Y^*}
\newcommand{\var}{\text{Var}}
\newcommand{\UCB}{\mathrm{UCB}}

\newcommand{\tR}{\tilde{R}}

\newcommand{\limsupT}{\limsup_{T\rightarrow \infty}}
\newcommand{\liminfT}{\liminf_{T\rightarrow \infty}}
\newcommand{\limT}{\lim_{T\rightarrow \infty}}
\newcommand{\limn}{\lim_{n\rightarrow \infty}}
\newcommand{\allucb}{\Lambda}
\newcommand{\notallucb}{\bar{\Lambda}}

\newcommand{\ssu}{u^*}

\newcommand{\hs}{\hat{s}}
\newcommand{\hf}{\hat{f}}
\newcommand{\fbd}{f_{\mathrm{bd}}}
\newcommand{\hA}{\hat{A}}
\newcommand{\hQ}{\hat{Q}}
\newcommand{\tQ}{\tilde{Q}}
\newcommand{\hC}{\hat{C}}
\newcommand{\cP}{\mathcal{P}}
\newcommand{\hb}{\hat{b}}
\newcommand{\tq}{\tilde{q}}
\newcommand{\tA}{\tilde{A}}
\newcommand{\tb}{\tilde{b}}
\newcommand{\AcI}{A_{\cI}}
\newcommand{\hAcI}{\hA_{\cI}}

\newcommand{\bcI}{b_{\cI}}
\newcommand{\proj}{\mathrm{proj}}

\newcommand{\cM}{\mathcal{M}}
\newcommand{\tcM}{\tilde{\mathcal{M}}}
\newcommand{\cQ}{\mathcal{Q}}

\newcommand{\welfare}{SW}
\newcommand{\ugain}{\mathrm{UtilGain}}

\usepackage{ulem}
\usepackage{xcolor}

\newcommand{\edit}[1]{{\color{black} #1}}

\DeclareMathAlphabet\mathbfcal{OMS}{cmsy}{b}{n}
\NewDocumentEnvironment{myproof}{o}
  {\IfNoValueTF{#1}{\paragraph{{Proof.} }} {\paragraph{{#1.} }} }
  {\hfill$\Halmos$}

\begin{document}

\RUNAUTHOR{Baek and Farias}

\RUNTITLE{Fair Exploration via Axiomatic Bargaining}

\TITLE{Fair Exploration via Axiomatic Bargaining}

\ARTICLEAUTHORS{%
\AUTHOR{Jackie Baek}
\AFF{Operations Research Center, Massachusetts Institute of Technology, \EMAIL{baek@mit.edu}}
\AUTHOR{Vivek F. Farias}
\AFF{Sloan School of Management, Massachusetts Institute of Technology, \EMAIL{vivekf@mit.edu}}
} 

\ABSTRACT{%

Exploration is often necessary in online learning to maximize long-term reward, but it comes at the cost of short-term `regret'. We study how this cost of exploration is shared across multiple groups. For example, in a clinical trial setting, patients who are assigned a sub-optimal treatment effectively incur the cost of exploration. When patients are associated with natural groups on the basis of, say, race or age, it is natural to ask whether the cost of exploration borne by any single group is `fair'.  So motivated, we introduce the `grouped' bandit model. We leverage the theory of axiomatic bargaining, and the Nash bargaining solution in particular, to formalize what might constitute a fair division of the cost of exploration across groups. On the one hand, we show that any regret-optimal policy strikingly results in the least fair outcome: such policies will perversely leverage the most `disadvantaged' groups when they can. More constructively, we derive policies that are optimally fair and simultaneously enjoy a small `price of fairness'. We illustrate the relative merits of our algorithmic framework with a case study on contextual bandits for warfarin dosing where we are concerned with the cost of exploration across multiple races and age groups. 

}%


\KEYWORDS{bandits; fairness; exploration; Nash bargaining solution; proportional fairness}

\maketitle

\section{Introduction}

Exploration is the act of taking actions whose rewards are highly uncertain in the hopes of discovering one with a large reward.
It is well-known that exploration is a key, and often necessary component in online learning problems. Exploration has an implicit cost, insomuch that exploring actions that are eventually revealed to be sub-optimal incurs `regret'. 
This paper studies how this cost of exploration is shared in a system with multiple stakeholders. At the outset, we present two practical examples that motivate our study of this issue. 

\textbf{Personalized Medicine and Adaptive Trials: } 
Multi-stage, adaptive designs \citep{kim2011battle,berry2012adaptive,berry2015brave,rugo2016adaptive}, are widely viewed as the frontier of clinical trial design. More generally, the ability to collect detailed patient level data, combined with real time monitoring (such as glucose monitoring for diabetes \citep{bergenstal2019automated,nimri2020insulin}) has raised the specter of learning personalized treatments. As a concrete example, consider the problem of finding the optimal dosage of warfarin, a blood thinner that is commonly used to treat blood clots.
The optimal dosage varies widely between patients (up to a factor of 10), and an incorrect dose can have severe adverse effects \citep{wysowski2007bleeding}.
Learning the appropriate personalized dosage is then naturally viewed as a contextual bandit problem \citep{bastani2020online} where the context at each time step corresponds to a patient's covariates, arms correspond to different dosages, and the reward is the observed efficacy of the assigned dose. In examining such a study in retrospect, it is natural to measure the `regret' incurred by distinct groups of patients (say, their race), measured by comparing the overall treatment efficacy for that group and the optimal treatment efficacy that could have been achieved for that group in hindsight. This quantifies the cost of exploration borne by that group. Now since covariates that impact dose efficacy may be highly correlated with race (e.g. genomic features), it is a priori unclear how a generic learning policy would distribute the cost of exploration across groups. More generally, we must have a way of understanding whether a given profile of exploration costs across groups is, in an appropriate sense, `fair'. 

\textbf{Revenue Management for Search Advertising: }
Ad platforms enjoy a tremendous amount of flexibility in the the choice of ads served against search queries. Specifically, this flexibility exists both in selecting a slate of advertisers to compete for a specific search, and then in picking a winner from this slate. Now a key goal for the platform is learning the affinity of any given ad for a given search. In solving such a learning problem -- for which many variants have been proposed \citep{graepel2010web,agarwal2014laser} -- we may again ask the question of who bears the cost of exploration, and whether the profile of such costs across various groups of advertisers is fair.  

\subsection{Bandits, Groups and Axiomatic Bargaining}

Delaying a formal development to later, a generic bandit problem is described by an uncertain parameter that must be learned, a set of feasible actions, and a reward function that depends on the unknown parameter and the action chosen. The set of actions available to the learner may change at each time, and the learner must choose which action to pick over time in a manner that minimizes `regret' relative to a strategy that, in each time step, picked an optimal action with knowledge of the unknown parameter. To this model, we add the notion of a `group'; in the warfarin example, a group might correspond to a specific race. 
Each group is associated with an arrival probability and a distribution over action sets. At each time step, a group and an action set is drawn from this distribution.
We refer to this problem as a `grouped' bandit. 
Now in addition to measuring overall regret in the above problem, we also care about the regret incurred by specific groups, which we can view as the cost of exploration borne by that group. 
As such, any notion of fairness is naturally a function of the profile of regret incurred across groups. 

In reasoning about what constitutes a `fair' regret profile in a principled fashion, we turn to the theory of axiomatic bargaining. There, a central decision maker is concerned with the incremental utility earned by each group from collaborating, relative to the utility the group would have earned on its own. In our bandit setting this incremental utility is precisely the reduction in regret for any given group relative to the optimal regret that group would have incurred should it have been `on its own'.
A bargaining solution is simply a solution to maximizing some (axiomatically justified) objective function over the set of achievable incremental utilities. The {\em utilitarian solution}, for instance, simply maximizes the sum of incremental utilities. Applied to the bandit problem, the utilitarian solution would simply minimize total regret, in effect yielding the usual regret optimal solution and ignoring the relevance of groups. 
The {\em Nash bargaining solution} maximizes an alternative objective, the Nash Social Welfare (SW) function. This latter solution is the unique solution to satisfy a set of axioms any `fair' solution would reasonably satisfy. { \em This paper develops the Nash bargaining solution to the (grouped) bandit problem.}

\subsection{Contributions}

In developing the Nash bargaining solution in the context of bandits, we focus on the `grouped' variant of two classical bandit models in the literature: $K$-armed bandits and linear contextual bandits.
We make the following contributions relative to this problem: 
\begin{itemize}
\item {\em Regret-Optimal Policies are Unfair:} We show that all regret-optimal policies 
share a structural property that make them `arbitrarily unfair' -- in the sense that the Nash SW is $-\infty$ for these solutions -- under a broad set of conditions on the problem instance.
We show that a canonical UCB policy is regret-optimal, hence UCB exhibits this unfairness property.
\item {\em Achievable Fairness:} We derive an instance-dependent upper bound on the Nash SW.
This can be viewed as a `fair' analogue to a regret lower bound (e.g. \cite{lai1985asymptotically}) for the problem, since a lower bound on achievable regret (forgoing any fairness concerns) would in effect correspond to an upper bound on the utilitarian SW for the problem.
\item {\em Nash Solution:} 
We develop policies that achieve the Nash solution.
Specifically, we introduce the policies PF-UCB for grouped $K$-armed bandits and PF-OAM for grouped linear contextual bandits, and we prove that each of these policies achieve the upper bound on the Nash SW for all instances.
\item {\em Price of Fairness for the Nash Solution:} We show that the `price of fairness' for the Nash solution is small: if $G$ is the number of groups, the Nash solution achieves at least $O(1/\sqrt{G})$ of the reduction in regret achieved under a regret optimal solution relative to the regret incurred when groups operate separately.
\item {\em Warfarin Dosing Case Study:} Applying our framework to a real-world dataset on warfarin dosing using race and age groups, we show that compared to a regret-optimal policy, the Nash solution is able to balance out reductions in regret across groups at the cost of a small increase in total regret.
\end{itemize}

\subsection{Simple Instance}
Before diving into the full model, we describe the simplest non-trivial instance of the grouped bandit model that captures the main idea of the problem we study.

\begin{example}[2-group, 3-arm bandit] \label{ex:3arm}
Suppose there are two groups A and B, and there are three arms with mean rewards satisfying $\theta_1 < \theta_2 < \theta_3$.
Suppose group A has access to arms 1 and 2, while group B has access to arms 1 and 3.
One of the two groups arrive at each time step, where each group arrives with probability 50\%.
\end{example}

In this example, the shared arm, arm 1, is suboptimal for both groups.
A reasonable bandit policy must pull arm 1 enough times (i.e. explore enough) to distinguish that it is worse than the other two arms.
Suppose, for simplicity, that $\theta_2$ and $\theta_3$ are known a priori.
The regret incurred when group A pulls arm 1, $\theta_2 - \theta_1$, is smaller than the regret when group B pulls arm 1, $\theta_3 - \theta_1$.
Then, it is intuitive that to minimize total regret, it is more efficient for group A to do the exploration and pull arm 1 rather than for group B to do so.
We show that indeed, the policy of exploring only with group A is optimal in terms of minimizing total regret.
However, such a policy results in group A incurring all of the regret, while group B incurs none --- that is, group B `free-rides' off of the learnings earned by group A.
It can be argued this outcome is `unfair'; yet, prior to this work, there was no framework to establish what would be a fairer solution for this instance.
The framework introduced in this paper formalizes what a fair division of exploration between groups A and B should be.

\subsection{Related Literature}
Two pieces of prior work have a motivation similar to our own.
\cite{jung2020quantifying} study a setting with multiple agents with a common bandit problem, where each agent can decide which action to take at each time. 
They show that `free-riding' is possible --- an agent that can access information from other agents can incur only $O(1)$ regret in several classes of problems. 
This result is consistent with the motivation for our work. 
\cite{raghavan2018externalities} study a very similar grouped bandit model to ours, and provides a `counterexample' in which a group can have a negative externality on another group (i.e. the existence of group B increases group A's regret). This example is somewhat pathological and stems from considering an instance-specific fixed time horizon; instead, if $T \rightarrow \infty$, all externalities become non-negative (details in \cref{app:negative_externality}).
Our grouped bandit model is also similar to \ii{sleeping bandits} \citep{kleinberg2010regret}, in which the set of available arms is adversarially chosen in each round. The known, fixed group structure in our model allows us to achieve tighter regret bounds compared to sleeping bandits.

There have also been a handful of papers (e.g. \cite{joseph2016fairness,liu2017calibrated,gillen2018online,patil2020achieving}) that study `fairness in bandits' in a completely different context. These works enforce a fairness criterion between \ii{arms}, which is relevant in settings where a `pull' represents some resource that is allocated to that arm, and these pulls should be distributed between arms in a fair manner.
In these models, the decision maker's objective (maximize reward) is distinct from that of a group (obtain `pulls'), unlike our setting (and motivating examples) where the groups and decision maker are aligned in their eventual objective. 

Our upper bound on Nash SW borrows classic techniques from the regret lower bound results of \cite{lai1985asymptotically} and \cite{graves1997asymptotically}. Our policy follows a similar pattern to recent work on regret-optimal, optimization-based policies for structured bandits (e.g. \cite{lattimore2017end,combes2017minimal,van2020optimal,hao2020adaptive}).
Our policy includes novel algorithmic techniques and does not require forced exploration, 
which we show, through simulations, improves practical performance.

Our fairness framework is inspired by the literature on fairness in welfare economics --- see \cite{young1995equity, sen1997economic}. 
Specifically, we study fairness in exploration through the lens of the axiomatic bargaining framework, first studied by \cite{nash1950bargaining}, who showed that enforcing four desirable axioms induces a unique fair solution. \cite{mas1995microeconomic} is an excellent textbook reference for this topic. 
This fairness notion is often referred to as proportional fairness, which has been studied extensively especially in the area of telecommunications \citep{kelly1998rate,jiang2005proportional}.

Lastly, the burden of exploration in learning problems has been studied in the \ii{decentralized} setting, where each time step represents a self-interested agent that decides on their own which action to take. 
One line of work in this setting aims to identify classes of bandit problems in which the greedy algorithm performs well (e.g. \cite{bastani2020mostly,chen2018incentivizing,kannan2018smoothed,raghavan2018externalities}).
Another stream of literature re-designs how the agents interact with the system to induce exploration, either by modifying the information structure (e.g. \cite{kremer2014implementing,mansour2015bayesian,papanastasiou2018crowdsourcing,immorlica2018incentivizing}),
or providing payments to incentivize exploration (e.g. \cite{frazier2014incentivizing,kannan2017fairness}).
The difference of our work compared to these aforementioned papers is the presence of a centralized decision maker.

\subsection{Roadmap of Paper} \label{sec:roadmap}
We outline the roadmap for the rest of this paper. 

\begin{itemize}
	\item In \cref{sec:model}, we introduce a general `grouped bandit' model, 
	and we establish a fairness framework for this model by defining the Nash SW and the Nash solution.
	We then introduce two examples of grouped bandit models: grouped $K$-armed bandits and grouped linear contextual bandits. 
	The results of Sections~\ref{sec:analyze_klucb} and \ref{sec:pfucb} pertain to the grouped $K$-armed bandit model, while \cref{sec:contextual_linear_bandits} pertains to the grouped linear contextual bandit model.
	\item In \cref{sec:analyze_klucb}, we state and prove the result that regret-optimal policies are unfair (\cref{corr:unfair}),
	and we show an upper bound on Nash SW (\cref{prop:nash_sw_ub}).
	\item In \cref{sec:pfucb}, we introduce the policy PF-UCB, which we prove is the Nash solution for the grouped $K$-armed bandit model (\cref{thm:main_fair_orig}).
	We then provide worst-case upper bounds on the price of fairness (\cref{thm:general_pof}).
	\item In \cref{sec:contextual_linear_bandits}, we study the grouped linear contextual bandit model.
	We show an upper bound on Nash SW (\cref{thm:sw_ub_linear}), and then we introduce a policy called PF-OAM, which we prove to be the Nash solution for grouped contextual bandits (\cref{thm:pfoam_nash_solution}).
	We also bound the price of fairness (\cref{prop:pof_linear}).
	\item \cref{sec:experiments} contains three sets of computational experiments.
	First, we evaluate the price of fairness on random instances to compare it to the theoretical worst-case bounds from \cref{sec:pfucb}.
	Second, we simulate several bandit policies and plot the trajectory of each group's regret over time to understand how well the theoretical asymptotic results translate to a finite-time setting.
	Lastly, we provide a case study on a real-world dataset regarding warfarin dosing.
	\item Finally, in \cref{sec:conclusion}, we discuss our modeling assumptions and possible alternatives that could be studied. 
	We then describe future directions and conclude.
\end{itemize}

\section{The Axiomatic Bargaining Framework for Bandits} \label{sec:model}

We first describe the generic `grouped' bandit model.
We then provide a background of the axiomatic bargaining framework of \cite{nash1950bargaining}, and describe how we apply this framework to the grouped bandit setting. 
Lastly, we describe two examples of the grouped bandit model studied in this work, the grouped $K$-armed bandit and the grouped linear contextual bandit. 

\subsection{Grouped Bandit Model} \label{sec:model:grouped_bandit}

Let $\theta \in \Theta$ be an unknown parameter and let $\cA$ be the action set. For every arm $a \in \cA$,  $(Y_n(a))_{n \geq 1}$ is an i.i.d. sequence of rewards drawn from a distribution $F(\theta, a)$ parameterized by $\theta$ and $a$. We let $\mu(a) = \bE[Y_1(a)]$ be the expected reward of arm $a$. In defining a {\em grouped} bandit problem, we let $\cG$ be a finite set of groups. Each group $g \in \cG$ is associated with a probability distribution $P^g$ over $2^{\cA}$, and a probability of arrival $p_g$; $\sum_g p_g = 1$. The group arriving at time $t$, $g_t$, is chosen independently according to this latter distribution; $\cA_t$ is then drawn according to $P^{g_t}$. An instance of the grouped bandit problem is specified by $\cI = (\cA, \cG,p, P, F, \theta)$, where all quantities except for $\theta$ are known.
At each time $t$, a central decision maker observes $g_t$ and $\cA_t$, chooses an arm $A_t \in \cA_t$ to pull and observes the reward $Y_t = Y_{N_t(A_t)+1}(A_t)$, 
where $N_t(a)$ is the total number of times arm $a$ was pulled up to but not including time $t$.
Let $\ssA_t \in \argmax_{a \in \cA_t} \mu(a)$ be an optimal arm at time $t$.
Given an instance $\cI$ and a policy $\pi$, the \ii{total regret}, and the \ii{group regret} for group $g \in \cG$ are respectively 
\begin{align*}
\ER_T(\pi, \cI) =\bE\left[  \sum_{t=1}^T  (\mu(\ssA_t) - \mu(A_t))\right] \ {\rm and \ }
\ER^g_T(\pi, \cI) =\bE\left[  \sum_{t=1}^T \bI(g_t = g) (\mu(\ssA_t) - \mu(A_t)) \right],
\end{align*}
where the expectation is over randomness in arrivals $(g_t, \cA_t)$, rewards $Y_n(a)$, and the policy $\pi$. 

Finally, so that the notion of an optimal policy for some class of instances, $\mathbfcal{I}$,  is well defined, we restrict our attention to \ii{consistent} policies which yield sub-polynomial regret for any instance in that class: $\conspolicies = \{\pi: \ER_T(\pi, \cI) = o(T^b) \ \forall \cI \in \mathbfcal{I}, \forall b > 0 \}$. 
We say that a policy $\pi^* \in \conspolicies$ is \textit{regret-optimal} for $\mathbfcal{I}$ if for any instance $\cI \in \mathbfcal{I}$ and any policy $\pi \in \conspolicies$,
\begin{align} \label{eq:regret_optimal}
\limsupT \frac{\ER_T(\pi^*, \cI)}{\log T} \leq \liminfT \frac{\ER_T(\pi, \cI)}{\log T}.
\end{align}

\subsection{Background: Axiomatic Bargaining} \label{sec:model:background}

We now describe the axiomatic bargaining framework, which we will apply to the grouped bandit model.
The axiomatic bargaining problem is specified by the number of agents $n$, a set of feasible utility profiles $U \subseteq \bR^n$, and a disagreement point $d \in \bR^n$, that represents the utility profile when agents cannot come to an agreement. 
A solution $f(\cdot, \cdot)$ to the bargaining problem selects an agreement $\ssu = f(U, d) \in U$, in which agent $i$ receives utility $\ssu_i$.
It is assumed that there is at least one point $u \in U$ such that $u > d$, and we assume $U$ is compact and convex. 

The bargaining framework proposes a set of axioms a fair solution $u^*$ should ideally satisfy:
\begin{enumerate}
	\item \ii{Pareto optimality}: There does not exist a feasible solution $u \in U$ such that $u \geq \ssu$ and $u \neq \ssu$.
	\item \ii{Symmetry}: 
	 If all entries of $d$ are equal and the space $U$ is symmetric (if $u, u'$ only differ in the permutation of its entries, then $u \in U$ if and only if $u' \in U$), then all entries of $\ssu$ are equal.
	\item \ii{Invariant to affine transformations}: Let $a \in \bR_+^n$, $b \in \bR^n$. If $U' = \{ a \circ u + b : u \in U\}$ and $d' = a \circ d + b$, then $f(U', d')_i = a_i \ssu_i + b_i$ (where $\circ$ refers to an element-wise product).
	\item \ii{Independence of irrelevant alternatives}: If $V \subseteq U$ where $\ssu \in V$, then $f(V, d) = \ssu$.
\end{enumerate}
Now invariant to affine transformations implies that $f(U,d) =  f(\{u-d:u \in U\},0) + d$. It is therefore customary to normalize the origin to the disagreement point, i.e. assume $d=0$, and implicitly assume that $U$ has been appropriately translated. So translated, $U$ is interpreted as a set of feasible utility {\em gains} relative to the disagreement point. The seminal work of \cite{nash1950bargaining} showed that there is a unique bargaining solution that satisfies the above four axioms, and it is the outcome that maximizes the \ii{Nash social welfare (SW) function} \citep{kaneko1979nash}: 
\begin{align*}
 \welfare(u) = \sum_{i=1}^n \log(u_i),
\end{align*}
where $\log a \triangleq -\infty$ for $a \leq 0$.
The unique solution $\ssu = \argmax_{u \in U} \welfare(u)$ is often referred to as the \ii{proportionally fair} solution, as it applies the following standard of comparison between feasible outcomes:
a transfer of utilities between two agents is favorable if the percentage increase in the utility gain of one agent is larger than the percentage decrease in utility gain of the other agent.
Mathematically, $\ssu$ satisfies:
\begin{align*}
\sum_{i=1}^n \frac{u_i - \ssu_i}{\ssu_i } \leq 0 \quad \forall u \in U, u \geq 0.
\end{align*}
That is, from $\ssu$, the aggregate proportional change in utility gains to any other feasible solution is non-positive.
We will interchangeably refer to $\ssu = \argmax_{u \in U} \welfare(u)$ as the \ii{Nash solution} or as \ii{proportionally fair}. If $u \in U$ such that $\welfare(u) = -\infty$, we say that $u$ is \ii{unfair}.
An unfair outcome corresponds to an outcome where there exists an agent who receives a non-positive utility gain.

\subsection{Fairness Framework for Grouped Bandits} \label{sec:fairness_framework}
We now consider the Nash bargaining solution in the context of the grouped bandit problem. To do so, we need to appropriately define the utility gain under any policy. 
\edit{
We begin by formalizing rewards at the disagreement point, which, informally, represents the best that a group can do if no information was shared across groups.
Given an instance $\cI \in \mathbfcal{I}$, define $\cI_g$ to be the `single-group' bandit instance where for any period $t$ in which $g_t \neq g$, we receive no reward under any action.
Formally, $\cI_g$ has the same set of groups $\cG$ and has the action set $\cA \cap \{a_0\}$, where $a_0$ is a null action with $\mu(a_0) = 0$.
Then, for any group $g' \neq g$, $P^{g'}$ is deterministically $\{a_0\}$ --- that is, only the null action is available when $g_t \neq g$.} 
Let us denote by $\pi^*_g$ a regret-optimal policy for instances of type $\cI_g$ so that for any instance of type $\cI_g$, and any other consistent policy $\pi'_g$ for instances of that type,
\begin{align} \label{eq:single_group_regret_optimal}
\limsupT \frac{\ER_T(\pi^*_g, \cI_g)}{\log T} \leq \liminfT \frac{\ER_T(\pi'_g, \cI_g)}{\log T}.
\end{align}
Letting $\tR^g_T(\cI) \triangleq \ER_T(\pi^*_g, \cI_g)$, we define, with a slight abuse of notation, the $T$-period utility earned by group $g$ under $\pi^*_g$, and any other consistent policy $\pi$ for instances of type $\cI$ respectively, as: 
\[
\bE\left[\sumt \bI(g_t =g) \mu(\ssA_t)\right]- \tR^g_T(\cI) \triangleq u_T^g(\pi^*_g) {\rm \ and \ }
\bE\left[\sumt \bI(g_t =g) \mu(\ssA_t)\right]- \ER^g_T(\pi, \cI) \triangleq u_T^g(\pi). 
\]
The $T$-period utility gain under a policy $\pi$ is then $u_T^g(\pi) - u_T^g(\pi^*_g) = \tR^g_T(\cI) - R^g_T(\pi, \cI)$. 
\edit{Note that the quantities $u_T^g(\pi^*_g)$ and $u_T^g(\pi)$ are comparable in that the number of arrivals of group $g$ are equal under both quantities.
}
Since our goal is to understand long-run system behavior, we define asymptotic utility gain for any group $g$: 
\begin{align} \label{eq:def_ugain}
\ugain^g(\pi, \cI) = \liminfT  \frac{\tR^g_T(\cI) - \ER^g_T(\pi, \cI)}{\log T}.
\end{align}
In words, $\ugain^g(\pi, \cI)$ is the decrease in regret for group $g$ under policy $\pi$, compared to the best that group $g$ could have done on its own (which is to run $\pi^*_g$ on instance $\cI_g$).
Equipped with this definition, we may now identify the set of incremental utilities for an instance $\cI$, as $U(\cI) = \{(\ugain^g(\pi, \cI))_{g \in \cG} :  \pi \in \conspolicies \}$. 
It is worth noting that while $U(\cI)$ is not necessarily convex, we can readily show that  the Nash solution remains the unique solution satisfying the fairness axioms presented in Section~\ref{sec:model:background} relative to $U(\cI)$ (details in Appendix~\ref{app:nash_sol_unique}). 

We now define the Nash SW function.
Since we find it convenient to associate a SW function with a policy (as opposed to a vector of incremental utilities), the Nash SW function for grouped bandits is then defined as:
\begin{align}  \label{eq:nash_welfare_function}
\welfare(\pi, \cI) =
 \sum_{g \in G(\cI)} \log \left( \ugain^g(\pi, \cI) \right),
\end{align}
where $\log u \triangleq -\infty$ for $u \leq 0$.
We finish by defining the Nash solution to the grouped bandit problem.
\begin{definition}\label{def:pf}
Suppose a policy $\pi^*$ satisfies $\welfare(\pi^*, \cI) = \sup_{\pi \in \conspolicies} \welfare(\pi, \cI)$ for every instance $\cI \in \mathbfcal{I}$. 
Then, we say that $\pi^*$ is the \ii{Nash solution} for $\mathbfcal{I}$ and that it is \ii{proportionally fair}.
\end{definition}

\subsection{Examples of Grouped Bandit Models} \label{sec:examples_grouped_bandit_models}
We now describe two examples of the grouped bandit model which impose additional assumptions on the rewards and group structure. 

\subsubsection{Grouped $K$-armed Bandit Model.}  \label{sec:karmed_bernoulli_bandit}
Let $\cA = [K]$. 
There is a fixed bipartite graph between the groups $\cG$ and the arms $\cA$; denote by $\cA^g \subseteq \cA$ the arms connected to group $g$ and by $\cG_a \subseteq \cG$ the groups connected to arm $a$. 
For each $g$, $P^g$ places unit mass on $\cA^g$ so that the set of arms available at time $t$ is $\cA_t = \cA^{g_t}$.
Assume $\theta \in (0, 1)^K$, and the single period reward $Y_1(a) \sim \mathrm{Normal}(\theta(a), 1)$.
We assume that $\theta(a) \neq \theta(a')$ for all $a \neq a'$. 
Since the set of arms available at each time step only depends on the arriving group, we denote by $\OPT(g) = \max_{a \in \cA^g} \theta(a)$ the optimal mean reward for group $g$.
We can write the $T$-period regret as 
\begin{align} \label{eq:regret_grouped}
\ER_T(\pi, \cI) = \sum_{g \in G} \sum_{a \in \cA^g} \Delta^g(a) \bE[N^g_T(a)],
\end{align}
where $N^g_T(a)$ is the number of times that group $g$ has pulled arm $a$ after $T$ time steps, and $\Delta^g(a) = \OPT(g) - \theta(a)$.

To define the disagreement point, we show that the KL-UCB policy of \cite{garivier2011kl} is a valid choice of $\pi^*$.
KL-UCB chooses the arm with the highest UCB at each time step, where the UCB is defined as 
\begin{align} \label{eq:define_UCB}
\UCB_t(a) = \max\{ q : N_t(a) \KL(\hmut(a), q) \leq \log t+ 3 \log \log t \},
\end{align}
where $\hmut(a)$ is the empirical mean of arm $a$ at time $t$, and $\KL(\alpha, \beta) = (\alpha-\beta)^2/2$ is the Kullback–Leibler divergence between two normal distributions with means $\alpha$ and $\beta$ and unit variance.
\edit{The following proposition shows that KL-UCB is optimal for single-group instances, and we characterize the regret at the disagreement point. The proof can be found in Appendix~\ref{sec:pf_klucb_disagreement}.

\begin{proposition} \label{prop:klucb_disagreement_pt}
Taking $\pi_g^*$ to be the KL-UCB policy satisfies \eqref{eq:single_group_regret_optimal}.
Moreover, 
\begin{align} \label{eq:disagreement_pt_regret}
\lim_{T \rightarrow \infty}
\frac{\tR^g_T(\cI)}{\log T} 
= \sum_{a \in \cA^g} \Delta^g(a) \Ngroupa,
\end{align}
where $\Ngroupa \triangleq 1/\KL(\theta(a), \OPT(g))$.
\end{proposition}
}

\subsubsection{Grouped Linear Contextual Bandit Model.} \label{sec:grouped_linear_contextual_bandit}
In this model, there is a context associated with each arrival which determines the set of available arms, and each group is associated with a \textit{distribution} over contexts.
Let $\theta \in \bR^d$ and $\cA \subseteq \bR^d$.
The reward from arm $a$ is distributed as $Y_1(a) \sim \mathrm{Normal}(\langle a, \theta \rangle , 1)$.
Let $\cM \subseteq \bR^d$ be the set of contexts, where $|\cM| = M < \infty$, and each $m \in \cM$ is associated with an action set $\cA(m) \subseteq \cA$.
Assume that $\cA(m)$ spans $\bR^d$ for every $m \in \cM$.
Each group $g \in \cG$ has a probability of arrival, $p^g$, and a distribution $P^g$ over contexts $\cM$.
At each time $t$, a group $g_t$ is drawn independently from $(p^g)_g$, then a random context $m_t \sim P^{g_t}$ is drawn.
The action set at time $t$ is $\cA_t = \cA(m_t)$.
Let $\cM^g$ be the contexts in the support of $P^g$.
Let $\OPT(m) = \max_{a \in \cA(m)} \langle a, \theta \rangle$ and $\Delta(m, a) = \OPT(m) - \langle a, \theta \rangle$.
For each context $m \in \cM$, we assume that the optimal arm, $\argmax_{a \in \cA(m)} \langle a, \theta \rangle$, is unique.

\cite{hao2020adaptive} provide a policy called optimal allocation matching (OAM) for the (non-grouped) linear contextual bandit model that satisfies \eqref{eq:single_group_regret_optimal}, and therefore we can take $\pi_g^*$ to be OAM in this setting.

\begin{remark}[Comparison to other linear contextual bandit formulations] \label{remark:formulation}
There are other formulations of the linear contextual bandit model in the literature
(e.g., \cite{goldenshluger2013linear,bastani2020online}), which is mathematically equivalent to the one we study.
The other formulation is one where a context $X_t \in \bR^d$ arrives at time $t$, and the expected reward from pulling an arm $a \in \cA$ is $\langle X_t, \beta_a \rangle$ for unknown parameters $(\beta_a)_{a \in \cA}$.
To convert such an instance to our model, we can let $\theta = (\beta_a)_{a \in \cA}$, and let the context $X_t$ be represented by the action set, where $\cA_t = \{(X_t, 0, \dots, 0), (0, X_t, \dots, 0), \dots, (0, 0, \dots, X_t)\}$.
\end{remark}

\section{Fairness-Regret Trade-off} \label{sec:analyze_klucb}

In this section, we prove that regret-optimal policies must necessarily be unfair for a wide class instances.
We state and prove the main result,
and then we turn to deriving an upper bound on achievable Nash SW.

\subsection{Unfairness of Regret Optimal Policies}\label{sec:unfairness_regret_optimal}

We first state the main result, which states that policies that minimize regret are arbitrarily unfair. 
In fact, we show that perversely the most `disadvantaged' group (in a sense we make precise shortly) bears the brunt of exploration in that it sees no utility gain relative to if it were on its own. 

\begin{theorem} \label{corr:unfair}
Let $\pi$ be a regret-optimal policy for grouped $K$-armed bandits. 
Let $\cI$ be an instance where $g_{\rm min} \triangleq \argmin_{g \in G} \OPT(g)$ is unique. 
Then, $\welfare(\pi, \cI) = -\infty$ and $\ugain^{g_{\rm min}}(\pi, \cI) = 0$. 
\end{theorem}

To prove this result, we first prove a regret lower bound for grouped $K$-armed bandits, along with a matching upper bound by a UCB policy.
Through this, we characterize a common property that all regret-optimal policies share that implies the desired result, $\ugain^{g_{\rm min}}(\pi, \cI) = 0$.

\begin{myproof}[Proof of \cref{corr:unfair}]
Let $\pi \in \conspolicies$ be any consistent policy.
We first lower bound the total number of pulls, $\bE\left[N_{T}(a)\right]$, of a suboptimal arm.
Denote by $\Asubg = \{a \in \cA^g : \theta(a) < \OPT(g)\}$ the suboptimal actions for group $g$, and denote by $\Asub = \{a \in \cA :  a \in \Asubg\;  \forall  g \in \cG_a \}$ the set of arms that are not optimal for any group.
Now since a consistent policy for the grouped $K$-armed bandit is automatically consistent for the vanilla $K$-armed bandit obtained by restricting to any of its component groups $g$, the standard lower bound of \cite{lai1985asymptotically} implies 
that for any $a \in \Asubg$, $\liminf_{T \rightarrow \infty} {\bE\left[N_{T}(a)\right]}/{\log T} \geq \Ngroupa$ where $\Ngroupa \triangleq {1}/{\KL(\theta(a), \OPT(g))}$ (this is formalized in \cref{prop:lb_group_num_pulls} in Appendix~\ref{sec:lower_bound_props}).
Since this must hold for all groups, we have that
\begin{align} 
\label{eq:lower_bound_pulls}
\liminf_{T \rightarrow \infty} \frac{\bE\left[N_{T}(a)\right]}{\log T}  \geq \Nopta
\end{align}
for all $a \in \Asub$ where $\Nopta = \max_{g \in \cG_a} \Ngroupa$. 
Now, denote by $\gmin(a) = \argmin_{g \in \cG_a}\OPT(g)$ the set of groups that have the smallest optimal reward out of all groups that have access to $a$. Then the smallest regret incurred in pulling arm $a$ is $\Delta^g(a)$ for any $g \in \gmin(a)$. With a slight abuse, we denote this quantity by $\Delta^{\gmin(a)}(a)$. 
Then, the lower bound \eqref{eq:lower_bound_pulls} immediately implies the following lower bound on total regret:
\begin{align} 
\label{eq:lower_bound_regret}
\liminf_{T \rightarrow \infty} \frac{\ER_T(\pi, \cI)}{\log T}  \geq \sum_{a \in \cA_{\text{sub}}} \Delta^{\gmin(a)}(a) \Nopta.
\end{align}
In fact, we show that the KL-UCB policy (surprisingly) achieves this lower bound; the intuition for this result is discussed in \cref{remark:klucb_optimal}.
Consequently, any regret-optimal policy must achieve the limit infimum in \eqref{eq:lower_bound_regret}. In turn, this implies that a policy $\pi \in \conspolicies$ is regret-optimal if and only if, the number of pulls of arms $a \in \Asub$ achieve the lower bound \eqref{eq:lower_bound_pulls}, i.e.,
\begin{align} 
\label{eq:property1opt}
\lim_{T \rightarrow \infty} \frac{\bE\left[N_{T}(a)\right]}{\log T}  = \Nopta \quad \forall a \in \Asub
\end{align}
and further that any pulls of arm $a$ from a group $g \notin  \gmin(a) $ must be negligible, i.e. 
\begin{align} 
\label{eq:property2opt}
\lim_{T \rightarrow \infty} \frac{\bE[N^g_T(a)]}{\log T} = 0 \quad \forall a \in \cA, g \notin \gmin(a).
\end{align}
Now, turning our attention to $g_{\min}$, we have by assumption that $g_{\min}$ is the only group in $\gmin(a)$ for all $a \in \cA^{g_{\min}}$. Consequently, by \eqref{eq:property2opt}, we must have that for any optimal policy,  
$\lim_{T \rightarrow \infty} {\bE[N^{g_{\min}}_T(a)]}/{\log T} = \lim_{T \rightarrow \infty} {\bE\left[N_{T}(a)\right]}/{\log T}$ for all $a \in \cA^{g_{\min}}$. And since $\Nopta = J^{g_{\min}}(a)$ for all $a \in \cA^{g_{\min}} \cap \Asub$, \eqref{eq:property1opt} then implies that the regret for group $g_{\min}$ is precisely 
\[
\lim_{T \rightarrow \infty}
\frac{\ER^{g_{\min}}_T(\pi, \cI)}{\log T}
=
\sum_{a \in \cA^{g_{\min}}} \Delta^{g_{\min}}(a) J^{g_{\min}}(a).
\]
But this is precisely $\lim_{T} {\tilde R^{g_{\min}}_T(\cI)}/{\log T}$. 
Thus, $\ugain^{g_{\rm min}}(\pi, \cI) = 0$. 
\end{myproof}

\cref{corr:unfair} formalizes the intuition described in \cref{ex:3arm}, that it is more `efficient' to explore with group A compared to group B. Since $\OPT(A) = \theta_2 < \theta_3 = \OPT(B)$, we have $g_{\min}=A$, and therefore group A's utility gain would be zero under any regret-optimal policy. Effectively, a regret-optimal policy completely sacrifices fairness for the sake of efficiency.

The proof also illustrates that if $g_{\rm max} \triangleq  \argmax_{g \in G} \OPT(g)$ is unique, then $g_{\max}$ incurs no regret from \ii{any} shared arm in a regret-optimal policy. 
If all suboptimal arms for $g_{\max}$ are shared with another group, then $g_{\max}$ incurs zero (log-scaled) regret in an optimal policy. In summary, regret-optimal policies are unfair, and achieve perverse outcomes with the most disadvantaged groups gaining nothing and the most advantaged groups gaining the most from sharing the burden of exploration.

\begin{remark}[Regret Optimality of KL-UCB] \label{remark:klucb_optimal}
The fact that KL-UCB is regret-optimal is somewhat surprising, as the required property \eqref{eq:property2opt} seems quite restrictive at first glance. 
The intuition for this result can be explained through the 2-group, 3-arm instance of \cref{ex:3arm}, where $\theta_1 < \theta_2 < \theta_3$ and only $\theta_1$ is unknown. 
In this instance, proving \eqref{eq:property2opt} corresponds to showing that group B never pulls arm 1.
Under KL-UCB, Group A or B pulls arm 1 at time $t$ if and only if $\UCB_t(1)$ is larger than $\theta_2$ or $\theta_3$ respectively.
$\UCB_t(1)$ can be shown to be equal to $\htheta_t(1)$ plus a term of order $\sqrt{\frac{\log t}{N_t(1)}}$; we refer to the latter term as the `radius' of the UCB. 
The radius increases logarithmically over time, but decreases as there are more pulls. 
Anytime $\UCB_t(1)$ increases past $\theta_2$, group A will pull arm 1, which causes the radius to shrink. 
Since the radius grows very slowly, it is unlikely that it ever gets to a point where $\UCB_t(1)$ is larger than $\theta_3$; and hence group B ends up essentially never pulling arm 1.
The full proof of this result can be found in Appendix~\ref{sec:proof_klucb_opt}, where the stated argument is used to prove \cref{lemma:main_epoch}.
We note that this result, along with the matching lower bound \eqref{eq:lower_bound_regret}, provides a complete regret characterization of the grouped $K$-armed bandit model (without fairness concerns).
\end{remark}

\subsubsection{Discussion: Existence of a Nontrivial Trade-off.} \label{sec:tradeoff:discussion}

The unfairness resulting from \cref{corr:unfair} is fueled by the fact that exploration is costly, which leads to contention across groups on how exploration is allocated.
However, it is worth noting that not every instance exhibits this behavior.
For instance, from \cref{ex:3arm}, suppose the ordering of arm rewards was instead $\theta_2 < \theta_1 < \theta_3$.
In this case, the shared arm, arm 1, is \textit{suboptimal} for group B but \textit{optimal} for group A. 
Then, under any consistent bandit policy, the expected number of times group A pulls arm 1 is \textit{linear} in $T$,
which allows the policy to learn $\theta_1$ without incurring any regret.
This substantially benefits group B, at no cost to group A.
Note that it is impossible for group A to benefit from group B, since group B does not have access to group A's suboptimal arm (arm 2), and hence the utility gain for group A is 0 under \textit{any} policy.
There is no shared suboptimal arm between the groups, and hence there is no contention in which group should explore.
This is a degenerate instance in that the feasible region $U(\cI)$ is a \textit{line} that is on the axis corresponding group B's utility which has a \textit{single} Pareto-optimal point, where the Nash SW associated with this point is $-\infty$.
This instance does not exhibit any non-trivial trade-off in group utilities, whereas the focus of our work is driven by such a trade-off.

In the linear contextual bandit setting, there is a line of work that aims to identify instances where no exploration is needed due to sufficient diversity in the contexts (e.g., \cite{bastani2020mostly,kannan2018smoothed,hao2020adaptive}).
This phenomenon could lead to one group helping another without incurring any cost of exploration --- this is analogous to group A helping group B in the previous 3-arm example.

\subsection{Upper Bound on Nash Social Welfare} \label{sec:ub_nash_sw}

The preceding question motivates asking what is in fact possible with respect to fair outcomes. To that end, we derive an instance-dependent upper bound on the Nash SW. We may view this as a `fair' analogue to instance-dependent lower bounds on regret.

Let $\cI$ be a grouped $K$-armed bandit instance with parameter $\theta$, and let $\pi \in \conspolicies$. 
Our goal is to upper bound $\welfare(\pi, \cI)$.
We first re-write ${\ER^g_T(\pi, \cI)}/{\log T}$.
Given a policy $\pi$, for any action $a$ and group $g$, let $q_T^g(a, \pi) \in [0, 1]$ be the \ii{percentage} of times that group $g$ pulls arm $a$, out of the total number of times arm $a$ is pulled. That is, $\bE[N^g_T(a)] = q_T^g(a, \pi) \bE[N_T(a)]$, where $\sum_{g \in G} q_T^g(a, \pi) = 1$ for all $a$.
Then,
\begin{align}  \label{eq:regret_param_N_t_log}
  \frac{\ER^g_T(\pi, \cI)}{\log T} = \sum_{a \in \Asubg} \Delta^g(a) q_T^g(a, \pi) \frac{\bE[N_T(a)]}{\log T}
  \geq  \sum_{a \in \Asubg \cap \Asub} \Delta^g(a) q_T^g(a, \pi) \frac{\bE[N_T(a)]}{\log T}.
\end{align}
Recalling $\ugain^{g}(\pi, \cI) = \liminfT  \frac{\tR^g_T(\cI) - \ER^g_T(\pi, \cI)}{\log T}$,
combining \eqref{eq:disagreement_pt_regret}, \eqref{eq:regret_param_N_t_log}, and \eqref{eq:lower_bound_pulls} yields:
\[
\ugain^{g}(\pi, \cI) 
\leq
\liminfT 
\sum_{a \in \Asubg} \Delta^g(a) \left( \Ngroupa - q_T^g(a, \pi) \Nopta \mathbf{1}\{a \in \Asub\} \right).
\]
Using the definition of $\welfare(\pi, \cI)$ and taking the $\liminf$ outside of the sum gives
\[
\welfare(\pi, \cI) \leq \liminfT  \sum_{g \in \cG} \log \bigg( \sum_{a \in \Asubg} \Delta^g(a) \left( \Ngroupa - q_T^g(a, \pi) \Nopta \mathbf{1}\{a \in \Asub\}  \right)\bigg). 
\]
But since $\sum_{g \in \cG} q_T^g(a, \pi) = 1$ for every $T$ and $a$, it must be that the limit infimum above is achieved for some vector $(q^g(a))$ satisfying $\sum_{g \in G} q^g(a) = 1$ for all $a$. This immediately yields an upper bound on $\welfare(\pi, \cI)$: let $\ssY(\cI)$ be the optimal value to the following program.
\begin{equation}
\begin{aligned} 
\ssY(\cI) =  \max_{q \geq 0} \quad & \sum_{g \in \cG} \log \bigg( \sum_{a \in \Asubg} \Delta^g(a) \left( \Ngroupa - q^g(a) \Nopta \right)\bigg)  \\
\text{s.t. } 
\quad
&\sum_{g \in \cG} q^g(a) = 1 \quad \forall a \in \Asub \\
& q^g(a) = 0 \quad \forall g \in G, a \notin \Asub \cap \cA_g.
\end{aligned} 
\label{eq:fair_opt_prob}
\end{equation}
Then, we have shown that the above program yields an upper bound to the Nash SW (omitted details can be found in \cref{sec:app_f_pi_upper_bound_proof}):
\begin{theorem} \label{prop:nash_sw_ub}
For every instance $\cI$ of grouped $K$-armed bandits, $\welfare(\pi, \cI) \leq \ssY(\cI)$ for any policy $\pi \in \conspolicies$.
\end{theorem}

\section{Nash Solution and Price of Fairness}
\label{sec:pfucb}

We turn our attention in this section to constructive issues: we first develop an algorithm that achieves the Nash SW upper bound of Theorem~\ref{prop:nash_sw_ub} and thus establish that this is the Nash solution for the grouped $K$-armed bandit. In analogy to the unfairness of a regret optimal policy, it is then natural to ask whether the regret under this Nash solution is large relative to optimal regret; we show thankfully that this `price of fairness' is relatively small. 

\subsection{The Nash Solution: PF-UCB}

The algorithm we present here `Proportionally Fair' UCB (or $\PFUCB$) works as follows: at each time step it computes the set of arms that optimize the (KL) UCB for some group. Then, when a group arrives, it asks whether any arm from this set has been `under-explored', where the notion of under-exploration is measured relative to an estimated optimal solution to \eqref{eq:fair_opt_prob}. Such an arm, if available, is pulled. Absent the availability of such an arm, a greedy selection is made.  

Specifically, let $\hat \theta_t$ be the empirical mean estimate of $\theta$ at time $t$. 
Define $\hat \Delta^g(a), \hat J^g(a), \hat J(a), \hat\cA_{\text{sub}}, \hat\cA^g_{\text{sub}}$ in the same way the quantities were originally defined, but using the estimate $\hat \theta_t$ instead of $\theta$.
Then, define $(\hq^g_t(a))_{g \in \cG, a \in \cA^g}$ to be the solution to the following optimization problem with the smallest Euclidean norm:
\begin{equation*}
\begin{aligned} 
 \max_{q \geq 0} \quad & \sum_{g \in \cG} \log \bigg( \sum_{a \in \hat\cA^g_{\text{sub}}} \hat \Delta^g(a) \left( \hat J^g(a) - q^g(a) \hat J(a) \right)\bigg) \\
\text{s.t. } 
\quad
&\sum_{g \in \cG} q^g(a) = 1 \quad \forall a \in \hat\cA_{\text{sub}} \\
& q^g(a) = 0 \quad \forall g \in G, a \notin \hat\cA_{\text{sub}}\cap \cA_g.
\end{aligned} 
\tag{$P(\htheta_t)$} \label{eq:fair_opt_prob_karm}
\end{equation*}

Note that finding such a solution constitutes a tractable convex optimization problem. Finally, we denote by $\AU_t(g) \in \argmax_{a \in \cA^g} \UCB_t(a)$ the arm with the highest UCB for group $g$ at time $t$, and by $\AKL = \{ \AU_t(g) : g  \in \cG\}$  the set of arms that have the highest UCB for {\em some} group. $\PFUCB$ then proceeds as follows. At time $t$:
\begin{enumerate}
    \item If there is an available arm $a \in \cA^{g_t} \cap \AKL$ such that $N^{g_t}_t(a) \leq \hq_t^g(a) N_t(a)$, pull $a$. If there are multiple arms matching this criteria, pull one of them uniformly at random.
    \item Otherwise, pull a greedy arm $A_t \in \argmax_{a \in \cA^{g_t}} \htheta_t(a)$.
\end{enumerate}
$\PFUCB$ constitutes a Nash solution for the grouped $K$-armed bandit. 
\begin{theorem} \label{thm:main_fair_orig}
For any instance $\cI$ of grouped $K$-armed bandits, $\welfare(\pi^{\text{PF-UCB}}, \cI) = Y^*(\cI)$. 
\end{theorem}

It is worth noting that relative to the existing optimization-based algorithms for structured bandits (e.g. \cite{lattimore2017end,combes2017minimal,van2020optimal,hao2020adaptive}), $\PFUCB$ does no forced sampling. In addition, we make no requirement that the solution to the optimization problem $P(\theta)$ is unique as these existing policies require. 

We provide a rough sketch of the proof of \cref{thm:main_fair_orig}; the full proof can be found in Appendix~\ref{sec:app:pfucb_fair_proof}.
\begin{myproof}[Proof Sketch of \cref{thm:main_fair_orig}]
The main result needed is the following characterization of group regret for all groups $g \in \cG$:
 \begin{align} \label{eq:group_regret_sketch}
 \lim_{T \rightarrow \infty}\frac{\ER^g_T(\pi^{\PFUCB}, \cI)}{\log T} = \sum_{a \in \cA^g} \Delta^g(a) q_*^g(a) \Nopta,
 \end{align}
where $q_*$ is the optimal solution to \eqref{eq:fair_opt_prob} with the smallest Euclidean norm.

Fix $g \in \cG$; we provide a proof sketch of \eqref{eq:group_regret_sketch}.
Let $\gpulls(a) = \bI(g_t = g, A_t = a)$ be the indicator for group $g$ pulling arm $a$ at time $t$.
There are two reasons why $\gpulls(a)$ would occur:
(i) $a = \AU_t(g')$ for some group $g'$, or
(ii) $a = \Ag_t(g)$.
We first show that the regret from (ii) is negligible:
\begin{proposition} \label{prop:greedy_not_opt}
For any group $g$ and arm $a \in \Asubg$ suboptimal for $g$, 
\begin{align*}
\sum_{t=1}^T \Pr(\gpulls(a), A^{\text{greedy}}_t(g) = a) = O(\log \log T).
\end{align*}
\end{proposition}

\ii{Sketch of \cref{prop:greedy_not_opt}}:
Let $R_t = \{\gpulls(a), A^{\text{greedy}}_t(g) = a\}$ be the event of interest.
Since $R_t$ involves pulling arm $a$, the estimator $\htheta_t(a)$ improves every time $R_t$ occurs. 
Therefore, it is sufficient to assume that $\htheta_t(a)$ is close $\theta(a)$; i.e. bound $\sumt \Pr(R'_t)$, where $R'_t = \{\gpulls(a), A^{\text{greedy}}_t(g) = a, \htheta_t(a) \in [\theta(a)-\delta, \theta(a)+\delta]\}$ for a small $\delta > 0$.
Let $a' = \argmax_{a \in \cA^g} \theta(a)$ be the optimal arm for group $g$.
For $R_t'$ to occur, since $a$ is the greedy arm, it must be that $\htheta_t(a') \leq \theta(a)+\delta$.
Since $\UCB_t(a') \geq \theta(a')$ (w.h.p.), the definition of the UCB \eqref{eq:define_UCB} implies that $N_t(a') \leq c \log t$ for some $c > 0$; i.e. $a'$ has not been pulled often.
Lastly, we show a probabilistic version of the lower bound of \cite{lai1985asymptotically}, proving that $N_t(a')> c \log t$ with high probability.
The result follows by combining the above arguments using a carefully constructed epoch structure on the time steps.

Therefore, all of the regret stems from pulls of type (i), pulls of arms that have the highest UCB for some group.
The fact that KL-UCB is a regret-optimal algorithm implies that the number of times each arm is pulled is optimal; i.e. \eqref{eq:property1opt} holds.
Therefore, we need to show that the pulls of arm $a$ are `split' between the groups according to $(q^g_*(a))_{g \in \cG}$.
The next result pertains to the program \eqref{eq:fair_opt_prob}, which states that if the empirical estimate $\htheta_t$ is close to the true parameter $\theta$, the approximate solution $\hq_t$ is also close to the true solution $q_*$.
Let $H_t(\delta) = \bI(\htheta_t(a) \in [\theta(a)-\delta, \theta(a)+\delta]\; \forall a \in \cA)$ be the event that the estimates for all arms are within $\delta$.
\begin{proposition} \label{prop:eps_close_delta}
For any $\eps > 0$, there exists $\delta > 0$ such that if $H_t(\delta)$, then $\hq_t^g(a) \in [q_*^g(a) - \eps, q_*^g(a) + \eps]$ for all $a \in \cA$ and $g \in \cG$.
\end{proposition}
Since both \eqref{eq:fair_opt_prob} and $(P(\htheta_t))$ may have multiple optimal solutions, \cref{prop:eps_close_delta} follows from an intricate analysis of the program to show that the two corresponding optimal solutions that minimize the Euclidean norm are close when $\theta$ and $\htheta_t$ are close. 
This result implies that when we have good empirical estimates of $\theta$ (i.e. $H_t(\delta)$ is true), the policy of `following' the solution $\hq_t^g(a)$ will give us the desired `split' of pulls between groups.
The final proposition shows that we do indeed have good empirical estimates of $\theta$, and \cref{thm:main_fair_orig} follows from combining these propositions.
\begin{proposition} \label{prop:subopt_pulls_ht}
Fix any $\delta > 0$.
For any group $g$ and arm $a \in \Asubg$ suboptimal for $g$, 
\begin{align*}
\sum_{t=1}^T \Pr(\gpulls(a), A^{\text{greedy}}_t(g) \neq a, \bar{H}_t(\delta)) = O(\log \log T).
\end{align*}
\end{proposition}
\ii{Sketch of \cref{prop:subopt_pulls_ht}}:
Let $E_t = \{\gpulls(a), A^{\text{greedy}}_t(g) \neq a, \bar{H}_t(\delta)\}$.
Divide the time interval into epochs, where epoch $k$ starts at time $s_k = 2^{2^k}$.
First, we bound the number of times that $E_t$ can occur during epoch $k$ to be at most $O(\log s_{k+1})$.
Next, we define $F_k = \{H_{s_k}(\delta/2), N_{s_k}(a) > c_a \log s_k \; \forall a \in \cA \}$ to be the event that at the start of epoch $k$, all arm estimates are accurate, and that all arms have been pulled an `expected' number of times ($c_a > 0$ is an arm-specific constant).
We show $\Pr(F_k) \geq 1-O\left(\frac{1}{\log s_k}\right)$ using the probabilistic lower bound from the proof of \cref{prop:greedy_not_opt}.
Lastly, we show that conditioned on $F_k$, the probability that $\bar{H}_t(\delta)$ occurs at \ii{any} time $t$ during epoch $k$ is $O\left(\frac{1}{\log s_k}\right)$.
Combining, using $\frac{\log s_{k+1}}{\log s_k}=2$, the expected number of times that $E_t$ occurs during one epoch is $O(1)$, and the result follows since there are $O(\log \log T)$ epochs. 
\end{myproof}

\subsection{Price of Fairness} \label{sec:price_of_fairness}

Whereas $\PFUCB$ is proportionally fair, what price do we pay with respect to efficiency? To answer this question we compute an upper bound on the `price of fairness'. Specifically, define 
\[
\textrm{SYSTEM}(\cI) = \mathsmaller{\sum}_{g \in \cG} \ugain^g(\pi^{\KLUCB}, \cI) 
\ {\rm and} \  
\textrm{FAIR}(\cI) = \mathsmaller{\sum}_{g \in \cG} \ugain^g(\pi^{\PFUCB}, \cI).
\]
Recall that $\ugain^g(\pi^{\KLUCB}, \cI)$ is the reduction in group $g$'s regret under a {\em regret-optimal} policy in the grouped setting relative to the optimal regret it would have endured on its own; $\textrm{SYSTEM}(\cI)$ aggregates this reduction in regret across all groups. 
Similarly, $\ugain^g(\pi^{\PFUCB}, \cI)$ is the reduction in group $g$'s regret under the Nash solution, and $\textrm{FAIR}(\cI)$ aggregates this across groups. 
The price of fairness (PoF) asks what fraction of the optimal reduction in regret is lost to fairness: 
\begin{align*}
\pof(\cI) \triangleq 
 \frac{ \textrm{SYSTEM}(\cI) - \textrm{FAIR}(\cI) }{\textrm{SYSTEM}(\cI)}.
\end{align*}
$\pof(\cI)$ is a quantity between 0 and 1, where smaller values are preferable. 
Note that $\pof(\cI)$ is a quantity determined entirely by the feasible space of utility gains, $U(\cI)$.

Letting $r(\cI)$ be a measure of the inherent asymmetry of utility gains across groups (to be precisely defined later),
we show the following upper bound on the price of fairness:
\begin{theorem} \label{thm:general_pof}
For an instance $\cI$ of grouped $K$-armed bandits, 
$\pof(\cI)  
\leq 1 - r(\cI) \frac{2 \sqrt{G}-1}{G}$.
\end{theorem}
The proof of \cref{thm:general_pof} relies on an analysis of the price of fairness for general convex allocation problems in \cite{bertsimas2011price} and may be found in Appendix~\ref{app:pof_proofs}. 
\edit{
The key takeaway from this result is that, treating the inherent asymmetry $r(\cI)$ as a constant, the Nash solution achieves at least $O(1/\sqrt{G})$ of the reduction in regret achieved under a regret optimal solution relative to the regret incurred when groups operate separately.
One way to interpret the merit of this result is to compare to a different, commonly used fairness notion.
Under a max-min notion of fairness, the results of \cite{bertsimas2011price} imply that the fair solution would only achieve $O(1/G)$ of the reduction in regret achieved under a regret optimal solution --- this is a significantly worse rate with respect to $G$.
}

$r(\cI)$ is defined as the following.
For an instance $\cI$, let $s^g(\cI) = \sup_{\pi \in \conspolicies^{+}(\cI)} \ugain^g(\pi, \cI)$ be the maximum achievable utility gain for group $g$, where $\conspolicies^{+}(\cI) = \{\pi \in \conspolicies: \ugain^g(\pi, \cI) \geq 0 \; \forall g \in \cG\}$.
That is, $s^g(\cI)$ is the largest possible utility gain for group $g$ without any other group incurring a negative utility gain.
Then, $r(\cI) = {\min_{g \in \cG} s^g(\cI)}/{\max_{g \in \cG} s^g(\cI)}$.

Whereas the bound above depends on the topology of the instance only through $r(\cI)$, a topology specific analysis may well yield stronger results. For instance, we exploit the structure of a special class of grouped $K$-armed bandit instances to derive the following improved upper bound: 
\begin{proposition} \label{theorem:pof_specialized}
Let $\cI$ be an instance of grouped K-armed bandits such that for every arm $a \in \cA$, either $\cG_a = \cG$ or $|\cG_a| = 1$.
Then $\pof(\cI) \leq \frac{1}{2}$.
\end{proposition}
This result shows that for a specific class of topologies, the price of fairness is a constant independent of any parameters including the number of groups or the mean rewards.
In Section~\ref{sec:experiments} we study the price of fairness computationally in the context of random families of instances.

\section{Extension to Grouped Linear Contextual Bandits} \label{sec:contextual_linear_bandits}

In this section, we extend the results of Sections~\ref{sec:analyze_klucb} and \ref{sec:pfucb} to the grouped linear contextual bandit setting.
We show an upper bound on Nash SW, and then we introduce a policy called PF-OAM, which we prove to be the Nash solution for grouped contextual bandits. 
We also bound the price of fairness, 
and lastly discuss the unfairness of regret-optimal policies.

\subsection{Upper Bound on Nash Social Welfare} \label{sec:ub_linear}
We first characterize the disagreement point $\tR^g_T(\cI)$, the optimal regret that each group would incur if they were on its own. 
\cite{hao2020adaptive} characterize this quantity by proving a regret lower bound and developing a policy called optimal allocation matching (OAM) with a matching upper bound.
We denote the vector $(\Delta(m, a))_{m \in \cM, a \in \cA}$ by $\Delta$.
\cite{hao2020adaptive} show that $\tR^g_T(\cI) = \cC(\cM^g, \Delta)$, where $\cC(\cdot, \cdot)$ is the optimal value of the following optimization problem:
\begin{equation} \label{opt:min_regret_contextual}
\begin{aligned} 
\cC(\cM, \Delta) = \min_{Q \geq 0} \quad & \mathsmaller{\sum}_{m \in \cM} \mathsmaller{\sum}_{a \in \cA(m)} Q(m, a) \Delta(m, a)\\
\text{s.t. } 
\quad
& Q(a) = \mathsmaller{\sum}_{m : a \in \cA(m)} Q(m, a) \quad \forall a \in \cA \\
& (Q(a))_{a \in \cA} \in \cQ(\cM, \Delta),
\end{aligned} 
\tag{$L(\cM, \Delta)$} 
\end{equation}
where $\cQ(\cM, \Delta)$ is the following polytope ensuring the consistency of the policy:
\begin{align*}
\cQ(\cM, \Delta) = \big\{ (Q(a))_{a \in \cA} : \;
& ||a||_{H_Q^{-1}}^2 \leq \Delta(m, a)^2/2 \quad  \forall m \in \cM, a \in \cA(m) \text { s.t. } \Delta(m, a) > 0,  \\
& H_Q = \mathsmaller{\sum_{a \in \cA}} Q(a) a a^{\top} \big\}.
\end{align*}
The variable $Q(m, a)$ implies that under a regret-optimal policy, context $m$ pulls arm $a$ approximately $Q(m, a) \log T$ times.
The constraints that define $\cQ(\cM, \Delta)$ can be thought of as the analog of the lower bound of \eqref{eq:lower_bound_pulls} for the $K$-armed setting to the linear bandit setting (this was first shown in \cite{lattimore2017end}, and generalized to the contextual setting in \cite{hao2020adaptive}).

Then, using a similar series of steps as \cref{sec:ub_nash_sw} for grouped $K$-armed bandits, we can prove an upper bound on the Nash SW for a grouped linear contextual bandit instance.  
We use the variable $Q^g(m, a)$ to denote the number of times a context $m$ from group $g$ pulls arm $a$.
Let ${Z}^*(\Delta)$ be the objective value of the following optimization problem:
\begin{align} \tag{$\fairopt(\Delta)$}\label{eq:fair_opt_prob_linear}
{Z}^*(\Delta) = \max_{Q \in \Gamma(\Delta)} \quad & \sum_{g \in \cG} \log \bigg( \cC(\cM^g, \Delta) - \sum_{m \in \mathcal{M}^g} \sum_{a \in \cA(m)} \Delta(m, a) Q^g(m, a) \bigg),
\end{align}
where $\Gamma(\Delta)$ is the feasibility set, defined as
\begin{align*} 
\Gamma(\Delta) = \{ (Q^g(m, a))_{g \in \cG, m \in \cM, a \in \cA}  :  \;
& Q^g(m, a) \geq 0 \quad \forall g, m, a, \\
& Q^g(m, a) = 0 \quad \forall g, m, a : m \notin \cM^g \text{ or } a \notin \cA(m), \\
& Q(a) = \mathsmaller{\sum}_{g \in \cG} \mathsmaller{\sum}_{m \in \cM^g: a \in \cA(m)} Q^g(m, a) \quad \forall a \in \cA, \\
& (Q(a))_{a \in \cA} \in \cQ(\cM, \Delta) \}.
\end{align*} 

We show that ${Z}^*(\Delta)$ is an upper bound on the Nash SW; the proof can be found in Appendix~\ref{sec:pf:sw_ub_linear}.
\begin{theorem} \label{thm:sw_ub_linear}
For any instance $\cI$ of grouped linear contextual bandits where $Z^*(\Delta)$ is continuous at $\Delta$,
$\welfare^*(\pi^{\PFUCB}, \cI) \leq {Z}^*(\Delta)$.
\end{theorem}

\subsection{PF-OAM: Nash Solution for Grouped Linear Contextual Bandits}
We now describe PF-OAM, a Nash solution for grouped linear contextual bandits.
At a high-level, it follows the same structure as PF-UCB. 
At each time step, we solve the optimization problem $\fairopt(\hat{\Delta}_t)$ using the empirical estimate $\hat{\Delta}_t$, and the policy `follows' the solution.

At time $t$, let $G_t = \sum_{s=1}^{t-1} A_s A_s^T$, and denote by $\hat{\theta}_t = G_t^{-1} \sum_{s=1}^{t-1} A_s Y_s$ the least squares estimator.
let $\hDelta_t(m, a) = \max_{a' \in \cA(m)} \langle \hat{\theta}_t, a' - a \rangle$ be the expected regret from pulling arm $a$ with context $m$, and let $\hDelta^{\min}_{t} = \min_{m, a: \hDelta_t^m(a) > 0} \hDelta_t^m(a)$ be the smallest non-zero regret.
Let $f_T = 2(1 + 1/\log T) \log T + cd \log (d \log T))$, where $c$ is an absolute constant ($f_T \approx 2 \log T$ for large $T$).
Let $\eps_T = 1/ \log \log T$.
$g_t$ and $m_t$ are the group and context respectively corresponding to time $t$, which are both observed.

\textbf{Initialization.}
For the first $d$ time steps, choose an action $a \in \cA_t$ that is not in the span of the previous actions chosen, $\{A_1, \dots, A_{t-1}\}$. This is always possible since we assumed that $\cA(m)$ spans $\bR^d$ for all $m \in \cM$. 

The initialization phase ensures that $G_t$ is invertible.
After this, the algorithm runs the following steps at each time:
\begin{enumerate}
\item \textbf{Exploitation.}
Let $\cD_t$ be the event that for all contexts $m \in \cM$ and arms $a \in \cA(m)$ such that $\hDelta_{t}(m, a) > 0$,
\begin{align*}
||a||^2_{G_{t}^{-1}}	\leq  \frac{(\hDelta_{t}(m, a))^2}{f_T} . 
\end{align*}
If $\cD_t$ holds, then we exploit, by pulling a greedy action:
$A_t \in \argmax_{a \in \cA_t}	\langle a, \htheta_t \rangle$.

\item \textbf{Solve Optimization Problem.}
Solve $\fairopt(\hat{\Delta}_t)$, and denote a solution by $\hat{Q}_t^{g}(m, a)$.
If there are multiple optimal solutions, choose the one with the smallest Euclidean norm.

\item \textbf{Forced Exploration.}
Let $S(t) = \sum_{s=1}^t \bI(\cD_s^c)$ be the number of times that the algorithm did not exploit in the past.

\item \textbf{Targeted Exploration.}
If there is an available arm $a \in \cA_t$ such that
\begin{align}  \label{eq:texplore_condition}
N_t^{g_t}(m_t, a) < \hat{Q}_t^{g_t}(m_t, a) f_T/2,
\end{align}
then pull this arm. 
Break ties arbitrarily if multiple arms satisfy this condition.

\item \textbf{Exploitation as backup.}
Pull $A_t \in \argmax_{a \in \cA_t}	\langle a, \htheta_t \rangle$.
\end{enumerate}

We show that this policy achieves the highest possible Nash SW, under some technical assumptions on the optimal solution to \eqref{eq:fair_opt_prob_linear}.
The proof can be found in Appendix~\ref{sec:app:pfoam_nash}.
\begin{theorem} \label{thm:pfoam_nash_solution}
Let $\cI$ be an instance of grouped linear contextual bandits where
the solution $(Q^g(m, a))$ to \eqref{eq:fair_opt_prob_linear} is unique, finite and continuous at $\Delta$.
Then, $\welfare(\pi^{\text{PF-OAM}}, \cI) = {Z}^*(\Delta)$.
\end{theorem}

\begin{remark}[Comparison to OAM] \label{remark:pfoam_vs_oam}
PF-OAM is inspired by the OAM policy from \cite{hao2020adaptive}, but there are major differences in the algorithm as well as its analysis.
At a high level, PF-OAM changes the optimization problem solved at each time step to one that maximizes Nash SW, rather than minimizing regret. 
Further, the condition to check whether to exploit ($\cD_t$) was modified, step 5 was changed to exploitation (which used to be LinUCB), and steps 3-5 were simplified.
The proof of \cref{thm:pfoam_nash_solution} involves significant changes from the analysis of OAM, which can be found in Appendix~\ref{sec:app:pfoam_nash}.
\end{remark}

\begin{remark}[Comparison to PF-UCB] \label{remark:pfoam_vs_pfucb}
As grouped linear contextual bandits is a generalization of grouped $K$-armed bandits, PF-OAM can be applied to the latter model.
However, the theoretical result for PF-UCB (\cref{thm:main_fair_orig}) require fewer assumptions than PF-OAM (\cref{thm:pfoam_nash_solution}), and hence our results are stronger in the $K$-arm setting via PF-UCB.
The policies are also algorithmically distinct in how it decides to explore: PF-UCB relies on querying the UCB policy, whereas PF-OAM relies on a combination of forced exploration and the optimization problem.
\end{remark}

\subsection{Price of Fairness} \label{sec:pof_linear}
The PoF result of \cref{thm:general_pof} extends to the contextual linear bandit setting, with an additional technical assumption that all utility gains are non-negative under the efficient outcome.

\begin{theorem} \label{prop:pof_linear}
Let $\cI$ be an instance of grouped linear contextual bandits where $\ugain^g(\pi^{\text{OAM}}, \cI) \geq 0$ for all groups $g$.
Then, $\pof(\cI) \leq 1 - r(\cI) \frac{2 \sqrt{G}-1}{G}$.
\end{theorem}

In the $K$-arm setting (\cref{thm:general_pof}), it can be shown that the utility gains of all groups are non-negative under the regret-optimal policy and hence such an assumption is unnecessary.
However, this may not necessarily be the case under the more general model of grouped linear contextual bandits.
The PoF is still well-defined when this assumption does not hold, hence bounding the PoF without such an assumption remains an open question.

\subsection{Unfairness of Regret-Optimal Policies} \label{sec:unfairness_linear}
As for characterizing the (un)fairness of regret-optimal policies, the same general intuition from the $K$-armed setting (\cref{corr:unfair}) will extend to the contextual setting: a regret-optimal policy will explore with those contexts who incur less regret from exploring.
Note that the solution to the optimization problem \eqref{opt:min_regret_contextual} assigns which contexts should explore and hence incur regret.
Since a group $g$ is composed of a subset of contexts, $\cM^g \subseteq \cM$, the group will be worse off if $\cM^g$ contains contexts that are assigned to explore.
Unlike the $K$-armed setting, identifying exactly which contexts will be assigned to explore is complex; it involves characterizing the solution to \eqref{opt:min_regret_contextual}.
For any given instance, the utility gain of group $g$ under a regret-optimal policy can be computed by solving \eqref{opt:min_regret_contextual} and $(L(\cM^g, \Delta))$.
One may be able to derive more general insights by analyzing the structure of the optimal solution to \eqref{opt:min_regret_contextual}; we leave this as a valuable open direction for future work.

\section{Experiments} \label{sec:experiments}

We consider three sets of experiments. 
The first seeks to understand the PoF for the grouped $K$-armed bandit in synthetic instances to shed further light on the impact of topology. 
In the second set of experiments, we evaluate the group regret trajectories over time under various policies grouped $K$-armed bandits; the goal is to understand how to theoretical results that rely on asymptotics translate to finite time. 
The third set of experiments is a real-world case study that returns to the warfarin dosing example discussed in motivating the paper where we seek to understand unfairness under a regret optimal policy and the extent to which the Nash solution can mitigate this problem.

\subsection{Price of Fairness} \label{sec:exp:pof}
In the first set of experiments, we evaluate the PoF on random instances to understand how it compares to the worst case bounds from \cref{sec:price_of_fairness}.
We generate random instances of the grouped $K$-armed bandit model and compute the PoF for each instance.
We consider two generative models that differ in how the bipartite graph matching groups to available arms is generated:
\begin{itemize}
  \item \ii{i.i.d.:} Each edge appears independently with probability 0.5, and $K=10$ is fixed. The mean reward of each arm is i.i.d. $\mathrm{U(0, 1)}$.
  \item \ii{Skewed:} $K=G+1$, and a group $g \in \{1, \dots, G-1\}$ has access to arms $\{g, G\}$, while the last group $g=G$ has access to all arms. The arm rewards satisfy $\theta(1) = \dots = \theta(G-1) < \theta(G) < \theta(G+1)$, which are generated randomly by sorting three i.i.d. $\mathrm{U(0, 1)}$ random variables.
\end{itemize}

For each of the two methods, we vary $G \in \{3, 5, 10, 50\}$, and generate 500 random instances for each parameter setting.
Given an instance $\cI$, note that one can exactly compute the asymptotic regret 
under $\KLUCB$, using \eqref{eq:lower_bound_regret}, as well as under $\PFUCB$, using \eqref{eq:group_regret_sketch}.
Therefore, $\pof(\cI)$ can be computed from $\cI$.

The results in \cref{tab:pof_Karm} shows that the PoF is very small in the `i.i.d.' setting, and contrary to the worst-case bound of \cref{thm:general_pof}, the PoF actually decreases as $G$ gets large. This suggests an interesting conjecture for future research: the PoF may actually grow negligible in large random bandit instances. The `Skewed' structure is motivated by our PoF analysis where we see that the PoF increases -- albeit slowly -- with $G$. 
These empirical results show that the worst-case bound of \cref{thm:general_pof} may not be representative of what one should expect in practice --- the worst-case upper bound may be quite pessimistic compared to an `average' instance.

\begin{table}[h]
\TableSpaced 
\caption{
The median and 95th percentile of the PoF for synthetic instances of the grouped $K$-armed bandit over 500 runs for each parameter setting.
}
\label{tab:pof_Karm}
\begin{center}
\begin{tabular}{c|cccc|cccc}
     \toprule
      & \multicolumn{4}{c|}{i.i.d.} &  \multicolumn{4}{c}{Skewed} \\
      $G$ & 3 & 5 & 10 & 50 & 3 & 5 & 10 & 50 \\
     \midrule 
 Median &  0.073 & 0.054 & 0.040 & 0.015 & 0.327 & 0.407 & 0.454 & 0.521 \\
95th percentile & 0.289 & 0.177 & 0.142 & 0.063 & 0.632 & 0.764 & 0.845 & 0.924 \\
\bottomrule 
\end{tabular}
\end{center}
\end{table}

\subsection{Group Regret Trajectories for \cref{ex:3arm}} \label{sec:experiments_trajectories}
\begin{figure}[h]
\begin{center}
\includegraphics[scale=0.5]{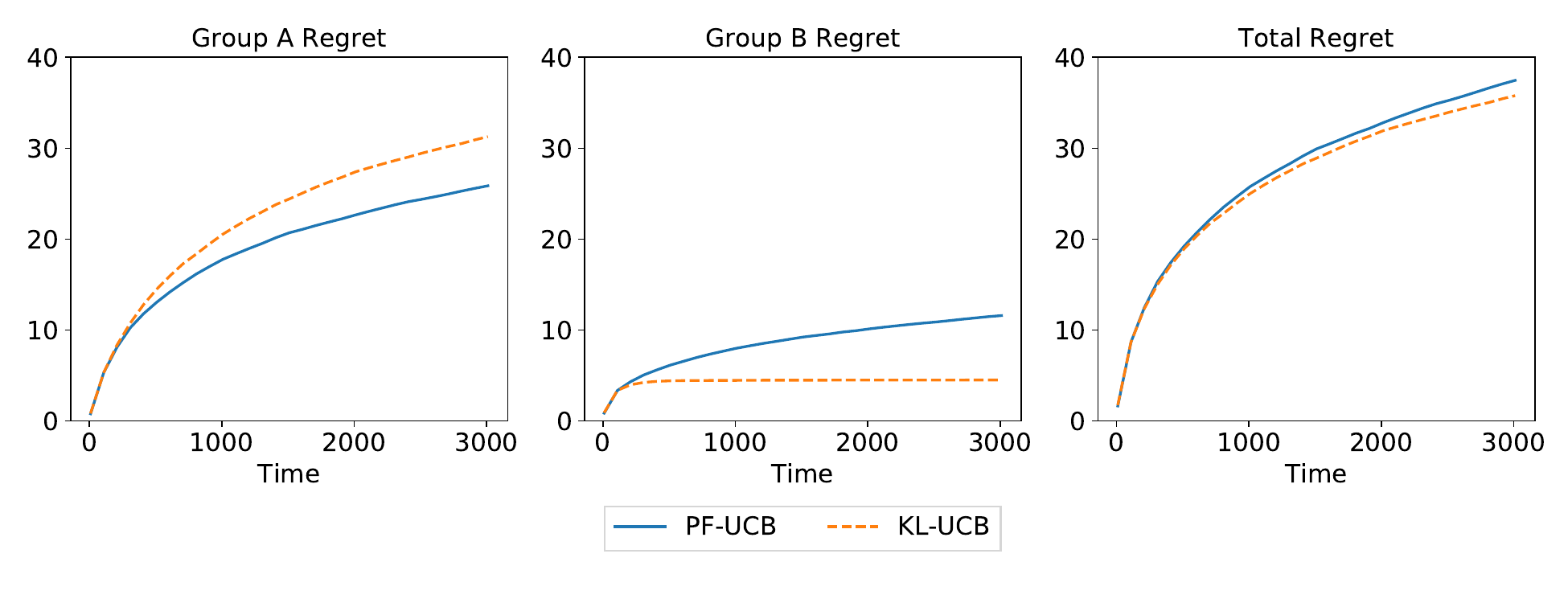}
\caption{Group regret as well as total regret over time under PF-UCB and KL-UCB for a grouped $K$-armed bandit instance with three arms and two groups. The plot shows an average over 500 runs.}
\label{fig:group_regret_trajectory}
\end{center}
\end{figure}

As our theoretical results are based on an asymptotic regime, these experiments aims to understand how well the theory translates to finite time.
We run KL-UCB and PF-UCB on a simple grouped $K$-armed bandit instance from \cref{ex:3arm} and we plot each group's regret over time.
There are three arms and two groups called A and B, where $\cA^A = \{1, 2\}$, and $\cA^B =\{1, 3\}$, and the probability of arrival for each group are equal ($p_A = p_B = 0.5$).
Let $\theta = (0, 0.5, 0.7)$,
and let the reward from pulling arm $a$ be distributed as $\text{Normal}(\theta_a, 1)$.
Then, arm 1 is the only suboptimal arm for either group, where $\Delta^A(1) = 0.5$ and $\Delta^B(1) = 0.7$.
We run KL-UCB and PF-UCB for $T=3000$ time steps, and the trajectory of each group's regret, as well as the total regret, is plotted over this horizon in \cref{fig:group_regret_trajectory}.

\textbf{Regret trajectories.}
The trajectories in \cref{fig:group_regret_trajectory} illustrate exactly the main findings we expected from the theory. 
Firstly, the trajectories of KL-UCB support the fact that KL-UCB is regret-optimal and is therefore unfair.
The theory implies that any regret-optimal policy should satisfy \eqref{eq:property2opt}; that is, 
a suboptimal arm (arm 1 in this instance) should only be pulled by the group with the smallest $\OPT(g)$ (group A in this instance).
Indeed, under KL-UCB, group B's regret stays essentially constant after around 200 time steps, which implies that group B almost never pulls arm 1 after then.
Under KL-UCB, the burden of exploration (pulling arm 1) is entirely put on group A.

On the other hand, under PF-UCB, the regret for both groups increases over time, implying that group B continues to pull arm 1.
This allows group A's regret to decrease as compared to KL-UCB, as the burden of exploration is now divided amongst both groups.
As expected, the total regret incurred is higher under PF-UCB, albeit a small amount (total regret is 4.7\% higher under PF-UCB at $T=3000$).
Both of these observations hold under a small number of time steps.

\begin{figure}[h]
\begin{center}
\includegraphics[scale=0.5]{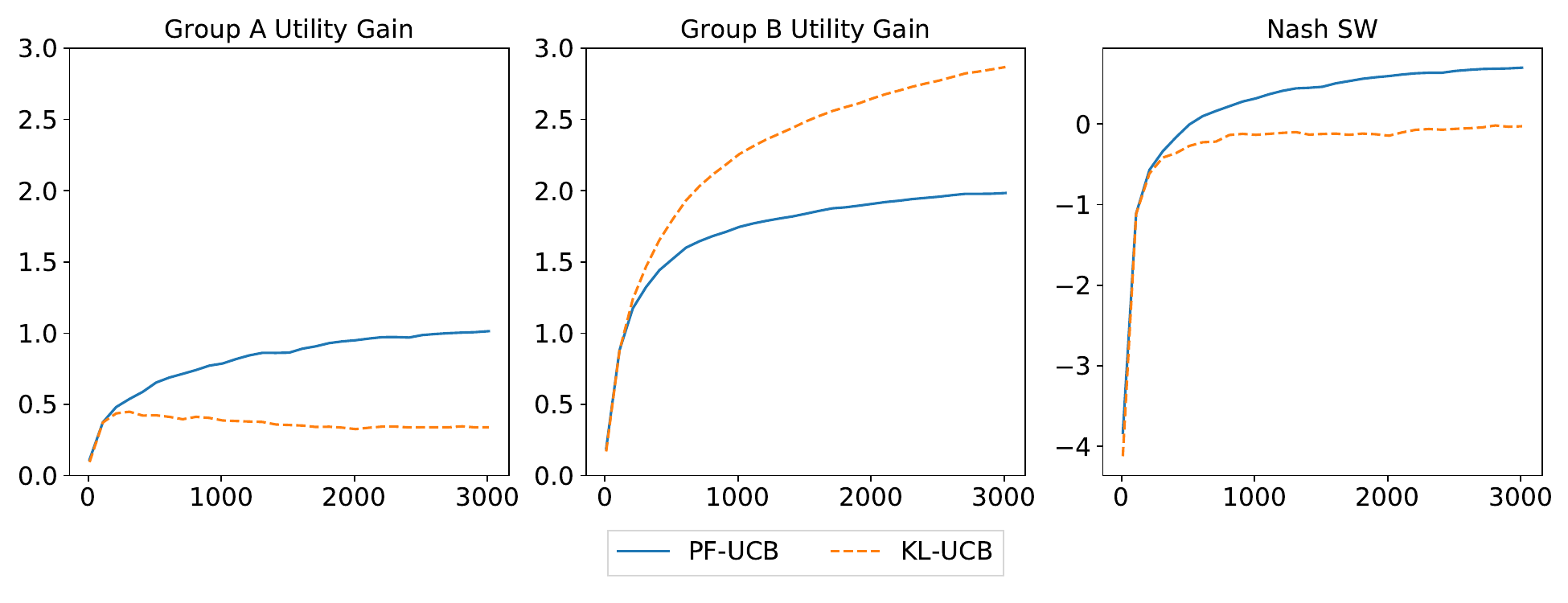}
\caption{Finite-time utility gains and Nash SW over time under PF-UCB and KL-UCB for a grouped $K$-armed bandit instance with three arms and two groups.}
\label{fig:utility_gain_trajectory}
\end{center}
\end{figure}

\textbf{Finite-time utility gains.}
Next, we evaluate a notion of utility gains in finite-time by computing
$(\tR^g_T(\cI) - \ER^g_T(\pi, \cI))/\log T$ for finite $T$, which is the definition of utility gain from \eqref{eq:def_ugain} without the $\liminf$.
We compute $\tR^g_T(\cI)$ for each group $g$, the regret at the disagreement point, by running KL-UCB on the single-group instance $\cI_g$ (as defined in \cref{sec:fairness_framework}).
Then, we can use these finite-time utility gains to compute a finite-time Nash SW.
\cref{fig:utility_gain_trajectory} shows the trajectories of the finite-time utility gains and Nash SW over time.
The utility gains for each group quickly diverge across the two policies, where PF-UCB results in a more `balanced' set of utilities across groups.
This results in a higher Nash SW under PF-UCB than KL-UCB within a short time frame.
The price of fairness at $T=3000$ is 6.5\%.

\subsection{Case Study: Warfarin Dosing}

Warfarin is a common blood thinner whose optimal dosage vastly varies across patients.
We perform an empirical case study on learning the optimal personalized dose of warfarin, modeled as a linear contextual bandit (as was done in \cite{bastani2020online}).
We use the race and age of patients as groups, 
and we evaluate the utility gains of the groups under the regret-optimal and fair outcomes, 
using the results for grouped linear contextual bandits from \cref{sec:contextual_linear_bandits}.

\textbf{Data:}
We use a publicly available dataset collected by PharmGKB \citep{whirl2012pharmacogenomics} containing data on 5700 patients who were treated with warfarin from 21 research groups over 9 countries.
The data contains demographical, clinical, and genetic covariates for each patient, as well as the optimal personalized dose of warfarin that was found by doctors through trial and error.

\textbf{Contexts:}
The grouped linear contextual bandit model assumes a finite number of contexts, and the computation of the optimization problems \eqref{opt:min_regret_contextual} and \eqref{eq:fair_opt_prob_linear} scales with the number of contexts.
Hence, for tractability, 
we use six features that are the most predictive of warfarin dosage, and for each feature, we discretize them into two bins using the empirical median.
The five features that we use are: age, weight, race, whether the patient was taking another drug (amiodarone), and two binary features capturing whether the patient has a particular genetic variant of genes Cyp2C9 and VKORC1, two genes that are known to affect warfarin dosage \citep{takeuchi2009genome}.
We also add an additional feature for the intercept, resulting in 6 features in total.

\textbf{Groups:}
We group the patients either by race or age. 
There are three distinct races which we label as A, B ,and C. 
For age, we split the patients into two age groups, where the threshold age was 70.

\textbf{Rewards:}
We bin the optimal dosage levels into three arms as was done in \cite{bastani2020online}: Low (under 3 mg/day), Medium (3-7 mg/day), and High (over 7 mg/day).
To ensure that the model is correctly specified, for each arm, we train a linear regression model using the entire dataset from the five contexts to the binary reward on whether the optimal dosage for that patient belongs in that bin.\footnote{This linear regression step is done to remove model misspecification errors, 
as the focus of this study is not to show that the linear model is a good fit for this dataset (this was demonstrated in \cite{bastani2020online}).
}
Let $\theta_a \in \bR^6$ be the learned linear regression parameter for each arm.
To model this as grouped linear contextual bandits as described in \cref{sec:grouped_linear_contextual_bandit}, we let $d=18$ and let $\theta = (\theta_{1}, \theta_{2}, \theta_{3}) \in \bR^d$.
When a patient with covariates $X \in \bR^6$ arrives at time $t$, the actions available are $\cA_t = \{(X, \zerovec, \zerovec), (\zerovec, X, \zerovec), (\zerovec, \zerovec, X)\}$, and their expected reward from arm $a$ is $\langle X, \theta_a \rangle$ for $a \in \{1, 2, 3\}$.

\textbf{Algorithms:}
We assume a patient is drawn i.i.d. from the dataset at each time step, and we compute the asymptotic group regret 
from the regret-optimal and fair policies, as well as the disagreement point, using the following method:
\begin{itemize}
  \item  \ii{Regret-optimal: }
Using the true parameter $\theta$, we solve \eqref{opt:min_regret_contextual} and obtain solution $(Q(m, a))_{m \in \cM, a \in \cA}$.
Then, the total (log-scaled) regret incurred by context $m$ is $\sum_{a \in \cA} \Delta(m, a) Q(m, a)$.
Since we assume the group arrivals are i.i.d., for each context, we allocate the regret to groups in proportion to the group's frequency.
That is, for each $m$, let $(w^g(m))_{g \in \cG}$, $\sum_{g \in \cG} w^g(m) = 1$ be the empirical distribution of groups among patients with context $m$.
Then, the total regret assigned to group $g$ is $\sum_{m \in [M]} w^g(m) \sum_{a \in \cA} \Delta(m, a) Q(m, a)$.
\item 
\ii{Fair: }
Using the true parameter $\theta$, we solve \eqref{eq:fair_opt_prob_linear} and obtain solution $(Q^g(m, a))_{g \in \cG, m \in \cM, a \in \cA}$.
Then, the group regret for $g$ is
$\sum_{m \in \cM^g}  \sum_{a \in \cA} \Delta(m, a) Q^g(m, a)$.
\item \ii{Disagreement point}: For each group $g$, the regret at the disagreement point, $\tR^g(\cI)$, is computed as the optimal value of $(L(\cM^g, \Delta))$, where $\cM^g$ are the set of contexts associated with group $g$.
\end{itemize}

The above computations provide asymptotic, log-scaled regret for each group. 
Then, the utility gain for each group for a policy ($\ugain^g(\pi, \cI)$) is computed by subtracting the group regret from the regret at the disagreement point.
The resulting utility gains for each group, as well as the total utility gain, can be seen in \cref{tab:oam_pfoam_regret}.

\begin{table}[h]
\TableSpaced 
\caption{
Utility gains ($\ugain^g(\pi, \cI)$) for each group under OAM (`Regret-optimal') and PF-OAM (`Fair') policies, where groups are either based on race or age.
}
\label{tab:oam_pfoam_regret}
\begin{center}
\begin{tabular}{cc|cccc|ccc}
     \toprule
      & & \multicolumn{4}{c|}{Race}  & \multicolumn{3}{c}{Age} \\
      & & A & B & C & Total & $< 70$ & $\geq 70$ &  Total\\
     \midrule 
\multirow{3}{*}{Regret} & Disagreement point &   25.6 & 74.8 & 78.6 & 179.1 &164.7 & 78.0 & 242.8 \\
& Regret optimal & 1.9 & 5.6 & 71.1 & 78.6& 151.6 & 23.2 & 174.8 \\
& Fair  &  0.0 & 25.4 & 54.0 & 79.4 & 149.3 & 29.3 & 178.7\\
     \midrule 
\multirow{2}{*}{Utility Gain} 
& Regret-optimal &  23.7 & 69.2 & 7.6 & 100.4 & 13.1 & 54.9 & 68.0  \\ 
& Fair & 25.6 & 49.4 & 24.6 & 99.6 & 15.4 & 48.7 & 64.1 \\
     \midrule 
& Relative change & +8.0\% & -28.6\% & +223.7\% & -0.80\%  & +17.6\% & -11.3\% & -6.1\% \\
\bottomrule 
\end{tabular}
\end{center}
\end{table}

\subsubsection{Results and Discussion.} \label{sec:exp:warfarin_discussion}
The results in \cref{tab:oam_pfoam_regret} show that for either groups based on race and age, the fair solution effectively `balances out' the utility gains across groups with a very small decrease in efficiency.
In both groupings, the group whose utility increased in the fair outcome compared to the regret-optimal outcome is the group who has the smallest absolute utility gain (group B for race and $\geq 70$ for age).
As expected, the total utility decreased, but only by 0.8\% and 6.1\% for race and age groupings respectively

The impact of incorporating fairness is higher under race groups compared to age groups.
One difference between the two groupings is that age was included in the feature vector, whereas race was not.
This implies that the context distributions for the two age groups do not overlap, whereas there is overlap under race groups.
Therefore, there is a potential for more `shared learning' across race groups.
Since a context can be associated with more than one race group, learning which arm is optimal for that context will directly help multiple groups.
On the other hand, a context is only associated with one age group, and therefore learning which arm is optimal for that context is only useful for one group; however, shared learning is still possible through learning the common unknown parameter $\theta$.
Therefore, there is a smaller opportunity for learning across groups if the group identity is part of the context, but the effect of incorporating fairness is still significant under this setting.

These results demonstrate the impact that accounting for fairness can have in a real-world application.
Specifically, we derived a bandit instance from a dataset on warfarin dosing, using the setup that was considered in \cite{bastani2020online}.
On this instance, the utility of the group that is the worst off can be improved significantly with an almost negligible impact on total utility, by essentially `rearranging' how exploration is shared between the groups.

\begin{remark}[Comparison to \cite{bastani2020online}] \label{remark:warfarin_comparison_to_bastani}
The only difference of our setup compared to the warfarin experiments in \cite{bastani2020online} is that 
the contexts are discretized into a finite set.
This discretization only serves a computational purpose --- without this, $(\bar{L}(\Delta))$ is a semi-infinite optimization problem, which is computationally intractable.
  This semi-infinite program stems back to a lower bound from \cite{graves1997asymptotically}, which all of the `optimization-based' bandit policies are based off of.
  Even though $|\cM|$ must be finite, we note that regret does not scale with $|\cM|$.
\end{remark}

\section{Discussion and Conclusion} \label{sec:conclusion}

\subsection{Alternative Fairness Notions} \label{sec:conclusion:modelling_assumptions}

\textbf{Axioms.}
The framework we propose is based on a set of axioms listed in \cref{sec:model:background}, 
which were the axioms studied when the cooperative bargaining problem was initially introduced in \cite{nash1950bargaining}.
However, it is possible to modify the axioms to derive other notions of fairness.
For example, \cite{kalai1975other} studies the bargaining problem under a different set of axioms, namely, when the independence of irrelevant alternatives axioms is replaced with a different one called monotonicity.
One could also study $\alpha$-fairness \citep{mo2000fair}, which interpolates between proportional and max-min notions of fairness.

The development of both policies derived in this work, PF-UCB and PF-OAM, can be extended to capture these fairness notions by modifying objective function of the optimization problem that is solved at each time step.
This would be a natural candidate policy that achieves this new notion of fairness; however, further analysis would be needed to theoretically guarantee that this is indeed the case.
Firstly, one needs to show that the solution to the optimization problem is indeed the unique solution that satisfies the new axioms.
This does not follow automatically because the feasible set $U(\cI)$ is not necessarily convex (this was shown in Appendix~\ref{app:nash_sol_unique} for the Nash solution).
Then, one needs to show that the utilities that result from the new bandit policy does indeed converge to the desired outcome.
The main result needed to show this is to show a \textit{stability} property on the optimization problem --- i.e. when the estimated parameter $\htheta_t$ is close to the true parameter $\theta$, then the solutions to the optimization problems are also close.
This seemingly simple property requires quite an involved analysis to prove; 
it essentially boils down to proving the conditions of Berge's maximum theorem \citep{berge1963topological}.
\cref{prop:eps_close_delta} shows this property for the Nash solution.

\textbf{Scope of the framework.}
Taking a higher level perspective, our work studies fairness with respect to an issue caused by the \textit{learning} process.
Exploration is only necessary because the parameter $\theta$ is unknown, and we study how to fairly apportion this necessary exploration between groups.
We assume that the `ideal' outcome is to assign everyone their best arm, and therefore if $\theta$ were known, the problem goes away (since exploration is not needed).
Now, there are other fairness issues that can arise in an algorithmic decision making setting that do not relate to learning.
For example, in our model, the asymmetry in access to arms across groups could arguably be considered unfair. 
Such an issue is indeed a concern related to fairness, but it is not related to \textit{learning};
i.e. the problem persists even if $\theta$ were known perfectly. 
Therefore, such a problem, though it relates to fairness, is an orthogonal issue to the problem of fair exploration that this work focuses on.

\subsection{Conclusion and Future Directions}

This paper provides the first framework in evaluating fairness for allocating the burden of exploration in an online learning setting. Though the phenomenon motivating this problem has been previously observed \citep{jung2020quantifying,raghavan2018externalities}, this work contains the first \ii{positive} results to the best of our knowledge. Our work establishes a rigorous, axiomatic framework that can be readily applied to a plethora of different learning models.

In terms of future directions, one valuable direction would be to formally modify the framework to a finite-time setting.
One of the main non-trivialities in doing this is in defining the disagreement point, which represents the best that a group could do if the group was on its own.
Since tight finite-time regret bounds are often difficult to derive, it would be ideal to define utility gains in a way that does not depend the existence of such results.
Second, there are a couple of technical questions that this paper raises.
The first is on the price of fairness, and whether the theoretical results can be improved, perhaps for a certain class of instances.
Another is to derive structural results on the unfairness of regret-optimal policies for the linear contextual bandit setting (as was described in \cref{sec:unfairness_linear}).
Lastly, another direction is to modify the modeling assumptions, some of which were described in \cref{sec:conclusion:modelling_assumptions}.
For example, if we used the axioms of \cite{kalai1975other} to define fairness, does the same flavor of policy achieve this new notion of fairness, and how does the price of fairness translate to this setting?

At a higher level, this work takes a step in understanding the granular behavior of algorithms and identifying systematic disparities in an algorithm's impact across a population.
Specifically, our work evaluates the granular impact of bandit algorithms, and we show that there is actually a
very significant systematic disparity: any algorithm whose goal is to minimize total regret will take advantage of and only explore with certain subpopulations.
Now, whether such a behavior is an undesirable or not (and whether one should instead use a `fair' policy) completely depends on the application and context at hand.
As researchers in algorithm design, we believe that it is important to try to understand the low-level behavior of the algorithms we design and to lay out any systematic variation of impact that deploying such an algorithm can have.

\OneAndAHalfSpacedXI
\bibliographystyle{informs2014}
\bibliography{ref}

\begin{thebibliography}{47}
\providecommand{\natexlab}[1]{#1}
\providecommand{\url}[1]{\texttt{#1}}
\providecommand{\urlprefix}{URL }

\bibitem[{Agarwal et~al.(2014)Agarwal, Long, Traupman, Xin, \protect\BIBand{}
  Zhang}]{agarwal2014laser}
Agarwal D, Long B, Traupman J, Xin D, Zhang L (2014) Laser: A scalable response
  prediction platform for online advertising. \emph{Proceedings of the 7th ACM
  international conference on Web search and data mining}, 173--182.

\bibitem[{Bastani \protect\BIBand{} Bayati(2020)}]{bastani2020online}
Bastani H, Bayati M (2020) Online decision making with high-dimensional
  covariates. \emph{Operations Research} 68(1):276--294.

\bibitem[{Bastani et~al.(2020)Bastani, Bayati, \protect\BIBand{}
  Khosravi}]{bastani2020mostly}
Bastani H, Bayati M, Khosravi K (2020) Mostly exploration-free algorithms for
  contextual bandits. \emph{Management Science} .

\bibitem[{Berge(1963)}]{berge1963topological}
Berge C (1963) Topological spaces, oliver and boyd ltd. \emph{Edinburgh and
  London Occurrence Handle} 114.

\bibitem[{Bergenstal et~al.(2019)Bergenstal, Johnson, Passi, Bhargava, Young,
  Kruger, Bashan, Bisgaier, Isaman, \protect\BIBand{}
  Hodish}]{bergenstal2019automated}
Bergenstal RM, Johnson M, Passi R, Bhargava A, Young N, Kruger DF, Bashan E,
  Bisgaier SG, Isaman DJM, Hodish I (2019) Automated insulin dosing guidance to
  optimise insulin management in patients with type 2 diabetes: a multicentre,
  randomised controlled trial. \emph{The Lancet} 393(10176):1138--1148.

\bibitem[{Berry(2012)}]{berry2012adaptive}
Berry DA (2012) Adaptive clinical trials in oncology. \emph{Nature reviews
  Clinical oncology} 9(4):199.

\bibitem[{Berry(2015)}]{berry2015brave}
Berry DA (2015) The brave new world of clinical cancer research: adaptive
  biomarker-driven trials integrating clinical practice with clinical research.
  \emph{Molecular oncology} 9(5):951--959.

\bibitem[{Bertsimas et~al.(2011)Bertsimas, Farias, \protect\BIBand{}
  Trichakis}]{bertsimas2011price}
Bertsimas D, Farias VF, Trichakis N (2011) The price of fairness.
  \emph{Operations research} 59(1):17--31.

\bibitem[{Chen et~al.(2018)Chen, Frazier, \protect\BIBand{}
  Kempe}]{chen2018incentivizing}
Chen B, Frazier P, Kempe D (2018) Incentivizing exploration by heterogeneous
  users. \emph{Conference On Learning Theory}, 798--818 (PMLR).

\bibitem[{Combes et~al.(2017)Combes, Magureanu, \protect\BIBand{}
  Proutiere}]{combes2017minimal}
Combes R, Magureanu S, Proutiere A (2017) Minimal exploration in structured
  stochastic bandits. \emph{arXiv preprint arXiv:1711.00400} .

\bibitem[{Frazier et~al.(2014)Frazier, Kempe, Kleinberg, \protect\BIBand{}
  Kleinberg}]{frazier2014incentivizing}
Frazier P, Kempe D, Kleinberg J, Kleinberg R (2014) Incentivizing exploration.
  \emph{Proceedings of the fifteenth ACM conference on Economics and
  computation}, 5--22.

\bibitem[{Garivier \protect\BIBand{} Capp{\'e}(2011)}]{garivier2011kl}
Garivier A, Capp{\'e} O (2011) The kl-ucb algorithm for bounded stochastic
  bandits and beyond. \emph{Proceedings of the 24th annual conference on
  learning theory}, 359--376.

\bibitem[{Gillen et~al.(2018)Gillen, Jung, Kearns, \protect\BIBand{}
  Roth}]{gillen2018online}
Gillen S, Jung C, Kearns M, Roth A (2018) Online learning with an unknown
  fairness metric. \emph{arXiv preprint arXiv:1802.06936} .

\bibitem[{Goldenshluger \protect\BIBand{}
  Zeevi(2013)}]{goldenshluger2013linear}
Goldenshluger A, Zeevi A (2013) A linear response bandit problem.
  \emph{Stochastic Systems} 3(1):230--261.

\bibitem[{Graepel et~al.(2010)Graepel, Candela, Borchert, \protect\BIBand{}
  Herbrich}]{graepel2010web}
Graepel T, Candela JQ, Borchert T, Herbrich R (2010) Web-scale bayesian
  click-through rate prediction for sponsored search advertising in microsoft's
  bing search engine. \emph{ICML}.

\bibitem[{Graves \protect\BIBand{} Lai(1997)}]{graves1997asymptotically}
Graves TL, Lai TL (1997) Asymptotically efficient adaptive choice of control
  laws incontrolled markov chains. \emph{SIAM journal on control and
  optimization} 35(3):715--743.

\bibitem[{Hao et~al.(2020)Hao, Lattimore, \protect\BIBand{}
  Szepesvari}]{hao2020adaptive}
Hao B, Lattimore T, Szepesvari C (2020) Adaptive exploration in linear
  contextual bandit. \emph{International Conference on Artificial Intelligence
  and Statistics}, 3536--3545 (PMLR).

\bibitem[{Immorlica et~al.(2018)Immorlica, Mao, Slivkins, \protect\BIBand{}
  Wu}]{immorlica2018incentivizing}
Immorlica N, Mao J, Slivkins A, Wu ZS (2018) Incentivizing exploration with
  selective data disclosure. \emph{arXiv preprint arXiv:1811.06026} .

\bibitem[{Jiang \protect\BIBand{} Liew(2005)}]{jiang2005proportional}
Jiang LB, Liew SC (2005) Proportional fairness in wireless lans and ad hoc
  networks. \emph{IEEE Wireless Communications and Networking Conference,
  2005}, volume~3, 1551--1556 (IEEE).

\bibitem[{Joseph et~al.(2016)Joseph, Kearns, Morgenstern, \protect\BIBand{}
  Roth}]{joseph2016fairness}
Joseph M, Kearns M, Morgenstern J, Roth A (2016) Fairness in learning: Classic
  and contextual bandits. \emph{arXiv preprint arXiv:1605.07139} .

\bibitem[{Jung et~al.(2020)Jung, Kannan, \protect\BIBand{}
  Lutz}]{jung2020quantifying}
Jung C, Kannan S, Lutz N (2020) Quantifying the burden of exploration and the
  unfairness of free riding. \emph{Proceedings of the Fourteenth Annual
  ACM-SIAM Symposium on Discrete Algorithms}, 1892--1904 (SIAM).

\bibitem[{Kalai \protect\BIBand{} Smorodinsky(1975)}]{kalai1975other}
Kalai E, Smorodinsky M (1975) Other solutions to nash's bargaining problem.
  \emph{Econometrica: Journal of the Econometric Society} 513--518.

\bibitem[{Kaneko \protect\BIBand{} Nakamura(1979)}]{kaneko1979nash}
Kaneko M, Nakamura K (1979) The nash social welfare function.
  \emph{Econometrica: Journal of the Econometric Society} 423--435.

\bibitem[{Kannan et~al.(2017)Kannan, Kearns, Morgenstern, Pai, Roth, Vohra,
  \protect\BIBand{} Wu}]{kannan2017fairness}
Kannan S, Kearns M, Morgenstern J, Pai M, Roth A, Vohra R, Wu ZS (2017)
  Fairness incentives for myopic agents. \emph{Proceedings of the 2017 ACM
  Conference on Economics and Computation}, 369--386.

\bibitem[{Kannan et~al.(2018)Kannan, Morgenstern, Roth, Waggoner,
  \protect\BIBand{} Wu}]{kannan2018smoothed}
Kannan S, Morgenstern JH, Roth A, Waggoner B, Wu ZS (2018) A smoothed analysis
  of the greedy algorithm for the linear contextual bandit problem.
  \emph{Advances in Neural Information Processing Systems}, 2227--2236.

\bibitem[{Kelly et~al.(1998)Kelly, Maulloo, \protect\BIBand{}
  Tan}]{kelly1998rate}
Kelly FP, Maulloo AK, Tan DK (1998) Rate control for communication networks:
  shadow prices, proportional fairness and stability. \emph{Journal of the
  Operational Research society} 49(3):237--252.

\bibitem[{Kim et~al.(2011)Kim, Herbst, Wistuba, Lee, Blumenschein, Tsao,
  Stewart, Hicks, Erasmus, Gupta et~al.}]{kim2011battle}
Kim ES, Herbst RS, Wistuba II, Lee JJ, Blumenschein GR, Tsao A, Stewart DJ,
  Hicks ME, Erasmus J, Gupta S, et~al. (2011) The battle trial: personalizing
  therapy for lung cancer. \emph{Cancer discovery} 1(1):44--53.

\bibitem[{Kleinberg et~al.(2010)Kleinberg, Niculescu-Mizil, \protect\BIBand{}
  Sharma}]{kleinberg2010regret}
Kleinberg R, Niculescu-Mizil A, Sharma Y (2010) Regret bounds for sleeping
  experts and bandits. \emph{Machine learning} 80(2):245--272.

\bibitem[{Kremer et~al.(2014)Kremer, Mansour, \protect\BIBand{}
  Perry}]{kremer2014implementing}
Kremer I, Mansour Y, Perry M (2014) Implementing the “wisdom of the crowd”.
  \emph{Journal of Political Economy} 122(5):988--1012.

\bibitem[{Lai \protect\BIBand{} Robbins(1985)}]{lai1985asymptotically}
Lai TL, Robbins H (1985) Asymptotically efficient adaptive allocation rules.
  \emph{Advances in applied mathematics} 6(1):4--22.

\bibitem[{Lattimore \protect\BIBand{} Szepesvari(2017)}]{lattimore2017end}
Lattimore T, Szepesvari C (2017) The end of optimism? an asymptotic analysis of
  finite-armed linear bandits. \emph{Artificial Intelligence and Statistics},
  728--737 (PMLR).

\bibitem[{Liu et~al.(2017)Liu, Radanovic, Dimitrakakis, Mandal,
  \protect\BIBand{} Parkes}]{liu2017calibrated}
Liu Y, Radanovic G, Dimitrakakis C, Mandal D, Parkes DC (2017) Calibrated
  fairness in bandits. \emph{arXiv preprint arXiv:1707.01875} .

\bibitem[{Mansour et~al.(2015)Mansour, Slivkins, \protect\BIBand{}
  Syrgkanis}]{mansour2015bayesian}
Mansour Y, Slivkins A, Syrgkanis V (2015) Bayesian incentive-compatible bandit
  exploration. \emph{Proceedings of the Sixteenth ACM Conference on Economics
  and Computation}, 565--582.

\bibitem[{Mas-Colell et~al.(1995)Mas-Colell, Whinston, Green
  et~al.}]{mas1995microeconomic}
Mas-Colell A, Whinston MD, Green JR, et~al. (1995) \emph{Microeconomic theory},
  volume~1 (Oxford university press New York).

\bibitem[{Mo \protect\BIBand{} Walrand(2000)}]{mo2000fair}
Mo J, Walrand J (2000) Fair end-to-end window-based congestion control.
  \emph{IEEE/ACM Transactions on networking} 8(5):556--567.

\bibitem[{Nash(1950)}]{nash1950bargaining}
Nash JF (1950) The bargaining problem. \emph{Econometrica: Journal of the
  econometric society} 155--162.

\bibitem[{Nimri et~al.(2020)Nimri, Battelino, Laffel, Slover, Schatz,
  Weinzimer, Dovc, Danne, \protect\BIBand{} Phillip}]{nimri2020insulin}
Nimri R, Battelino T, Laffel LM, Slover RH, Schatz D, Weinzimer SA, Dovc K,
  Danne T, Phillip M (2020) Insulin dose optimization using an automated
  artificial intelligence-based decision support system in youths with type 1
  diabetes. \emph{Nature medicine} 26(9):1380--1384.

\bibitem[{Papanastasiou et~al.(2018)Papanastasiou, Bimpikis, \protect\BIBand{}
  Savva}]{papanastasiou2018crowdsourcing}
Papanastasiou Y, Bimpikis K, Savva N (2018) Crowdsourcing exploration.
  \emph{Management Science} 64(4):1727--1746.

\bibitem[{Patil et~al.(2020)Patil, Ghalme, Nair, \protect\BIBand{}
  Narahari}]{patil2020achieving}
Patil V, Ghalme G, Nair V, Narahari Y (2020) Achieving fairness in the
  stochastic multi-armed bandit problem. \emph{Proceedings of the AAAI
  Conference on Artificial Intelligence}, volume~34, 5379--5386.

\bibitem[{Raghavan et~al.(2018)Raghavan, Slivkins, Wortman, \protect\BIBand{}
  Wu}]{raghavan2018externalities}
Raghavan M, Slivkins A, Wortman JV, Wu ZS (2018) The externalities of
  exploration and how data diversity helps exploitation. \emph{Conference on
  Learning Theory}, 1724--1738 (PMLR).

\bibitem[{Rugo et~al.(2016)Rugo, Olopade, DeMichele, Yau, van’t Veer, Buxton,
  Hogarth, Hylton, Paoloni, Perlmutter et~al.}]{rugo2016adaptive}
Rugo HS, Olopade OI, DeMichele A, Yau C, van’t Veer LJ, Buxton MB, Hogarth M,
  Hylton NM, Paoloni M, Perlmutter J, et~al. (2016) Adaptive randomization of
  veliparib--carboplatin treatment in breast cancer. \emph{New England Journal
  of Medicine} 375(1):23--34.

\bibitem[{Sen \protect\BIBand{} Foster(1997)}]{sen1997economic}
Sen A, Foster JE (1997) \emph{On Economic Inequality} (Oxford university
  press).

\bibitem[{Takeuchi et~al.(2009)Takeuchi, McGinnis, Bourgeois, Barnes, Eriksson,
  Soranzo, Whittaker, Ranganath, Kumanduri, McLaren
  et~al.}]{takeuchi2009genome}
Takeuchi F, McGinnis R, Bourgeois S, Barnes C, Eriksson N, Soranzo N, Whittaker
  P, Ranganath V, Kumanduri V, McLaren W, et~al. (2009) A genome-wide
  association study confirms vkorc1, cyp2c9, and cyp4f2 as principal genetic
  determinants of warfarin dose. \emph{PLoS Genet} 5(3):e1000433.

\bibitem[{Van~Parys \protect\BIBand{} Golrezaei(2020)}]{van2020optimal}
Van~Parys B, Golrezaei N (2020) Optimal learning for structured bandits.
  \emph{Available at SSRN 3651397} .

\bibitem[{Whirl-Carrillo et~al.(2012)Whirl-Carrillo, McDonagh, Hebert, Gong,
  Sangkuhl, Thorn, Altman, \protect\BIBand{} Klein}]{whirl2012pharmacogenomics}
Whirl-Carrillo M, McDonagh EM, Hebert J, Gong L, Sangkuhl K, Thorn C, Altman
  RB, Klein TE (2012) Pharmacogenomics knowledge for personalized medicine.
  \emph{Clinical Pharmacology \& Therapeutics} 92(4):414--417.

\bibitem[{Wysowski et~al.(2007)Wysowski, Nourjah, \protect\BIBand{}
  Swartz}]{wysowski2007bleeding}
Wysowski DK, Nourjah P, Swartz L (2007) Bleeding complications with warfarin
  use: a prevalent adverse effect resulting in regulatory action.
  \emph{Archives of internal medicine} 167(13):1414--1419.

\bibitem[{Young(1995)}]{young1995equity}
Young HP (1995) \emph{Equity: in theory and practice} (Princeton University
  Press).

\end{thebibliography}

\clearpage
\begin{APPENDICES}{}

\section{Deferred Descriptions}

\subsection{Negative Externality Example from \texorpdfstring{\cite{raghavan2018externalities}}{Raghavan et al. (2018)}} \label{app:negative_externality}

\cite{raghavan2018externalities} provide an example of an instance where there exists a sub-population that is better off when UCB is run on that sub-population alone, compared to running UCB on the entire population.
The example they provide depends on the total time horizon $T$.
We claim that this does not occur when you fix an instance and consider asymptotic log-scaled regret, $\lim_{T \rightarrow \infty} \frac{\ER_T}{\log T}$.

Fix any time $T_0$, and consider the two-armed instance according to $T=T_0$ from Definition 1 of \cite{raghavan2018externalities}.
The population consists of three groups that depend on their starting location: A, B, and C.
The sub-population consisting of B and C is dubbed the ``minority'', while A is the ``majority''.
Note that only B has access to both arms and hence it is the only group that can ever incur regret.
Group B pulls the arm that has a higher UCB, defined as $\htheta_t(a) + \sqrt{\frac{\alpha \log T_0}{N_t(a)}}$ for some tuning parameter $\alpha > 0$.

We first summarize informally how the negative externality arises.
Because arms 1 and 2 are so close together, even after $O(T_0)$ time steps, which arm has a higher UCB is not dominated by the difference between their empirical means, but it is dominated the second term of the UCB: $\sqrt{\frac{\alpha \log T_0}{N_t(a)}}$, which is just a function of the number of pulls $N_t(a)$.
That is, group B essentially ends up pulling the arm that has fewer pulls.
Therefore, when only the minority exists, since C only pulls arm 2, arm 1 ends up having a higher UCB, and hence B ends up always pulling arm 1.
However, if the majority group exists, arm 1 always has more pulls than arm 2 since there are more people from A than C.
Then, B ends up essentially always pulling arm 2.
If arm 2 is the arm that has a lower true reward than arm 1, then regret is higher when the majority group exists --- therefore, the existence of the majority can have a ``negative externality'' on the minority.

However, if we fix this instance and let $T \rightarrow \infty$, then no matter which arms is better, from \cref{thm:klucb_opt_stochastic_arrivals}, the total log-scaled regret is 0 from running KL-UCB.
Moreover, when the majority does not exist, then the minority incurs non-zero log-scaled regret when $\theta_1 < \theta_2$.
Therefore, the presence of the majority can only help the minority.

This example from \cite{raghavan2018externalities} shows that the presence of the majority can negatively affect the minority in the early time steps (i.e. $t < T_0$).
In the asymptotic regime, such a negative externality corresponds to adding $o(\log T)$ regret, which is deemed insignificant in our setting.



\section{Proof Preliminaries} \label{sec:proof:prelimiaries}

\subsection{Notation} 
\label{subsec:proof:notation}

For all of the subsequent proofs, we assume that an instance $\cI$ is \ii{fixed}.
We often use big-O notation, which is with respect to $T \rightarrow \infty$, unless otherwise specified.
The big-O hides constants that may depend on any other parameter other than $T$, including the instance $\cI$.
In general, when we introduce a \ii{constant}, it may depend on any other parameters other than $T$ .
We are usually not concerned with the values of the constants as we are concerned with asymptotic results (though we do concern ourselves with constants in front of the leading term, usually $\log T$).
We sometimes re-use letters like $c$ for constants but they do not refer to the same value. 

The UCB of an arm is defined as:
\begin{align} \label{eq:define_UCB}
\UCB_t(a) = \max\{ q : N_t(a) \KL(\hmut(a), q) \leq \log t+ 3 \log \log t \}.
\end{align}

Let $\pulls(a)$ be the indicator for arm $a$ being pulled at time $t$, and let $\gpulls(a)$ be the indicator for when arm $a$ is pulled by group $g$.
We define the class of log-consistent policies:
\begin{definition} \label{def:log_consistent}
A policy $\pi$ for the grouped bandit problem is \ii{log-consistent} for if for any instance $(\theta, G, (p_g)_{g \in G}, (\cA_g)_{g \in G})$, 
for any group $g$,
\begin{align} \label{eq:reasonable_alg_assumption}
\bE\left[ \sum_{a \in \cA_{\text{sub}}(g)} N_{T}^g(a) \right]  = O(\log T).
\end{align}

That is, the expected number of times that group $g$ pulled a suboptimal arm by time $t$ is logarithmic in the number of arrivals of $g$.
\end{definition}

\subsection{Commonly Used Lemmas} 
\label{subsec:common_lemmas}

We state a few lemmas that are used several times for both \cref{thm:klucb_opt_stochastic_arrivals} and \cref{thm:main_fair_orig}.
These lemmas do not depend on the policy that is used.
The first result shows that the number of times that an arm's UCB is smaller than its true mean is small.
\begin{lemma} \label{commonlemma:allucb}
Let $\allucb_t = \{ \UCB_t(a) \geq \theta(a) \; \forall a \in \cA\}$ be the event that the UCB for every arm is valid at time $t$.
\begin{align*}
\sumt \Pr(\notallucb_t) = O(\log \log T).
\end{align*}
\end{lemma}
\begin{myproof}
For a fix arm $a$, $\sumt \Pr(\UCB_t(a) < \theta(a)) = O(\log \log T)$ follows from Theorem 10 of \cite{garivier2011kl}, plugging in $\delta = \log t + 3 \log \log t$ as is done in the proof of Theorem 2 of \cite{garivier2011kl}.
The result follows from a union bound over all actions $a \in \cA$.
\end{myproof}

The second lemma states a relationship between the radius of the UCB of an arm and the number of pulls of the arm.
\begin{lemma} \label{commonlemma:ucb_radius_logarithmic_pulls}
Let $0 < \alpha < \beta < 1$. 
There exists a constant $c > 0$ such that if $\htheta_t(a) \leq \alpha$ and $\UCB_t(a) \geq \beta$, then $N_t(a) < c \log t$.
\end{lemma}
\begin{myproof}
Suppose $\htheta_t(a) \leq \alpha$ and $\UCB_t(a) \geq \beta$.
Then, $\KL(\htheta_t(a), \UCB_t(a)) \geq \KL(\alpha, \beta)$.
Let $c = \frac{4}{\KL(\alpha, \beta)}$.
By definition of the UCB \eqref{eq:define_UCB}, $N_t(a) \leq \frac{\log t + 3 \log \log t}{\KL(\htheta_t(a), \UCB_t(a))} \leq c \log t$.
\end{myproof}

This result essentially states that if the radius of the UCB of an arm is larger than a constant, then the number of pulls of the arm is at most $O(\log t)$; this result follows simply from the definition of the UCB \eqref{eq:define_UCB}.
The next result states that if an arm $a$ is pulled, then its empirical mean will be close to its true mean.
\begin{lemma} \label{commonlemma:empirical_mean_close_when_pulled}
For any group $g$ and arm $a \in \cAg$, if $L < \theta(a) < U$,
\begin{align*} 
\sumt \Pr(\pulls(a), \htheta_t(a) \notin [L, U]) = O(1).
\end{align*}
where big-$O$ hides constants that may depend on the instance and $L, U$.
\end{lemma}

\begin{myproof}
Let $\hthetan(a)$ be the empirical mean after $n$ pulls of arm $a$.
Let $E_{t,n}$ be the event that the number of times arm 1 has been pulled before time $t$ is exactly $n$.
\begin{align*}
&\sum_{t=1}^T \Pr(\pulls(a), \htheta_t(a) \notin [L, U]) \\
=& \sum_{t =1}^{T} \sum_{n=1}^T \Pr(\pulls(a),\hthetan(a) \notin [L, U], E_{t, n}) \\
=& \sum_{n=1}^T \sum_{t =1}^{T} \Pr(\hthetan(a) \notin [L, U] \vg \pulls(a),E_{t, n}) \Pr(\pulls(a),E_{t, n})
\end{align*}
If $F_{t, n} = \{\pulls(a),E_{t, n}\}$, then for any $n$,
the events $F_{1, n}, \dots, F_{T, n}$ are disjoint.
Then, by the law of total probability, $\Pr(\hthetan(a) \notin [L, U]) \geq \sum_{t =1}^{T}  \Pr(\hthetan \notin [L, U] | F_{t, n}) \Pr(F_{t, n})$.
Therefore,
\begin{align*}
\sum_{t=1}^T \Pr(\pulls(a), \hthetat(a) \notin [L, U])
\leq  \sum_{n=1}^T  \Pr(\hthetan(a) \notin [L, U])
\leq  \sum_{n=1}^T  \exp(-\alpha n).
\end{align*}
for some $\alpha > 0$ since the rewards of arm $a$ are Bernoulli.
Therefore, $\sum_{t=1}^T \Pr(\pulls(a), \hthetat(a) \notin [L, U]) = O(1)$.
\end{myproof}


\section{Proof that KL-UCB is Regret Optimal}
\label{sec:proof_klucb_opt}

In this section, we prove that the KL-UCB policy is regret-optimal.
At each time step, $\pi^{\KLUCB}$ chooses the arm with the highest UCB, defined as \eqref{eq:define_UCB}, out of all arms available.
\begin{theorem} \label{thm:klucb_opt_stochastic_arrivals}
For all instances $\cI$ of the grouped $K$-armed bandit,
\begin{align} 
\liminf_{T \rightarrow \infty} \frac{\ER_T(\pi^{\KLUCB}, \cI)}{\log T}  \leq \sum_{a \in \cA_{\text{sub}}} \Delta^{\gmin(a)}(a) \Nopta.
\end{align}
\end{theorem}

The first step of the proof is to show that the \ii{number of pulls} of a suboptimal arm is optimal:
\begin{proposition} \label{theorem:number_pulls}
Let $a \in \cA_{\text{sub}}$ be a suboptimal arm.
KL-UCB satisfies
\begin{align*}
 \limsup_{T \rightarrow \infty} \frac{\bE\left[N_T(a)\right] }{\log T}
 \leq \Nopta.
\end{align*}
\end{proposition}
This result can be shown using the existing analysis of KL-UCB from \cite{garivier2011kl}.
The next step is to analyze how these pulls are distributed across groups.
In particular, we need to show that a group never pulls a suboptimal arm $a$ if $g \notin \gmin(a)$.
This is the result of the next theorem:
\begin{proposition} \label{thm:main_theorem}
Let $a \in \cA$. Let $g \in G_a$, $g \notin \gmin(a)$ be a group that has access to the arm but is not the group that has the smallest optimal out of $G_a$.
Then, KL-UCB satisfies
\begin{align*}
 \bE\left[N_T^g(a) \right] = O(\log \log T),
\end{align*}
where the big-$O$ hides constants that depend on the instance.
\end{proposition}

This result implies that for any arm $a$, the regret incurred by group $g \notin \gmin(a)$ pulling the arm is $o(\log T)$, and is equal to 0 when scaled by $\log T$.
\cref{thm:klucb_opt_stochastic_arrivals} then follows from combining \cref{theorem:number_pulls} and \cref{thm:main_theorem}.

In this section, we prove \cref{thm:main_theorem}.
Let $a \in \cA$ and let $A \in \gmin(a)$ be a group that has access to that arm with the smallest OPT.
Let group $B \notin \gmin(a)$ be another group that has access to arm $a$.
Let $\thetaA, \thetaB$ be the optimal arms for group A and B respectively. 
We use $\thetaA, \thetaB$ to refer to both the arm and the arm means.
Our goal is to show $\bE\left[N_T^B(a) \right] = O(\log \log T)$.
\begin{align*}
\bE\left[N_T^B(a) \right]
&= \sum_{t=1}^T \Pr(\pulls^B(a)) \\
& = \sum_{t=1}^T \Pr(\pulls^B(a),  \UCB_t(\thetaB) \geq \thetaB)
      + \sum_{t=1}^T \Pr(\pulls^B(a), \UCB_t(\thetaB) < \thetaB).
\end{align*}

The second sum can be bounded by \cref{commonlemma:allucb}, since 
$ \sumt \Pr(\pulls^B(a), \UCB_t(\thetaB) < \thetaB) \leq \sumt \Pr(\notallucb_t) = O(\log \log T)$.
Therefore, our goal is to show 
\begin{align} \label{eq:klopt_main_eq}
\sumt \Pr(\pulls^B(a),  \UCB_t(\thetaB) \geq \thetaB) = O(\log \log T).
\end{align}

We state a slightly more general result that implies \eqref{eq:klopt_main_eq}.
\begin{lemma} \label{lemma:main_epoch}
Suppose we run any log-consistent policy $\pi$.
Let $r > 0$ be fixed.
For any $a \in \cA$,
\begin{align*}
\sumt \Pr(\pulls(a), \UCB_t(a) \geq \OPT(\gmin(a)) + r) = O(\log \log T),
\end{align*}
where the constant in the big-$O$ may depend on the instance and $r$.
\end{lemma}
The rest of this section proves \cref{lemma:main_epoch}.

\subsection{Probabilistic Lower Bound of $N_t(a)$ for Grouped Bandit} \label{sec:lower_bound_props}

One of the main tools used in the proof of \cref{lemma:main_epoch} is a high probability lower bound on the number of pulls of a suboptimal arm.
Let $\arr = \sum_{s=1}^g \bI(g_t = g)$ be the number of arrivals of group $g$ by time $t$.
Let $W^g_t = \{\arr \geq \frac{p_g t}{2}\}$ be the event that the number of arrivals of group $g$ is at least half of the expected value.
We condition on the event $W^g_t$ to ensure that a group has arrived a sufficient number of times.
\begin{proposition}  \label{prop:lower_bound_stochastic}
Let $g$ be a group, and let $a \in \Asubg$ be a suboptimal arm for group $g$.
Fix $\eps \in (0, 1)$.
Suppose we run a log-consistent policy as defined in \cref{def:log_consistent}.
Then,
\begin{align} \label{eq:lb_prob_rate}
\Pr\left(N_{t}(a) \leq \frac{(1-\eps)\log t}{\KL(\theta(a), \OPT(g))}  \;\bigg|\; W^g_t  \right) = O\left(\frac{1}{\log t}\right),
\end{align}
where the big-$O$ notation is with respect to $t \rightarrow \infty$.
\end{proposition}
The proof of this result can be found in \cref{sec:proof_lower_bound_stochastic}.
For an arm $a \notin \cA_{\text{sub}}$, we have the following stronger result:
\begin{proposition}  \label{prop:lower_bound_opt_stochastic}
Let $a$ be an arm that is optimal for some group $g$.
Suppose we run a log-consistent policy.
Then, for any $b > 0$,
\begin{align*}
\Pr\left(N_t(a) \leq b \log t \;\big|\; W^g_t\right) = O\left(\frac{1}{\log t}\right),
\end{align*}
where the big-$O$ notation is with respect to $t \rightarrow \infty$ and hide constants that depend on both $b$ and the instance.
\end{proposition}

\begin{proposition} \label{prop:lb_group_num_pulls}
Let $\cI$ be a grouped $K$-armed bandit instance, and let $\pi \in \conspolicies$.
Let $g$ be a group, and let $a \in \Asubg$ be a suboptimal arm for group $g$.
Then,
\begin{align*}
\liminfT \frac{\bE[N_T(a)]}{\log T} \geq \Ngroupa.
\end{align*}
\end{proposition}

\subsection{Proof of \cref{lemma:main_epoch}} \label{sec:main_lemma_proof}
\textbf{Outline:} 
Let $A \in \gmin(a)$ be a group that has the smallest optimal out of all arms with access to $a$.
The main idea of this lemma is that group A does not ``allow'' the UCB of arm $a$ to grow as large as $\OPT(A) + r$, as group A would pull arm $a$ once the UCB is above $\OPT(A)$.
\cref{prop:lower_bound_stochastic} implies that $\UCB_t(a)$ is not larger than $\OPT(A)$ with high probability.
If this occurs at time $t$, since the radius of the UCB grows slowly (logarithmically), the earliest time that the UCB can grow to $\OPT(A) +r$ is $t^\gamma$, for some $\gamma > 1$.
We divide the time steps into epochs, where if epoch $k$ starts at time $s_k$, it ends at $s_k^\gamma$.
This exponential structure gives us $O(\log \log T)$ epochs in total, and we show that the expected number of times that $\UCB_t(a) > \OPT(A) + r$ during one epoch is $O(1)$.

\noindent \textbf{Proof:}
We denote by $\theta_a$ the true mean reward of arm $a$ and by $\htheta_t$ the empirical mean reward of $a$ at the start of time $t$. 
Let $U = \OPT(\gmin(a)) + r$.
Let $A \in \gmin(a)$, and let $\thetaA = \OPT(A)$.
If $a \notin \Asub$, then let $\thetaA = \OPT(A) + r/2$.
Let $b > 0$ such that $\frac{\KL(\theta_a, U)}{\KL(\theta_a, \thetaA)}= 1+b$.
Define $\theta_u \in [\theta_a, \thetaA]$ such that $\frac{\KL(\theta_u, U)}{\KL(\theta_a, \thetaA)}= 1+\frac{b}{2}$.
We have $\theta_a < \theta_u < \thetaA  < U$.
Define $\gamma \triangleq 1+\frac{b}{4}$.
Let $\eps > 0$ such that $\frac{1-\eps}{1+\eps}\cdot \frac{\KL(\theta_u, U)}{\KL(\theta_a, \thetaA)} = \gamma$.

By \cref{commonlemma:empirical_mean_close_when_pulled}, 
$\sumt \Pr(\pulls(a), \htheta_t(a) > \theta_u) = O(1)$.
Therefore, we can assume $\htheta_t(a) \leq \theta_u$.
Denote the event of interest by $E_t = \{\pulls(a), \UCB_t(a) \geq \thetaA + r, \htheta_t(a) \leq \theta_u\}$.
Our goal is to show $\sumt \Pr(E_t) = O(\log \log T)$.

Divide the time interval $T$ into $K = O(\log \log T)$ epochs.
Let epoch $k$ start at $s_k \triangleq \left\lceil 2^{\gamma^k}  \right\rceil$ for $k \geq 0$. Let $\cT_k = \{s_k, s_k+1, \dots, s_{k+1}-1\}$ be the time steps in epoch $k$.
This epoch structure satisfies the following properties:
\begin{enumerate}
    \item The total number of epochs is $O(\log \log T)$.
    \item $\frac{\log s_{k+1}}{\log s_k} = \gamma$ for all $k \geq 0$.
\end{enumerate}

We will treat each epoch separately.
Fix an epoch $k$.
Our goal is to bound $ \bE\left[\sum_{t \in \cT_k} \bI(E_t) \right] $.
\cref{commonlemma:ucb_radius_logarithmic_pulls} implies that there exists a constant $c > 0$ such that if $E_t$ occurs, it must be that $N_t(a) < c \log t$.
Hence, 
\begin{align*}
\sum_{t \in \cT_k} \bI(E_t) \leq c \log s_{k+1}.
\end{align*}

Define the event $G_t = \left\{N_{t}(a) \geq (1-\eps)\frac{\log t}{\KL(\mu, \thetaA)}\right\}$.
The following claim says that if $G_{s_k}$ is true, then $E_t$ never happens during that epoch. 
\begin{claim} \label{claim:good_event_epoch}
Suppose $G_{s_k}$ is true.
Let $t_0$ be such that if $t \geq t_0$, $\log \log t \leq \eps \log t$.
Then, if $s_k \geq t_0$,$\sum_{t = s_k}^{s_{k+1}} \bI(E_t) = 0$.
\end{claim}
This result follows from the fact that the event $G_{s_k}$ implies that the radius of the UCB is ``small'' at time $s_k$, 
and the epoch is defined so that the radius will not grow large enough that $E_t$ can occur during epoch $k$.
Therefore, we have the following:
\begin{align*} 
\bE\left[\sum_{t \in \cT_k} \bI(E_t) \right]
= \bE\left[\sum_{t \in \cT_k} \bI(E_t) \bigg| \bar{G}_{s_k}  \right]
\Pr\left(\bar{G}_{s_k} \right) 
\leq c \log s_{k+1}\Pr\left(\bar{G}_{s_k} \right).
\end{align*}

We can bound $\Pr\left(\bar{G}_{s_k} \right)$ using the probabilistic lower bound of \cref{prop:lower_bound_stochastic}.
\begin{claim} \label{claim:prob_g_h}
$\Pr\left( \bar{G}_{s_k} \right) \leq  O\left(\frac{1}{\log s_{k}}\right)$.
\end{claim}

Then, property 2 of the epoch structure implies 
$\bE\left[\sum_{t \in \cT_k} \bI(E_t) \right] = O(1)$.
Since the number of epochs is $O(\log \log T)$,
\begin{align*}
 \bE\left[\sum_{t = 1}^{T} \bI(E_t)\right]
  \leq \sum_{k=1}^K \bE\left[\sum_{t \in \cT_k} \bI(E_t)\right]
  =O(\log \log T),
\end{align*}
as desired.

\subsection{Proof of Claims} \label{sec:proof_of_claims}

\begin{myproof}[Proof of \cref{claim:good_event_epoch}]
Let $t = s_k > t_0$ and let $t' \geq t$ such that $E_{t'}$ is true.
By definition of KL-UCB,
\begin{align*}
N_{t'}(a) \leq \frac{\log {t'} + 3 \log {t'}}{\KL(\htheta_{t'}, \UCB_{t'}(\theta))}.
\end{align*}
Since $E_{t'}$ implies $\UCB_{t'}(a) > \thetaB$ and $\htheta_{t'} \leq \theta_u$, we have $N_{t'}(a) \leq \frac{\log {t'} + 3 \log {t'}}{\KL(\theta_u, \thetaB)}$.
Since $G_{s_k}$ is true, $N_{t'}(a) \geq (1-\eps)\frac{\log s_k}{\KL(\theta_a, \thetaA)}$.
Therefore, it must be that
\begin{align}
&(1-\eps)\frac{\log s_k}{\KL(\theta_a, \thetaA)}
\leq \frac{\log t' + 3 \log \log t'}{\KL(\theta_u, \thetaB)}
\leq \frac{(1+\eps)\log t'}{\KL(\theta_u, \thetaB)} \nonumber \\
\Rightarrow\;&\frac{1-\eps}{1+\eps}\cdot \frac{\KL(\theta_u, \thetaB)}{\KL(\theta_a, \thetaA)} \log s_k \leq  \log t' \nonumber \\
\Rightarrow \;& t' \geq s_k^\gamma. \nonumber
\end{align}
This implies that $t'$ is not in epoch $k$.
\end{myproof}

\begin{myproof}[Proof of \cref{claim:prob_g_h}]

For group $g = A$,
\cref{prop:lower_bound_stochastic} (or \cref{prop:lower_bound_opt_stochastic} if $a \notin \Asub$) states that
\begin{align*}
\Pr\left(\bar{G}_{s_k} \vg R_{s_k}^g \right) = O\left(\frac{1}{\log s_k} \right).
\end{align*}
(We show in \cref{app:sec:klucb_log_consistent} that KL-UCB is log-consistent.)

Now we need to bound $\Pr(\bar{R}_{s_k}^g) = \Pr\left( M_{s_k}(A) \leq \frac{p_A s_k}{2}\right)$.
Note that $M_s(A) = \sum_{t=1}^s Z_i^A$, where $Z_t^A \overset{\text{iid}}{\sim} \text{Bern}(p_A)$.
By Hoeffding's inequality,
\begin{align*}
\Pr\left(M_{s_{k}}(A) \leq \frac{p_A s_k}{2}\right) < \exp\left(-\frac{1}{2} p_A^2 s_{k}\right).
\end{align*}

Combining, we have
\begin{align*}
\Pr(\bar{G}_k) 
&\leq \Pr(\bar{R}_k) + \Pr(\bar{G}_k \vv R_k) \leq O\left(\frac{1}{\log s_k}\right).
\end{align*}

\end{myproof}

\section{Deferred Proofs for \cref{thm:klucb_opt_stochastic_arrivals}}

For any $\eps > 0$, let
\begin{align*}
K_{\eps}^g(x) =\left\lceil \frac{1+\eps}{\KL(\theta_a, \OPT(g))} \left( \log x + 3 \log \log x \right) \right\rceil.
\end{align*}
To show both \cref{theorem:number_pulls} and the fact that KL-UCB is log-consistent, we make use of the following lemma.

\begin{lemma} \label{lemma:nopullsafterK}
Let $a \in \cA$. Let $g \in G_a$ be a group in which $a$ is suboptimal.
For any $\eps > 0$,
\begin{align} \label{eq:lemma_nopullsafterKT}
\bE\left[ \sum_{t=1}^{T} \bI(\gpulls(a), N_t(a) \geq K_{\eps}^g(T)) \right] = O(\log \log T).
\end{align} 
\end{lemma}
\begin{myproof}
Let $\eps > 0$.
Recall that $\ssA_g$ is the optimal arm for group $g$, and $\OPT(g)$ is the mean reward of $\ssA_g$.
\begin{align*}
&\bE\left[ \sum_{t=1}^{T} \bI(\gpulls(a), N_t(a) \geq K_{\eps}^g(T)) \right] \\
&= \bE\left[ \sum_{t=1}^{T} \bI(\gpulls(a), N_t(a) \geq K_{\eps}^g(T), \UCB_t(\ssA_g) \geq \OPT(g)) \right]
+ \bE\left[ \sum_{t=1}^{T} \bI(\gpulls(a), \UCB_t(\ssA_g) < \OPT(g)) \right]
\end{align*}

The second term is $O(\log \log T)$ from \cref{commonlemma:allucb}.
We will show that the first term is $O(1)$.
Let $\hmu_s(a)$ be the empirical mean of $a$ after $s$ pulls.
Consider the event $\{A_t =a, g_t=g, N_t(a) =s, \UCB_t(\ssA_g) \geq \OPT(g)\}$, where $s \geq K_n$.
Suppose this is true at time $t$.
Then, it must be that $\UCB_t(a) \geq \OPT(g)$.
For this to happen, by definition of KL-UCB, it must be that
\begin{align} \label{eq:klucb_m_condition}
s\KL(\htheta_s(a), \OPT(g)) \leq \log t + 3 \log \log t.
\end{align}
Since $s \geq K_{\eps}^g(T)$ and $t \leq T$, we must have
\begin{align} \label{eq:kl_r_condition}
\KL(\htheta_s(a), \OPT(g)) \leq \frac{ \log T + 3 \log \log T}{K_{\eps}^g(T)} = \frac{\KL(\theta_a, \OPT(g))}{1+\eps}.
\end{align}

Let $r > \theta_a$ such that $\KL(r, \OPT(g)) = \frac{\KL(\theta_a, \OPT(g))}{1+\eps}$.
Then, for \eqref{eq:kl_r_condition} to occur, it must be that $\hmu_s(a) \geq r$.
Then, we have
\begin{align*}
& \bE\left[ \sum_{t=1}^{T} \bI(\gpulls(a), N_t(a) \geq K_{\eps}^g(n), \UCB_t(\ssA_g) \geq \OPT(g)) \right] \\
=&  \bE\left[ \sum_{t=1}^{T}  \sum_{s=K_n}^{\infty}  \bI(\gpulls(a), N_t(a)=s, \UCB_t(\ssA_g) \geq\OPT(g)) \right]\\
\leq&  \bE\left[ \sum_{t=1}^{T}  \sum_{s=K_n}^{\infty}  \bI(\gpulls(a), N_t(a)=s, \hmu_s(a) \geq r ) \right] \\
=&  \bE\left[  \sum_{s=K_n}^{\infty} \bI(\hmu_s(a) \geq r ) \sum_{t=1}^{T}  \bI(\gpulls(a), N_t(a)=s ) \right]\\
\leq&  \sum_{s=K_n}^{\infty} \Pr(\hmu_s(a) \geq r).
\end{align*}

Since $r > \mu(a)$, there exists a constant $C_3 > 0$ that depends on $\eps$ and $r$ such that $\Pr(\mu_s(a) \geq r) \leq \exp(-s C_3).$
Therefore, $\sum_{s=K_n}^{\infty} \Pr(\hmu_s(a) \geq r) = O(1)$ and we are done.

\end{myproof}

\subsection{Proof that KL-UCB is log-consistent} \label{app:sec:klucb_log_consistent}

This basically follows from \cref{lemma:nopullsafterK}.
Let $\eps = 1/2$.
Fix a group $g$, and let $a$ be a suboptimal arm for $g$.
\begin{align*}
\bE[N^g_{T}(a)]
&= \bE\left[\sum_{t=1}^{T} \bI(\gpulls(a)) \right] \\
&\leq K_{\eps}^g(T) + \bE\left[\sum_{t=1}^{t_{g(n)}} \bI(\gpulls(a), N_t(a) \geq K_{\eps}^g(T)) \right] \\
&= K_{\eps}^g(T) + \log \log(T).
\end{align*}
We are done since $K_{\eps}^g(T) = O(\log T)$.

\subsection{Proof of \cref{theorem:number_pulls}} \label{sec:proof:theorem:number_pulls}

Let $a \in \cA_{\text{sub}}$ be a suboptimal arm.
Let $\eps > 0$.
Let
\begin{align*}
K_T = \max_{g \in G_a} K_{\eps}^g(T).
\end{align*}
Clearly, the maximum is attained in the group $g$ with the smallest $\OPT(g)$, so.
\begin{align*}
K_T =\left\lceil \frac{1+\eps}{\KL(\theta_a, \OPT(\gmin(a)))} \left( \log T + 3 \log \log T \right) \right\rceil.
\end{align*}

\begin{align*}
\bE[N_T(a)]
&= \bE\left[\sum_{t=1}^T \bI(A_t = a) \right] \\
&\leq K_T + \bE\left[\sum_{t=1}^T \bI(A_t = a, N_t(a) \geq K_T) \right] \\
&\leq K_T + \sum_{g \in G_a} \bE\left[\sum_{t=1}^T \bI(\gpulls(a), N_t(a) \geq K_T) \right] \\
&\leq K_T + \sum_{g \in G_a} O(\log \log T).
\end{align*}
where the last inequality follows from \cref{eq:lemma_nopullsafterKT} of \cref{lemma:nopullsafterK}.
Since this holds for any $\eps > 0$, the desired result holds.

\subsection{Proof of Propositions~\ref{prop:lower_bound_stochastic}, \ref{prop:lower_bound_opt_stochastic}, \ref{prop:lb_group_num_pulls}} \label{sec:proof_lower_bound_stochastic}
Let $g$ be a group, and let $j$ be a suboptimal arm for group $g$; i.e. $\theta_j < \OPT(g)$.
Fix $\eps > 0$.
We assume that the event $W_t^g = \{\arr \geq \frac{p_g t}{2}\}$ holds.
Fix $\delta > 0$ such that $\frac{1-\delta}{1+\delta} = 1 - \eps$. Let $a = \delta/2$.
We construct another instance $\gamma$ where arm $j$ is replace with $\lambda$ so that arm $j$ is the optimal arm for $g$ in the same manner as the Lai-Robbins proof.
Specifically, $\lambda > \theta_j$ such that
\begin{align*}
\KL(\theta_j, \lambda) = (1+\delta) \KL(\theta_j, \OPT(g)).
\end{align*}

Our goal is to bound the probability of event $\left\{N_{t}(j) \leq \frac{(1-\delta)\log t}{\KL(\theta_j, \lambda)}\right\}$, which we split into two events:
\begin{align*}
C_t &= \left\{N_{t}(j) \leq \frac{(1-\delta)\log t}{\KL(\theta_j, \lambda)}, L_{N_{t}(j)} \leq (1-a) \log t \right\}, \\
E_t &= \left\{N_{t}(j) \leq \frac{(1-\delta)\log t}{\KL(\theta_j, \lambda)}, L_{N_{t}(j)} > (1-a) \log t\right\},
\end{align*}
where $L_m = \sum_{i =1}^m \log \left( \frac{f(Y_i ; \theta_j)}{f(Y_i ; \lambda)}\right)$.

Assumption \eqref{eq:reasonable_alg_assumption}, 
there exists a constant $c$ such that 
if $t$ is large enough that $\Pr(W_t^g) \geq 1/2$,
\begin{align*}
\bE_{\gamma}\left[\sum_{a \in \cA_{\text{sub}}} N_t^g(a) \vgg W_t^g\right] \leq  c\log t.
\end{align*}
Since $j$ is the unique optimal arm under $\gamma$,
\begin{align*}
\bE_{\gamma}\left[\arr - N^g_{t}(j) \vgg W_t^g\right] \leq  c\log t.
\end{align*}
Using Markov's inequality and using the fact that $\arr \geq \frac{p_g t}{2}$, we get
\begin{align*}
\Prgamma\left( N^g_{t}(j) \leq \frac{(1-\delta) \log t}{\KL(\theta_j, \lambda)} \vgg W_t^g \right) 
=& \Prgamma\left(\arr - N^g_{t}(j) \geq \arr - \frac{(1-\delta) \log t}{\KL(\theta_j, \lambda)} \vgg W_t^g \right)  \\
\leq& \Prgamma\left(\arr - N^g_{t}(j) \geq \frac{p_g t}{2} - \frac{(1-\delta) \log t}{\KL(\theta_j, \lambda)} \vgg W_t^g \right) \\
\leq& \frac{\bE\left[\arr - N^g_{t}(j) \vg W_t^g \right]}{\frac{p_g t}{2} - \frac{(1-\delta) \log t}{\KL(\theta_j, \lambda)}} \\
=& O\left(\frac{\log t}{t}\right).
\end{align*}

\noindent
\textbf{Bounding $\Pr(C_t\vv W_t^g )$: }
Following through with the same steps as the original proof, we can replace (2.7) with
\begin{align*}
\Prtheta(C_t\vv W_t^g) \leq t^{1-a} \Prgamma(C_t\vv W_t^g)
\leq t^{1-a} O\left(\frac{\log t}{t}\right)
= O\left(\frac{\log t}{t^a}\right).
\end{align*}

\noindent
\textbf{Bounding $\Pr(E_t \vv W_t^g)$: }
Next, we need to show a probabilistic result in lieu of (2.8) of \cite{lai1985asymptotically}.
Let $m = \frac{(1-\delta) \log t}{\KL(\theta_j, \lambda)}$ and let $\alpha> 0$ such that $(1+\alpha) = \frac{1-a}{1-\delta}$.
We need to upper bound
\begin{align*}
\Prtheta\left(\max_{j \leq m} L_j > (1-a) \log t \right)
&= \Prtheta\left(\max_{j \leq m} L_j > (1+\alpha) \KL(\theta_j, \lambda) m \right) \\
&\leq \Prtheta\left(\max_{j \leq m}\{L_j - j\KL(\theta_j, \lambda)\}  > \alpha \KL(\theta_j, \lambda) m \right).
\end{align*}

Let $Z_i = \log \left( \frac{f(Y_i ; \theta_j)}{f(Y_i ; \lambda)}\right) - \KL(\theta_j, \lambda)$. We have $\bE[Z_i] = 0$.
Let $\text{Var}(Z_i) = \sigma^2$.
Then, by Kolmogorov's inequality, we have
\begin{align*}
\Prtheta\left(\max_{j \leq m}\sum_{i=1}^j Z_i  > \alpha \KL(\theta_j, \lambda) m \right)
&\leq \frac{1}{\alpha^2 \KL(\theta_j, \lambda)^2 m^2 } \text{Var}\left(\sum_{i=1}^m Z_i \right) \\
& = \frac{\sigma^2}{\alpha^2 \KL(\theta_j, \lambda)^2 m } \\
& = O\left(\frac{1}{\log t}\right),
\end{align*}
since $m = \Theta(\log t)$.

\noindent
\textbf{Combine: }
Combining, we have
\begin{align*}
\Prtheta\left(N_{t}(j) \leq \frac{(1-\delta)\log n}{\KL(\theta_j, \lambda)}\vgg W_t^g \right)
&= \Prtheta(C_n\vg W_t^g) + \Prtheta(E_n\vg W_t^g) \\
&= O\left(\frac{\log t}{t^a}\right) + O\left(\frac{1}{\log t}\right).
\end{align*}

Since $\KL(\theta_j, \lambda) \leq (1+\delta) \KL(\theta_j, \OPT(g))$ and $\frac{1-\delta}{1+\delta} = 1 - \eps$, we have
\begin{align} 
\Prtheta\left(N_t(j) \leq \frac{(1-\eps)\log t}{\KL(\theta_j, \OPT(g))}\vgg W_t^g\right)
 \leq O\left(\frac{1}{\log t}\right)
\end{align}
as desired.

\begin{myproof}[Proof of \cref{prop:lower_bound_opt_stochastic}]
The proof of this result follows the same steps as \cref{prop:lower_bound_stochastic}.
Let $\eps = 1/2$ and let $\theta^* > \theta_j$ so that $\frac{1-\eps}{\KL(\theta_j, \theta^*)} = b$.
In the proof of  \cref{prop:lower_bound_stochastic}, replace $\OPT(g)$ with $\theta^*$.
Then, the same proof goes through and we get $\Pr\left(N_t(j) \leq b \log n \;\big|\; R^g_t\right) = O\left(\frac{1}{\log t}\right)$.
\end{myproof}

\begin{myproof}[Proof of \cref{prop:lb_group_num_pulls}]
Let $g$ be a group, and let $a \in \Asubg$ be a suboptimal arm for group $g$.
Let $\pi \in \conspolicies$.
The following can be shown using the same proof as \cref{prop:lower_bound_stochastic}:
\begin{align*}
\liminfT \frac{\bE[N_T(a) \;|\;  W^g_T]}{\log T} \geq  \Ngroupa.
\end{align*}
The only change in the proof is to use the definition of a consistent policy instead of a log-consistent policy, which proves that the LHS of \eqref{eq:lb_prob_rate} goes to 0 as $t$ increases (rather than an explicit rate).

Let $\eps > 0$.
Let $T'$ be large enough that for all $T \geq T'$, $\Pr(W^g_T) > 1-\eps$, and that $\frac{\bE[N_T(a) \;|\;  W^g_T]}{\log T} \geq  \Ngroupa + \eps$.
Then, for any $T \geq T'$,
\begin{align*}
\frac{\bE[N_T(a)]}{\log T}
&\geq \frac{\bE[ N_T(a) \;|\;  W^g_T ]\Pr(W^g_T)}{\log T} \\
&\geq  \frac{\bE[ N_T(a) \;|\;  W^g_T ] (1-1/T)}{\log T} \\
&\geq  \Ngroupa (1-\eps)^2.
\end{align*}
The result follows since $\eps > 0$ was arbitrary.
\end{myproof}

\section{Proof of \texorpdfstring{\cref{thm:main_fair_orig}}{Theorem \ref{thm:main_fair_orig}}} \label{sec:app:pfucb_fair_proof}

In this section, we prove that $\PFUCB$ is the Nash solution.
\cref{thm:main_fair_orig} is a corollary of the following theorem that characterizes the group regret under $\PFUCB$.
\begin{theorem} \label{thm:main_fair}
Let $\cI$ be an instance of grouped $K$-armed bandits.
Let $q$ be the optimal solution to \eqref{eq:fair_opt_prob} with the smallest Euclidean norm.
Then, for all groups $g \in \cG$,
 \begin{align*}
 \lim_{T \rightarrow \infty}\frac{\ER^g_T(\pi^{\PFUCB}, \cI)}{\log T} = \sum_{a \in \cA^g} \Delta^g(a) q^g(a) \Nopta.
 \end{align*}
\end{theorem}

First, we provide the proof of \cref{thm:main_fair} using the results of Propositions~\ref{prop:greedy_not_opt}, \ref{prop:eps_close_delta}, and \ref{prop:subopt_pulls_ht} (propositions stated in the proof sketch).
Then, we prove Propositions~\ref{prop:greedy_not_opt}, \ref{prop:subopt_pulls_ht}, and \ref{prop:total_pulls} in \cref{app:main_fair_props}.
Lastly, we prove \cref{prop:eps_close_delta} in \cref{app:proof:eps_close_delta}.

\subsection{Proof of \cref{thm:main_fair}}

We first state a result which states that the number of pulls of each arm is optimal.
\begin{proposition} \label{prop:total_pulls}
For any $a \in \Asub$, $\PFUCB$ satisfies
\begin{align} \label{eq:total_pulls_limit}
 \lim_{T \rightarrow \infty} \frac{\bE[N_T(a)]}{\log T}  = \Nopta.
\end{align}
\end{proposition}

\begin{myproof}[Proof of \cref{thm:main_fair}]
Fix a group $g$ and an arm $a \in \Asubg$.
Let $\eps > 0$. Let $\delta \in (0, \delta_0)$ according to \cref{prop:eps_close_delta}. Let $H_t = H_t(\delta)$.
\begin{align}
\bE[N^g_T(a)] 
&= \sum_{t=1}^T \Pr(\gpulls(a)) \nonumber \\ 
&= \sum_{t=1}^T (  \Pr(\gpulls(a), A^{\text{greedy}}_t(g) \neq a, H_t) \nonumber \\
& \quad 
+ \Pr(\gpulls(a), A^{\text{greedy}}_t(g) = a)
+ \Pr(\gpulls(a), A^{\text{greedy}}_t(g) \neq a, \bar{H}_t) )  \nonumber 
\\
&\leq \sum_{t=1}^T \Pr(\gpulls(a), A^{\text{greedy}}_t(g) \neq a, a \in \AKL, H_t) + O( \log \log T). \label{eq:g_pulls_decomp}
\end{align}
where the last step follows from \cref{prop:subopt_pulls_ht} and \cref{prop:greedy_not_opt}. 

First, assume that $a \notin \Asub$.
That is, there exists a group $g'$ such that $a$ is optimal for $g'$.
We claim that $\Pr(\gpulls(a) \vg a \in \AKL, H_t) = 0$.
Notice that when $H_t$ is true, $a$ is not the greedy arm for $g$, and moreover, $a \notin \hAsub$.
Therefore, from \eqref{eq:fair_opt_prob_karm}, $q^g(a) = 0$.
Since $a$ is also not the greedy arm for $g$, $g$ will not pull arm $a$ under PF-UCB.
Therefore, $\gpulls(a) = 0$ when $H_t$ is true.
This implies that if $a \notin \Asub$,
\begin{align} \label{eq:not_in_Asub}
\lim_{T \rightarrow \infty}\frac{\bE[N_T^g(a)]}{\log T} =  0.
\end{align}

Next, assume $a \in \Asub$.
By definition of the algorithm, if $\{\pulls^g(a), A^{\text{greedy}}_t(g) \neq a\}$ occurs, then $N_t^g(a) \leq \hq_t^g(a) N_t(a)$. If $H_t(\delta)$, then $\hq_t^g(a) \leq q_t^g(a) + \eps$.
Therefore, $\sumt \bI(\pulls^g(a), a \in \AKL, H_t(\delta)) \leq (q_t^g(a) + \eps) N_T(a)$.
Then, using \eqref{eq:g_pulls_decomp}, we can write
\begin{align*}
\limsupT \frac{\bE[N_T^g(a)] }{\log T}
&= 
\limsup_{T \rightarrow \infty}
\frac{\bE\left[\sumt \bI(\pulls^g(a), a \in \AKL, H_t(\delta))\right] + O(\log \log T)}{\log T} \\
& \leq   \limsup_{T \rightarrow \infty}  \frac{(q^g(a) + \eps)\bE[N_T(a)]}{\log T } \\
& \leq (q^g(a) + \eps) J(a),
\end{align*}
where the last inequality follows from \cref{prop:total_pulls}.
Since this holds for all $\eps > 0$,
\begin{align} \label{eq:limsup_NgTa}
\limsupT \frac{\bE[N^g_T(a)] }{\log T} \leq  q^g(a) J(a).
\end{align}

\cref{prop:total_pulls} implies that \eqref{eq:limsup_NgTa} must be an equality all $g$, since $\sum_g q^g(a) = 1$ for all $a$.
If this weren't the case, then $\limsupT \frac{\bE[N_T(a)] }{\log T}$ would be strictly less than $J(a)$, which would be a contradiction.

Moreover, we claim that \eqref{eq:limsup_NgTa} and \eqref{eq:total_pulls_limit} implies $\limT \frac{\bE[N^g_T(a)] }{\log T} =  q^g(a) J(a)$ for all $g$.
By contradiction, suppose there exists a $g' \in \cG$ such that 
$\liminfT \frac{\bE[N^{g'}_T(a)] }{\log T} =  q^{g'}(a) J(a) - \alpha$ for some $\alpha > 0$.
Then, \eqref{eq:total_pulls_limit} implies that $\limsupT \sum_{g \neq g'} \frac{\bE[N^{g'}_T(a)] }{\log T} \geq (1- q^{g'}(a)) J(a) + \alpha$, which is a contradiction.
Therefore, for every $g$,
\begin{align*} 
\limT \frac{\bE[N^g_T(a)] }{\log T} =  q^g(a) J(a).
\end{align*}

Combining with \eqref{eq:not_in_Asub} yields the desired result:
\begin{align} \label{eq:pf_desired}
\lim_{T \rightarrow \infty} \frac{\ER^g_T(\pi^{\text{PF-UCB}}, \cI)}{\log T}
= \lim_{T \rightarrow \infty} \frac{\sum_{a \in \cA} \Delta^g(a) \bE[N^g_T(a)] }{\log T} 
= \sum_{a \in \Asub} \Delta^g(a) q^g(a) \Nopta.
\end{align}
\end{myproof}

\subsection{Proof of Propositions~\ref{prop:greedy_not_opt}, \ref{prop:subopt_pulls_ht}, and \ref{prop:total_pulls}} \label{app:main_fair_props}

\begin{myproof}[Proof of \cref{prop:greedy_not_opt}]
Let $g \in \cG$ and let $a \in \Asubg$. We bound $\sumt \Pr(\gpulls(a), a = \Ag_t(g))$.
We can assume that the events $\htheta_t(a) \in [\theta(a) - \delta, \theta(a) + \delta]$ and $\allucb_t$ occur 
using \cref{commonlemma:empirical_mean_close_when_pulled} and \cref{commonlemma:allucb} respectively.
Since $a$ is the greedy arm, it must be that $\htheta_t(a') \leq \theta(a) + \delta$ for all $a' \in \cAg$.

Define the event 
\begin{align*}
R_t = \{ 
		\Ag_t(g) = a,
		\allucb_t, 
		\htheta_t(a) \leq \theta(a) + \delta,
		\htheta_t(a') \leq \theta(a) + \delta \; \forall a' \in \cAg
		\}.
\end{align*}
Our goal is to bound $\sumt \Pr(R_t)$.

For $R_t$ to occur, $\htheta_t(a') \leq \theta(a) + \delta$ (since $a$ is the greedy arm) and $\UCB_t(a') \geq \OPT(g)$ (since $\allucb_t$) for all $a' \in \Aoptg$.
By \cref{commonlemma:ucb_radius_logarithmic_pulls} there exists a constant $c > 0$ such that if $N_t(a') > c \log t$ for some $a' \in \Aoptg$, $R_t$ cannot happen.
Moreover, for every $a' \in \Aoptg$, $\Pr(N_t(a') < c \log t) < O \left( \frac{1}{\log t} \right)$ from \cref{prop:lower_bound_opt_stochastic}.

Divide the time period into epochs, where epoch $k$ starts at time $s_k = 2^{2^k}$.
Let $\cT_k$ be the time steps in epoch $k$.
Let $G_k = \{N_{s_k}(a) > 3c \log s_k\; \forall a \in \Aoptg\}$ be the event that all optimal arms were pulled at least $3c \log s_k$ times by the start of epoch $k$.
If $G_k$ occurs, since $s_k = \sqrt{s_{k+1}}$, $N_{s_{k+1}}(a) > \frac{3}{2} r\log s_{k+1} > r \log s_{k+1}$, and hence $R_t$ can never happen during epoch $k$.
Moreover, $\Pr(\bar{G}_k) = O \left( \frac{1}{\log s_k} \right)$ for any $k$.

Suppose we are in a ``bad epoch'', where $G_k$ does not occur.
We claim that $R_t$ can't occur more than $O(\log s_{k+1})$ times during epoch $k$.
For $R_t$ to occur, the arm $j$ with the highest UCB 
satisfies $\UCB_t(j) \geq \OPT(g)$ and $\htheta_t(j) \leq \theta(a) + \delta$.

\begin{claim} \label{claim:ucb_radius_logarithmic_stronger}
For any action $j \in \cAg$, $\sum_{t=1}^{s} \Pr(\AU_t(g) = j, \UCB_t(j) \geq \OPT(g), \htheta_t(j) \leq \theta(a) + \delta \vg \bar{G}_k) = O(\log s)$.
\end{claim}
Using \cref{claim:ucb_radius_logarithmic_stronger} and taking a union bound over all actions $j$ implies $\sum_{t \in \cT_k} \Pr(R_t \vg \bar{G}_k) = \sum_{t \in \cT_k} \sum_{j \in \cAg} \Pr(R_t, \AU_t(g) = j \vg \bar{G}_k ) = O(\log s_{k+1})$.
Since $\Pr(\bar{G}_k) = O \left( \frac{1}{\log s_k} \right)$, $\sum_{t \in \cT_k} \Pr(R_t) = O(1)$.
Since there are $O(\log \log T)$ epochs, $\sumt \Pr(R_t) = O(\log \log T)$.
\end{myproof}

\begin{myproof}[Proof of \cref{prop:subopt_pulls_ht}]
Let $H_t = H_t(\delta)$.
Fix a group $g$ and an arm $a \in \Asubg$.
For $g$ to pull $a$ when $A^{\text{greedy}}_t(g) \neq a$, it must be that $a \in \AKL$.

First, assume $a \notin \Asub$. Then, there exist groups $G \subseteq \cG$ in which $a$ is optimal.
If $a$ is the greedy arm for some $g' \in G$, 
then $a \notin \hAsub$, implying $a$ is not considered in the optimization problem $(\hat{P}_t)$.
In this case, group $g$ would never pull arm $a$.
Therefore, it must be that $a$ is not the greedy arm for all groups in $G$.
We show the following lemma, which proves the proposition for an arm $a \notin \Asub$.
\begin{lemma} \label{lemma:optarm_pulled_not_greedy}
Let $a \notin \Asub$, and let $G$ be the set of groups in which $a$ is optimal.
Then,
\begin{align*}
\sumt \Pr(\pulls(a), A^{\text{greedy}}_t(g) \neq a\; \forall g \in G, a \in \AKL) = O(\log \log T).
\end{align*}
\end{lemma}

Now assume $a \in \Asub$.
We assume that the events $\allucb_t$ and $\hmu_t(a) \in [\theta(a)-\delta, \theta(a) + \delta]$ hold using \cref{commonlemma:allucb} and \cref{commonlemma:empirical_mean_close_when_pulled}.
Since $a \in \AKL$ and $\allucb_t$, it must be that $\UCB_t(a) \geq \OPT(\gmin(a))$.
Let $E_t = \{\gpulls(a), \allucb_t, \hmu_t(a) \in [\theta(a)-\delta, \theta(a) + \delta], \UCB_t(a) \geq \OPT(\gmin(a)) \}$
Our goal is to show
\begin{align*}
\bE\left[\sumt \bI(E_t, \bar{H}_t)\right] = O(\log \log T).
\end{align*}

Divide the time interval into epochs, where epoch $k$ starts at time $s_k = 2^{2^k}$.
Let $K = O(\log \log T)$ be the total number of epochs.
Let $\cT_k$ be the time steps in epoch $k$.

Let $H_k = \cap_{t \in \cT_k} H_t$.
Clearly, if $H_k$ is true, then by definition, $\sum_{t \in \cT_k}\bI(E_t, \bar{H}_t)= 0$.
Therefore, we can write
\begin{align*}
\bE\left[\sumt \bI(E_t, \bar{H}_t)\right]
=\sumk \bE\left[\sum_{t \in \cT_k} \bI(E_t, \bar{H}_t)\right] 
= \sumk \left( \bE\left[ \sum_{t \in \cT_k} \bI(E_t, \bar{H}_t) \vgg \bar{H}_k \right] \Pr(\bar{H}_k) \right) 
\end{align*}
We bound the expectation and the probability separately.

\noindent
\textbf{1) Bounding $\bE\left[ \sum_{t \in \cT_k} \bI(E_t, \bar{H}_t) \vgg \bar{H}_k \right]$: }
If $E_t$ occurs at some time step $t$, $\UCB_t(a) \geq \OPT(\gmin(a))$ and $\htheta_t(a) \leq \theta(a) + \delta$.
By \cref{commonlemma:ucb_radius_logarithmic_pulls} it must be that $N_t(a) = O(\log t)$.
Clearly, $N_s(a) \geq \sum_{t=1}^s \bI(E_t)$, implying that $\sum_{t \in \cT_k} \bI(E_t) = O(\log s_{k+1})$.
Therefore, $\sum_{t \in \cT_k} \bI(E_t, \bar{H}_t) \leq \sum_{t=1}^{s_{k+1}} \bI(E_t) = O(\log s_{k+1})$

\noindent
\textbf{2) Bounding $\Pr(\bar{H}_k)$: }
For $a \in \Asub$ let $c_a = \frac{0.9}{\KL(\theta(a), \OPT(\gmin(a)))}$. 
For $a \notin \Asub$, let $c_a = 1$.
Let $F_k = \{\hmu_{s_k}(a) \in [\theta(a) - \delta/2, \theta(a) + \delta/2], N_{s_k}(a) \geq c_a \log s_k \; \forall a \in \cA\}$ be the event that at time $s_k$, all arms $a$ have been pulled $c_a \log s_k$ times and all arms are within an ``inner'' boundary (half as small as the boundary defined for $H_t$).
We bound $\Pr(\bar{H}_k)$ by conditioning on the event $F_k$.
Firstly, we bound $\Pr(\bar{F}_k)$ using the probabalistic lower bound of \cref{prop:lower_bound_stochastic}-\ref{prop:lower_bound_opt_stochastic}:
\begin{lemma} \label{lemma:F_k}
For any $k$, $\Pr(\bar{F}_k) = O\left(\frac{1}{\log s_k}\right)$.
\end{lemma}

Next, we show that if $F_k$ is true, then $H_k$ occurs with probability at least $1 - O\left(\frac{1}{\log s_k}\right)$.
\begin{lemma} \label{lemma:outside_boundary_start_in_boundary}
For any action $a$, $\Pr\left(\hmu_t(a) \notin [\theta(a) - \delta, \theta(a) + \delta] \text{ for some } t \in \cT_k\;|\; F_k \right) \leq O\left(\frac{1}{\log s_k}\right)$.
\end{lemma}

Therefore,
\begin{align*}
\Pr(\bar{H}_k) 
\leq  \Pr(\bar{F_k})+ \Pr(\bar{H}_k \vg F_k)  
= O\left(\frac{1}{\log s_k}\right).
\end{align*}

\noindent
\textbf{3) Combine:}
Combining, we have
\begin{align*}
\bE\left[\sumt \bI(E_t, \bar{H}_t)\right]
\leq& \sumk \left( O(\log s_{k+1}) O\left(\frac{1}{\log s_{k}}\right) \right) \\
\leq& \sumk O(1) \\
=&  O( \log \log T),
\end{align*}
where the last inequality follows due to the fact that $\frac{ \log s_{k+1}}{\log s_k} = 2$ for any $k$.
\end{myproof}

\begin{myproof}[Proof of \cref{prop:total_pulls}]
Let $a \in \Asub$.
We need to show $\limsupT \frac{\bE[N_T(a)]}{ \log T} \leq J(a)$, as the lower bound is implied by \eqref{eq:lower_bound_pulls}.
By \cref{prop:greedy_not_opt}, the number of times $a$ is pulled when $a$ is the greedy arm for some group $g$ is $O(\log \log T)$.
Therefore,
\begin{align*}
\bE[N_T(a)] = \sumt \Pr(\pulls(a), a \in \AKL) + O(\log \log T).
\end{align*}
The result follows from the fact that KL-UCB is optimal (same argument as \cref{theorem:number_pulls}).
\end{myproof}

\subsubsection{Deferred Proofs of Lemmas}

\begin{myproof}[Proof of \cref{claim:ucb_radius_logarithmic_stronger}]
Recall that $G_k = \{N_{s_k}(a) > 3c \log s_k\; \forall a \in \Aoptg \}$.
We will show $\sumt \Pr(\AU_t = j,\UCB_t(j) \geq \OPT(g), \htheta_t(j) \leq \theta(a) + \delta \vg \bar{G}_k) = O(\log \log T)$.
From \cref{commonlemma:ucb_radius_logarithmic_pulls}, there exists a constant $c'$ such that if $N_t(j) > c' \log T$ then, $\{\UCB_t(j) \geq \OPT(g), \htheta_t(j) \leq \theta(a) + \delta\}$ cannot occur.
\begin{align}
&\sum_{t\in \cT_k} \Pr(\AU_t(g) = j,\UCB_t(j) \geq \OPT(g), \htheta_t(j) \leq \theta(a) + \delta \vg \bar{G}_k)  \nonumber \\
=& \sum_{n=1}^{c' \log T}\sum_{t\in \cT_k} \Pr(\AU_t(g) = j,\UCB_t(j) \geq \OPT(g), \htheta_t(j) \leq \theta(a) + \delta, N_t(a) = n \vg \bar{G}_k) \nonumber \\
\leq& \sum_{n=1}^{c' \log T} \sum_{t\in \cT_k} \Pr(\AU_t(g) = j, N_t(a) = n \vg \bar{G}_k). \label{eq:ucb_radius_log_stronger}
\end{align}

Our goal is to show that $\sum_{t\in \cT_k} \Pr(\AU_t(g) = j, N_t(a) = n \vg \bar{G}_k) = O(1)$ for any $n$.
Fix $n$, and write
\begin{align*}
\sum_{t\in \cT_k} \Pr(\AU_t(g) = j, N_t(j) = n \vg \bar{G}_k) 
= \bE \left[   \sum_{t\in \cT_k} \bI(\AU_t(g)=j, N_t(j)=n) \vgg \bar{G}_k  \right] 
\end{align*}

Let $L_t = \bI(\AU_t(g)=j, N_t(j)=n)$ be the indicator for the event of interest. Our goal is to count the number of times $L_t$ occurs.
Let $Y_m = \{\exists \; t : \sum_{s = 1}^t L_s = m\}$ be the event that $L_s$ occurs at least $m$ times.
Note that for $Y_m$ to occur, it must be that $Y_{m-1}$ occurred.
Therefore, by explicitly writing out the expectation, we have
\begin{align*}
\bE \left[    \sum_{t=1}^T \bI(\AU_t(g)=j, N_t(j)=n) \vgg \bar{G}_k  \right] 
\leq& \sum_{m \geq 1} m \Pr(Y_m \vg \bar{G}_k) \\
=& \sum_{m \geq 1} m \Pr(Y_m \vg Y_{m-1}, \bar{G}_k) \Pr(Y_{m-1} \vg \bar{G}_k).
\end{align*}

We claim that there exists a $\lambda \in (0, 1)$ such that $\Pr(Y_m \vg Y_{m-1}, \bar{G}_k) \leq \lambda$.
Let $\tau$ be the time when $L_s$ occurred for the $m-1$'th time, which exists since $Y_{m-1}$ is true.
For $Y_m$ to occur, it must be that arm $j$ was not pulled at time $\tau$, even though arm $j$ is the UCB.
Given that $j$ is the UCB, there exists a group $g$ in which $N_{\tau}^g(a) \leq  \hq_t^g(a) N_{\tau}(a)$.
If such a group arrives, it will pull $j$ with probability at least $\frac{1}{K}$.
Therefore, at time $\tau$, the probability that arm $j$ will be pulled is at least $\min_{g \in G} \frac{p_g}{K}$.
Then, $\lambda = 1-\min_{g \in G} \frac{p_g}{K}$ satisfies $\Pr(Y_m \vg Y_{m-1}, \bar{G}_k) \leq \lambda$.

Therefore, 
\begin{align*}
\bE \left[    \sum_{t=1}^T \bI(\AU_t=j, N_t(j)=n) \vgg \bar{G}_k \right] 
&= \sum_{m \geq 1} m \Pr(Y_m \vg Y_{m-1}, \bar{G}_k) \Pr(Y_{m-1} \vg \bar{G}_k)  \\
&\leq \sum_{m \geq 1} m \lambda^m  \\
&= O(1).
\end{align*}

Substituting back into \eqref{eq:ucb_radius_log_stronger} gives
\begin{align*}
\sumt \Pr(\AU_t = j,\UCB_t(j) \geq \OPT(g), \htheta_t(j) \leq \theta(a) + \delta \vg \bar{G}_k)  
\leq \sum_{n=1}^{c' \log T} O(1) 
= O(\log T).
\end{align*}

\end{myproof}

\begin{myproof}[Proof of \cref{lemma:optarm_pulled_not_greedy}]
Let $a \notin \Asub$, let $G$ be the set of groups in which $a$ is an optimal arm.
We condition on whether $a$ is the UCB for some group in $G$.

First, suppose $a = \AU_t(g)$ for some group $g \in G$, implying $\theta(a) = \OPT(g)$.
We can assume $\htheta_t(a) > \OPT(g) - \delta$ from \cref{commonlemma:empirical_mean_close_when_pulled}.
Then, if $a$ is not the greedy arm for $g$, there exists a suboptimal arm $j \in \Asubg$ with higher mean but lower UCB than $a$.
This implies that the UCB radius of $j$ is smaller than the UCB radius of $a$, implying that $j$ was pulled more times: $N_t(j) \geq N_t(a)$.
We show that this event cannot happen often.
Let $E_t = \{\pulls(a), A^{\text{greedy}}_t(g) \neq a, a \in \AKL, a = \AU_t(g), \htheta_t(a) > \OPT(g) - \delta\}$.
For any $j \in \Asubg$,
\begin{align*}
&\sumt \bI(E_t, N_t(j) \geq N_t(a), \htheta_{t}(j) > \OPT(g) - \delta)  \\
\leq& \sumt \sum_{n=1}^t \sum_{n_j=n}^t \bI(E_t, \htheta_{n_j}(j) > \OPT(g) - \delta, N_t(j) = n_j, N_t(a)=n)  \\
\leq& \sum_{n_j=1}^T \bI(\htheta_{n_j}(j) > \OPT(g) - \delta)  \sum_{n=1}^{n_j} \sum_{t=n}^T \bI(E_t, N_t(a)=n)  \\
\leq& \sum_{n_j=1}^T \bI(\htheta_{n_j}(j) > \OPT(g) - \delta) n_j,
\end{align*}
where the last inequality uses $\sum_{t=n}^T \bI(E_t, N_t(a)=n) \leq 1$ (since pulling arm $a$ increasing $N_t(a)$ by 1).
Since $\Pr(\htheta_{n}(j) > \OPT(g) - \delta) \leq \exp(-c n)$ for some constant $c > 0$, $\sumt \Pr(E_t, N_t(j) \geq N_t(a), \htheta_{t}(j) > \OPT(g) - \delta) = O(1)$.
Taking a union bound over actions $j \in \Asubg$ gives us the desired result:
\begin{align*}
\sumt \Pr(\pulls(a), A^{\text{greedy}}_t(g) \neq a\; \forall g \in G, a \in \AKL, \exists g \in G : a = \AU_t(g)) = O(\log \log T).
\end{align*}

Now, suppose $a \notin \AU_t(g)$ for all $g \in G$.
This means that there is another group $h$ where $a = \AU_t(h)$, but $a$ is suboptimal for $h$.
We assume $\allucb_t$ holds.
Let $a_h$ be an optimal arm for $h$. 
Since $\allucb_t$, $\UCB_t(a_h) \geq \OPT(h)$.
Therefore, it must be that $\UCB_t(a) \geq \OPT(h)$.
By \cref{lemma:main_epoch},
\begin{align*}
\sumt \Pr(\pulls(a), \UCB_t(a) \geq \OPT(h)) = O(\log \log T).
\end{align*}
This finishes the proof.
\end{myproof}

\begin{myproof}[Proof of \cref{lemma:F_k}]

Fix $a \in \cA$ and time $t$. 
We will show $\Pr(\hmu_{s_k}(a) \in [\theta(a) - \delta/2, \theta(a) + \delta/2], N_{s_k}(a) \geq c_a \log s_k) \geq 1- O\left(\frac{1}{\log t}\right)$. Then the result follows from taking a union bound over actions.
We first show that $\PFUCB$ is log-consistent.
\begin{lemma} \label{prop:scaled_log_consistent}
$\NFUCB$ is log-consistent.
\end{lemma}

Let $g \in \gmin(a)$. 
Since $\Pr(M_t(a) < \frac{p_g}{2} t) \leq \exp(-\frac{1}{2}p_g t)$, we can assume that there have been at least $\frac{p_g}{2}t$ arrivals of $g$ by time $t$.
Then, using \cref{prop:lower_bound_stochastic} and \cref{prop:lower_bound_opt_stochastic}, we know that at time $t$, $\Pr(N_{t}(a) < c_a \log t | M_t(a) \geq \frac{p_g}{2} t ) \leq O \left( \frac{1}{\log t} \right)$.
Next, 
we show that the probability of the event $\htheta_{t}(a) \notin [\theta(a) - \delta/2, \theta(a) + \delta/2]$ given that we have more than $c_a \log t$ pulls of $a$ is small.
\begin{align*}
&\Pr(\htheta_{t}(a) \notin [\theta(a) - \delta/2, \theta(a) + \delta/2] \vg N_{t}(a) \geq c_a \log t) \\
=& \sum_{n=c_a \log t}^t \Pr(\htheta_{n}(a) \notin [\theta(a) - \delta/2, \theta(a) + \delta/2] \vg N_t(a) = n) \Pr(N_t(a) = n) \\
\leq& \sum_{n=c_a \log t}^t \exp(- c_1 n) \Pr(N_t(a) = n) \\
\leq& c_3 \exp(- c_2 \log t)  \\
\leq& \frac{c_3}{t^{c_2}},
\end{align*}
for some constants $c_1, c_2, c_3 > 0$ that depends on the instance, $a$, and $\delta$.
Combining, we have that for any action $a$, $\Pr(\hmu_{s_k}(a) \in [\theta(a) - \delta/2, \theta(a) + \delta/2], N_{s_k}(a) \geq c_a \log s_k) \geq 1- O\left(\frac{1}{\log t}\right)$.

\end{myproof}

\begin{myproof}[Proof of \cref{lemma:outside_boundary_start_in_boundary}]
Let $U_a = \theta(a) + \delta$ and $U^I_a = \theta(a) + \delta/2$.
Let $\eta = U_a - U^I_a$.
Since $F_k$ is true, $N_{s_k}(a) \geq c_a \log s_k$.
Let $n_1 = N_{s_k}(a)$.
Let $\theta^n(a)$ be the empirical average of arm $a$ after $n$ pulls.
We will bound
\begin{align*}
 \Pr(\cup_{n_2 = n_1+1}^\infty \{\hmu^{n_2}(a) \notin [L_a, U_a]\} \vg \hmu^{n_1}(a) \in [L^I_a, U^I_a]).
\end{align*}
For any $n_2$, $\hmu^{n_2}(a) > U_a$ implies $\hmu^{n_2}(a) > \hmu^{n_1}(a) + \eta$.
Fix $n_2 > n_1$. Let $m = n_2 - n_1$.
\begin{align*}
\left\{\hmu^{n_2}(a) > U_a\right\}
&= \left\{\sum_{i=1}^{n_2} X_i > n_2 U_a \right\} \\
&= \left\{n_1 \hmu^{n_1}(a) +  \sum_{i=n_1+1}^{n_2} X_i > n_2 U_a \right\} \\
&= \left\{\sum_{j=1}^{m} X_{n_1+j} > n_1 (U_a - \hmu^{n_1}(a)) + m U_a \right\} \\
&= \left\{\sum_{j=1}^{m} (X_{n_1+j} - \mu) > n_1 (U_a - \hmu^{n_1}(a)) +  m (U_a-\mu) \right\} 
\end{align*}

\textbf{Case $m \leq n_1$: }
Since $U_a - \mu > 0$ and $U_a - \hmu^{n_1}(a) > \eta$ if $F_k$ is true,
\begin{align*}
\Pr\left(\bigcup_{m=1}^{n_1} \{\hmu^{n_1+m}(a) > U_a\} \vgg F_k \right)
&\leq \Pr\left(\bigcup_{m=1}^{n_1} \left\{\sum_{j=1}^{m} (X_{n_1+j} - \mu) > n_1 \eta\right\} \vgg F_k \right) \\
&\leq \Pr\left(\max_{m=1, \dots, n_1} S_m > n_1 \eta \vgg F_k \right),
\end{align*}
where $S_m =\sum_{j=1}^{m} (X_{n_1+j} - \mu)$.
Given that $X_{n_1+j} - \mu$ are zero mean independent random variables,
by Kolomogorov's inequality, we have
\begin{align*}
\Pr\left(\bigcup_{m=1}^{n_1} \{\hmu^{n_1+m}(a) > U_a\} \vgg F_k \right)
\leq \frac{1}{n_1^2 \eta^2} \var(S_{n_1}) 
= \frac{\sigma^2}{n_1 \eta^2} 
= \frac{\sigma^2}{\eta^2} \cdot \frac{1}{ c_a \log s_k},
\end{align*}
where $\sigma_2 = \var(X_1)$.

\textbf{Case $m > n_1$: }
\begin{align*}
\Pr\left(\bigcup_{m=n_1}^{\infty} \{\hmu^{n_1+m}(a) > U_a\} \vgg F_k \right)
&\leq \Pr\left( \bigcup_{m=n_1}^{\infty} \left\{\frac{\sum_{j=1}^{m} (X_{n_1+j} - \mu)}{m} > U_a - \mu \right\} \vgg F_k \right) \\
&\leq \sum_{m=n_1}^{\infty} \Pr\left( \frac{\sum_{j=1}^{m} (X_{n_1+j} - \mu)}{m} > U_a - \mu  \vgg F_k \right) \\
&\leq \sum_{m=n_1}^{\infty}  \exp(-m D) \\
&= \frac{\exp(-n_1 D)}{1 - \exp(-D)} \\
&= \frac{1}{s_k^{c_a D}(1 - \exp(-D))},
\end{align*}
for a constant $D > 0$ that depends on $U_a - \mu$ and $\sigma^2$.

Therefore,
\begin{align*}
&\Pr\left(\bigcup_{m=1}^{\infty} \{\hmu^{N_{s_k}(a)+m}(a) > U_a\} \vgg F_k \right) \\
\leq&
\Pr\left(\bigcup_{m=1}^{n_1} \{\hmu^{N_{s_k}(a)+m}(a) > U_a\} \vgg F_k \right)
 + \Pr\left(\bigcup_{m=n_1}^{\infty} \{\hmu^{N_{s_k}(a)+m}(a) > U_a\} \vgg F_k \right) \\
\leq& \frac{\sigma^2}{\eta^2} \cdot \frac{1}{ c_a \log s_k} + \frac{1}{s_k^{c_a D}(1 - \exp(-D))} \\
=& O\left(\frac{1}{\log s_k}\right),
\end{align*}
as desired.
\end{myproof}

\begin{myproof}[Proof of \cref{prop:scaled_log_consistent}]
Fix a group $g$.
At time $t$, if group $g$ arrives, the $\NFUCB$ pulls either the UCB arm or the greedy arm.
The original regret analysis of KL-UCB from \cite{garivier2011kl} shows that
\begin{align*}
\sumt \Pr(A_t \notin \Aoptg, A_t = \AU_t, g_t = g) = O(\log T).
\end{align*}
\cref{prop:greedy_not_opt} shows that the number of times the greedy arm is pulled and incurs regret is $O(\log \log T)$.
Combining, the total regret is $O(\log T)$.
\end{myproof}

\subsection{Proof of \texorpdfstring{\cref{prop:eps_close_delta}}{\ref{prop:eps_close_delta}}}
\label{app:proof:eps_close_delta}

\begin{myproof} 

First, we prove the statement with respect to the variables $(s^g)_{g \in \cG}$.
Let $f_s(s) = \sum_{g \in\cG} \log s^g$,
and let
$s_*^g = \sum_{a \in \cA^g} \Delta^g(a) \left(\Ngroupa - q_*^g(a)\Nopta \right)$ and
$\hs_t^g = \sum_{a \in \cA^g} \hDelta^g(a) \left(\hNgroupa - \hq_t^g(a)\hNopta \right)$. 
Since $f_s$ is strictly concave with respect to $s$, $s_*^g$ is unique.
Define the event $H_t(\delta) = \{\htheta_t(a) \in [\theta(a)-\delta, \theta(a)+\delta]$ for all $a \in \cA\}$.

\begin{lemma} \label{lemma:convex_prob_wrt_s}
For any $\eps > 0$, there exists $\delta > 0$ such that if $H_t(\delta)$, then $\hs_t^g \in [s_*^g - \eps, s_*^g + \eps]$ for all $g \in \cG$.
\end{lemma}

This shows that if $H_t(\delta)$, then the variables $\hat{s}^g_t$ are close to $s_*^g$ for all $g$.
Next, we need to show that the corresponding $q$'s are also close.
Let $\proj(z, P)$ be the projection of point $z$ onto a polytope $P$.

Let $Q = \{q : \sum_{g \in G} q^g(a) = 1 \; \forall a \in \Asub, q^g(a) = 0 \; \forall g \in G, a \notin \Asub,  q^g(a) \geq 0 \; \forall g \in G, a \in \cA\}$ be the feasible space.
Let $S^g(q, \tilde{\theta}) = \sum_{a \in \cA^g} \tilde{\Delta}^g(a) \left(\tilde{J}^g(a) - q^g(a)\tilde{J}(a) \right)$, where $\tilde{\Delta}^g(a)$, $\tilde{J}^g(a)$, and $\tilde{J}(a)$ are computed with $\tilde{\theta}$.

Given $s = (s^g)_{g \in \cG}$, let $Q(s, \tilde{\theta}) = \{ q^g(a) \in Q : S^g(q, \tilde{\theta}) = s^g\}$ be the set of all feasible $q$'s that corresponds to the solution $s$ under the parameters $\tilde{\theta}$.
Note that $Q(s, \tilde{\theta})$ is a linear polytope, and we can write it as $Q(s, \tilde{\theta}) = \{q : A(\tilde{\theta}) q = b(s), q \geq 0\}$ for a matrix $A(\tilde{\theta})$ and a vector $b(s)$.
We are interested in the polytopes $Q(s, \theta)$ and $Q(\hs_t, \htheta_t)$, which correspond the optimal solutions of \eqref{eq:fair_opt_prob} and \eqref{eq:fair_opt_prob_karm} respectively.
The next two lemmas state that these polytopes are close together:
\begin{lemma} \label{lemma:convex_projection_close}
Let $\eps > 0$.
There exists $\delta > 0$ such that if $H_t(\delta)$,
for any $\hq \in Q(\hs_t, \htheta_t)$,
$||\proj(\hq, Q(s, \theta)) - \hq||_2 \leq \eps$.
\end{lemma}	

\begin{lemma} \label{lemma:convex_projection_close2}
Let $\eps > 0$.
There exists $\delta > 0$ such that if $H_t(\delta)$, for any $q \in Q(s, \theta)$, $||\proj(q, Q(\hs_t, \htheta_t)) - q||_2 \leq \eps$.
\end{lemma}

Let $q_* = \argmin_{q \in Q(s, \theta)} ||q||^2_2$, $\hq= \argmin_{q \in Q(\hs_t, \htheta_t)} ||q||^2_2$.
Our goal is to show $||q_* - \hq||_1 \leq \eps$.
Let $R(\eta) = \{q \in Q(s, \theta) : ||q||_2 \leq ||q_*||_2 + \eta\}$ for $\eta > 0$.
Since the function $||\cdot||^2_2$ is strongly convex and $q_*$ is minimizer, we have the following result:
\begin{claim} \label{claim:convex_norm}
For every $\eps > 0$, there exists $\eta > 0$ such that if $q \in R(\eta)$, then $|| q - q_* ||_2 \leq \eps$.
\end{claim}

First, assume $||\hq_t||_2 \leq ||q_*||_2$.
Let $\eta > 0$ be from \cref{claim:convex_norm} using $\eps = \frac{\eps}{2}$.
Let $\delta > 0$ be from \cref{lemma:convex_projection_close} using $\eps =\min\{ \frac{\eps}{2}, \eta\}$.
Let $q' = \proj(\hq, Q(s, \theta)) \in Q(s, \theta)$.
From \cref{lemma:convex_projection_close}, $||\hq_t - q'||_2 \leq \eta$, implying $||q'||_2 \leq ||\hq_t||_2 + \eta  \leq ||q_*||_2 + \eta$.
Therefore, $q' \in R(\eta)$. \cref{claim:convex_norm} implies $||q' - q_*|| \leq \frac{\eps}{2}$.
Let $\delta > 0$ correspond to $\frac{\eps}{2}$ from \cref{lemma:convex_projection_close}, so that $||\hq_t - q'||_2 \leq \frac{\eps}{2}$.
Then,
\begin{align*}
||\hq_t - q_*||_2 
\leq ||\hq_t - q'||_2  + ||q' - q_*||_2
\leq \eps.
\end{align*}
An analogous argument shows the same result in the case that $||q_*||_2 \leq ||\hq_t||_2$ using \cref{lemma:convex_projection_close2}.

\end{myproof}

\subsubsection{Proof of Lemmas}

We first state an additional lemma:

\begin{lemma} \label{lemma:convex_prob_q_change_objective}
For any $\eps > 0$ there exists a $\delta > 0$ such that if $H_t(\delta)$, then
for any feasible solution $q$, $|f(q) - \hat{f}(q)| < \eps$.
\end{lemma}

\begin{myproof}[Proof of \cref{lemma:convex_prob_q_change_objective}]
Let $q$ be a feasible solution.
Let $S^g(q, \tilde{\theta}) = \sum_{a \in \cA^g} \tilde{\Delta}^g(a) \left(\tilde{J}^g(a) - q^g(a)\tilde{J}(a) \right)$, where $\tilde{\Delta}^g(a)$, $\tilde{J}^g(a)$, and $\tilde{J}(a)$ are computed with $\tilde{\theta}$.

For each $g$, let $\eps_g > 0$ be such that if $|\tilde{s}^g - s_*^g| \leq \eps_g$, then $|\log s_*^g - \log \tilde{s}^g| \leq \frac{\eps}{G}$.
$\Delta^g(a), \Ngroupa$, and $\Nopta$ are all differentiable functions of $\theta$ with finite derivatives around $\theta_*$.
Then, it is possible to find $\delta_g > 0$ such that if $H_t(\delta_g)$, 
$|\hat{\Delta}^g(a) \left(\hat{J}^g(a) - q^g(a)\hat{J}(a) \right) - {\Delta}^g(a) \left({J}^g(a) - q^g(a){J}(a) \right)| \leq \frac{\eps_g}{|\cA|}$. Summing over actions, $|S^g(q, \hat{\theta}_t) - S^g(q, \hat{\theta})| \leq \eps_g$.
Then, if $H_t(\delta_g)$, $|\log S^g(q, \htheta) - \log S^g(q, \theta)| \leq \frac{\eps}{G}$.
Take $\delta = \min_{g \in \cG} \delta_g$.
If $H_t(\delta)$ is true, $|f(q) - \hat{f}(q)| < \eps$.
\end{myproof}

\begin{myproof}[Proof of \cref{lemma:convex_prob_wrt_s}]
Let $\eps > 0$.
Let $S_{\eps} = \{ s : |s^g - s_*^g| \leq \eps \; \forall g\}$ be the set around $s_*$ of interest.
Our goal is to show that $f_s(\hs) \in S_{\eps}$.
Let $\fbd = \max\{f(s) : s \in \mathrm{bd}(S_{\eps})\} < \ssf$ be the largest $f$ on the boundary of $S_{\eps}$.
Then, if $f_s(s) > \fbd$, it must be that $s \in S_{\eps}$. 
(Since the entire line between $s$ and $s_*$ must have a value of $f_s$ that is higher than $f_s(s)$ due to concavity, and it must cross the boundary.)
Therefore, we need to show $f_s(\hs_t) > \fbd$.
Let $\hq_t$ be the corresponding solution to $\hs_t$.
Then, $f_s(\hs_t) = \hf_t(\hq_t)$.
Let $\delta > 0$ as in \cref{lemma:convex_prob_q_change_objective} with $\eps = \ssf - \fbd$.
Then, if $H_t(\delta)$ is true,
\begin{align*}
f_s(\hs_t) 
= \hf_t(\hq_t)
\geq \hf_t(q_*) 
\geq f(q_*) -  (\ssf - \fbd) = \fbd,
\end{align*}
where the second inequality follows from \cref{lemma:convex_prob_q_change_objective}.

\end{myproof}

\begin{myproof}[Proof of \cref{lemma:convex_projection_close}]
Let $\eps > 0$.
Let $n$ be the dimension of $q$.
We will make use of the following closed form formula for the projection onto a linear subspace:
\begin{fact} \label{fact:proj}
Let $P = \{x : Ax = b\}$.
The orthogonal projection of $z$ onto $P$ is
$\proj(z, P) = z  - A^\top (A A^\top)^{-1} (Az - b)$.
\end{fact}

Let $Q = Q(s, \tilde{\theta})$, and let $A, b$ be the corresponding parameters of the linear constraints; i.e. $Q = \{ x : Ax = b, x \geq 0\}$.
Similarly, let $\hQ = Q(\hs_t, \htheta_t)$, and let $\hA, \hb$ be defined similarly.
Note that \cref{fact:proj} only works with equality constraints.

We define a distance between two linear polytopes.
We use the notation $P(D, f) = \{x : Dx = f\}$.
Then, $Q = P(A, b)$, $\hQ = P(\hA, \hb)$.
\begin{definition}
For two polytopes $P(A, b)$ and $P(A', b')$, the distance is defined as
$d(P(A, b), P(A', b')) = \max\{||A - A'||_2, ||b - b'||_2\}$.
\end{definition}

Note that for every $\alpha >0$, there exists $\delta > 0$ such that $H_t(\delta)$ implies $d(Q, \hQ) \leq \alpha$ using \cref{lemma:convex_prob_wrt_s}.
For any $\cI \in 2^{[n]}$, let $P_{\cI} = P(A_{\cI}$, $b_{\cI}) = \{x : Ax = b, x_i = 0 \; \forall i \in \cI\}$.

\begin{claim} \label{claim:proj_constant}
There exists a constant $C \geq 1$ such that
for any $\cI \in 2^{[n]}$ and any $\tA, \tb$ of same dimensions as $A_{\cI}, b_{\cI}$,
if $\tq \in P(\tA, \tb)$ with $\tq \leq 1$ (for all elements), then
$||\tq - \proj(\tq, P_{\cI}) ||_2 \leq C d(P_{\cI}, P(\tA, \tb))$.
\end{claim}

\begin{myproof}[Proof of \cref{claim:proj_constant}]
From \cref{fact:proj}, we have
$||\tq - \proj(\tq, P_{\cI}) ||_2  =  ||\AcI^\top (\AcI \AcI^\top)^{-1} (\AcI\tq - \bcI)||_2$.
Since $\tq \in P(\tA, \tb)$, $\tA\tq = \tb$.
Let $\lambda = \max_{\cI} ||\AcI^\top (\AcI \AcI^\top)^{-1}||_2$ and let $d = d(P_{\cI}, P(\tA, \tb))$.
Therefore,
\begin{align*}
||\tq - \proj(\tq, P_{\cI}) ||_2 
&\leq \lambda || (\AcI - \tA)\tq + (\tb - \bcI) ||_2 \\
&\leq \lambda \left(   || \AcI - \tA ||_2 ||\tq||_2 + || \tb - \bcI ||_2  \right)\\
&\leq 2 \lambda n d.
\end{align*}
Therefore, $C = 2 \lambda n$.
\end{myproof}

We now describe an iterative process to prove this result.

Let $Q^0 = \{q : Aq = b\}$ ($Q$ without the non-negativity constraint),
and same with $\hQ^0 = \{q : \hA q = \hb\}$.
Let $\alpha_0 = d(Q^0, \hQ^0)$.
Let $\tq^0 = \proj(\hq, Q^0)$.
By \cref{claim:proj_constant}, $||\hq - \tq^0 ||_2 \leq C \alpha_0$.
If $\tq^0 \geq 0$, then STOP here.

Otherwise, find an index $i$ which violates the non-negativity constraint using the following method:
\begin{itemize}
	\item Let $q \in Q$ be an arbitrary feasible point ($q \geq 0$).
	\item From the point $\tq^0$, move along the direction towards $q$. Let $p^0$ be the first point on this line where $p^0$ is non-negative.
	\item Since $Q$ is simply $Q^0$ with non-negativity constraints and both sets are convex, $p^0 \in Q$.
	\item Let $i$ be an index where $\tq^0_i < 0$ and $p^0_i = 0$ (the last index to become non-negative).
\end{itemize}
Since $\hq \geq 0$, it must be that $\hq_i \leq C \alpha_0$  since $||\tq^0 - \hq|| \leq C \alpha_0$.

Let $Q^1$ be the same polytope as $Q^0$, but with the additional constraint that $q_{i} = 0$ --- call this constraint $C$. Let $A^1, b^1$ be the corresponding equality constraints for $Q^1$.
Let $\hQ^1$ be the same polytope as $\hQ$, but with the additional equality constraint that $q_i = \hq_i$ --- call this constraint $\hC$. Let $\hA^1, \hb^1$ be the equality constraints for $\hQ^1$.
Note that the only difference between constraints $C$ and $\hC$ is the right hand side, which differ by at most $C \alpha_0$.
Therefore, $d(Q^1, \hQ^1) \leq d(Q^0, \hQ^0) + C \alpha_0 \leq 2C \alpha_0$.
Clearly, $\hq \in \hQ^1$.
Let $\tq^1 = \proj(\hq, Q^1)$.
Applying \cref{claim:proj_constant} again, we have
$||\hq - \tq^1 ||_2 \leq C (2C \alpha_0) = 2 C^2 \alpha_0$.
If $\tq^1 \geq 0$, then STOP here.

Otherwise, let $j$ be the index which violates the non-negativity constraint found using the same method as before; except this time, we draw a line between $\tq^1$ towards $p^0 \in Q$. We let $p^1$ be the first point where $p^1 \geq 0$.
Then, we repeat the above process. 
We define $Q^2$ to be the same polytope as $Q^1$, with the additional constraint that $q_j = 0$.
$\hQ^2$ is defined as $\hQ^1$ with the additional constraint $q_j = \hq_j$.
Then, $\hq_j \leq 2 C^2 \alpha_0$.
Therefore, $d(Q^2, \hQ^2) \leq d(Q^1, \hQ^1) + 2 C^2 \alpha_0 \leq 2C \alpha_0 +  2C^2 \alpha_0 \leq 4C^2 \alpha_0$.
Applying \cref{claim:proj_constant}, we get
$||\hq - \tq^2 ||_2 \leq C (4C^2 \alpha_0) = 4 C^3 \alpha_0$.
If $\tq^2 \geq 0$, then STOP here.

\textbf{After stopping: }
If this process stopped at iteration $m$, then $\tq^m  \in Q$ and
$||\hq - \tq^m ||_2 \leq 2^m C^{m-1} \alpha_0$.
It must be that $m \leq n$.
If $\alpha_0 = \frac{\eps}{2^n C^{n-1}}$, then $||\hq - \tq^m ||_2 \leq \eps$.
Then, $||\proj(\hq, Q) - \hq||_2 \leq \eps$.
Let $\delta > 0$ such that $H_t(\delta)$ implies $d(Q, \hQ) \leq \alpha_0$.
\end{myproof}

\begin{myproof}[Proof of \cref{lemma:convex_projection_close2}]
This proof follows essentially the same steps as the proof of \cref{lemma:convex_projection_close} by swapping $Q$ and $\hQ$.
The main difference is that we are projecting $q$ onto $Q(\hs_t, \htheta_t)$, but this must hold for all possible values of $\hs_t, \htheta_t$ (using a single $\delta$).
Due to this, the only thing we have to change from the proof of \cref{lemma:convex_projection_close} is \cref{claim:proj_constant}.
We must show that there exists a constant $C$ where \cref{claim:proj_constant} is satisfied for all possible values of $\hs_t, \htheta_t$.
The only place where $C$ relies on a property of the polytope $P_{\cI}$ is in choosing $\lambda$.
Therefore our goal is to uniformly upper bound 
$\max_{\cI} ||\hAcI^\top (\hAcI \hAcI^\top)^{-1}||_2$ for all possible $\hAcI$ that can be induced by all possible $\hs_t, \htheta_t$.

Note that since we assume that $H_t(\delta_0)$ holds,
the possible matrices $\hA$ lie in a compact space (since every element of the matrix $\hA$ can be at most $\delta_0$ apart).
Since $||A^\top (A A^\top)^{-1}||_2$ is a continuous function of the elements of the matrix $A$, $\lambda_1 = \max_{\hA} ||\hA^\top (\hA \hA^\top)^{-1}||_2$ exists.
Moreoever, for any $\cI$, $||\hAcI^\top (\hAcI \hAcI^\top)^{-1}||_2 \leq C(n) ||\hA^\top (\hA \hA^\top)^{-1}||_2$ for a constant $C(n)$.
Therefore, by replacing $\lambda$ with $\lambda_1 C(n)$, \cref{claim:proj_constant} holds.
\end{myproof}



\section{Proof of \texorpdfstring{\cref{thm:pfoam_nash_solution}}{Theorem \ref{thm:pfoam_nash_solution}}} \label{sec:app:pfoam_nash}

\cref{thm:pfoam_nash_solution} is a corollary of the following theorem, which bounds the number of times each group and context pulls each arm.
For any context $m \in \cM$, let $\Asub(m) = \{a  \in \cA(m) \; :\; \Delta(m, a) > 0\}$ be the set of arms that are suboptimal for that context.
Let $(Q^g(m, a))_{g, m, a}$ be the solution to \eqref{eq:fair_opt_prob_linear}, where $\Delta$ is the true parameter.
\begin{theorem} \label{prop:num_pull_ub_linear}
For $\eps >0 $ and any $g \in \cG, m \in \cM^g, a \in \Asub(m)$,
\begin{align*}
\limsupT
\frac{\bE[ N^g_T(m, a)]}{\log T} 
\leq Q^g(m, a) + \eps.
\end{align*}
\end{theorem}



We prove \cref{prop:num_pull_ub_linear} in \cref{sec:main_subresult_linear}, where several sub-results are proven in \cref{sec:linear_lemmas}.
Then, in \cref{sec:pf_main_result_linear}, we prove \cref{thm:pfoam_nash_solution} using \cref{prop:num_pull_ub_linear}.

\subsection{Proof of \cref{prop:num_pull_ub_linear}} \label{sec:main_subresult_linear}
We now prove \cref{prop:num_pull_ub_linear}. 
The proofs of lemmas can be found in the next section.
Fix $\eps >0 $ and $g \in \cG, m \in \cM^g, a \in \Asub(m)$.

We first introduce some notation.
Let $\pull_t^g(m, a) = \bI(g_t = g, m_t = m, A_t = a)$.
At time $t$, denote by $\fexplore_t, \texplore_t$ and $ \bexploit_t$ the indicator variables that represent whether the algorithm pulled an arm in the forced exploration, targeted exploration, and backup exploitation step respectively at time $t$.

First, we state a stability result on the solutions $\hQ_t(m, a)$ --- that is, if $\htheta_t \approx \theta$, then $\hQ_t \approx Q$.
This result follows from the assumption that the solution to the optimization problem \eqref{eq:fair_opt_prob_linear} is unique, continuous, and finite.
\begin{lemma} \label{prop:eps_close_delta_linear}
For any $\eps > 0$, there exists $\delta > 0$ such that if $H_t(\delta)$, then $\hQ_t^g(m, a) \in [Q^g(m, a) - \eps, Q^g(m, a) + \eps]$ for all $g \in \cG$, $m \in \cM$ and $a \in \Asub(m)$.
\end{lemma}

Next, we write Lemma A.2 from \cite{hao2020adaptive} which will be often used.
\begin{lemma} \label{lemma:HaoA.2}
Suppose for $t \geq d$, $G_t$ is invertible. For any $\delta \in (0, 1)$, we have
\begin{align*}
\Pr(\exists t \geq d, \exists a \in \cA, \text{ s.t. } |\langle a, \htheta_t-\theta \rangle| \geq ||a||_{G_t^{-1}} f_{n, \delta}^{1/2} ) \leq \delta,
\end{align*}
where $f_{n, \delta} = 2(1+1/\log n) \log (1/\delta) + cd \log (d \log n)$, where $c > 0$ is a universal constant.
We write $f_{n} = f_{n, 1/n}$ for short.
\end{lemma}

Define the event $\mathcal{B}_t$ as
\begin{align*}
\mathcal{B}_t = \{ \exists t \geq d, \exists a \in \cA \text{ s.t. } |\langle a , \htheta_t - \theta \rangle| \geq ||a||_{G_t^{-1}} f_T^{1/2}\}.
\end{align*}
\cref{lemma:HaoA.2} shows that $\Pr(\cB_t) \leq 1/T$.
Therefore, the total regret when $\cB_t$ is true is $O(1)$, and hence we can assume that $\mathcal{B}^c_t$ holds.

We state a lemma showing that the expected number of times the policy goes past step 1 is logarithmic.
\begin{lemma} \label{lemma:logarithmic_exploration}
$\bE[S(T)] = O(\log T)$.
\end{lemma}

Our goal is to bound $\bE[N_T^g(m, a)] =\bE[\sumt \pull^g_t(m, a)]$.
We can decompose $\pull^g_t(m, a)$ into which step of the policy occurred at time $t$ (either step 1, 3, 4, or 5).
In particular, we will show that the number of times that $\pull^g_t(m, a)$ during steps 1, 3, and 5 are negligible --- essentially all pulls are derived from step 4, targeted exploration.
The following three lemmas bound the regret from steps 1, 3, and 5 respectively.
The first lemma bounds regret from time steps in which the condition $\cD_t$ holds, which corresponds to step 1 of the policy (exploitation); this result follows directly from Lemma A.3 of \cite{hao2020adaptive}.
\begin{lemma} \label{lemma:step1}
$\bE\bigg[ \sumt \Delta(m_t, A_t) \bI(\cD_t, \mathcal{B}^c_t) \bigg]  = o(\log T)$
\end{lemma}

Next, we bound the regret from time steps corresponding to step 3 (forced exploration).
\begin{lemma}  \label{lemma:step3}
$\bE\bigg[ \sumt \Delta(m_t, A_t)  \bI(\fexplore_t) \bigg]  = o(\log T)$
\end{lemma}
\begin{myproof}[Proof of \cref{lemma:step3}]
Since we only force explore when an arm was pulled less than $\eps_T S(t)$ times, $\Nf(t) \leq |\cA| \eps_T S(t)$ surely.
Then, $\bE\bigg[ \sumt \Delta(m_t, A_t)  \bI(\fexplore_t) \bigg] \leq \Delta^{\max} \bE[\Nf(T)] \leq |\cA| \eps_T \bE[S(T)]$. 
The lemma follows from the fact that $\bE[S(T)] = O(\log T)$ from \cref{lemma:logarithmic_exploration}, and that $\eps_T \rightarrow 0$.
\end{myproof}

Lastly, we bound the regret from time steps corresponding to step 5 (backup exploitation).
\begin{lemma} \label{lemma:step5}
$\bE\bigg[ \sumt \Delta(m_t, A_t)  \bI(\bexploit_t, \mathcal{B}_t^c) \bigg] = o(\log T)$.
\end{lemma}
The proof of \cref{lemma:step5} can be found in the next section.

Since $\Delta^g(m, a) > 0$, showing that regret is sub-logarithmic implies that the number of times that $\bI(\pull_t^g(m, a))$ happens is sub-logarithmic.
That is, \cref{lemma:step1}, \cref{lemma:step3} and \cref{lemma:step5} imply that the number of times $\bI(\pull_t^g(m, a))$ happens in steps 1, 3, or 5 is sub-logarithmic; i.e.,
\begin{align}\label{eq:notstep4_sublog}
\bE\bigg[\sumt \bI(\pull_t^g(m, a), \texplore_t^c)\bigg] = o(\log T).
\end{align}

Lastly, we bound the pulls from targeted exploration.
Let $E_t = \{N_t(a) \geq \frac{\eps_T S(t)}{2} \; \forall a \in \cA\}$ be the event that all arms have been pulled at least $\eps_T S(t)/2$ times.
We state a lemma stating that we can assume that both $E_t$ and $H_t(\delta)$ hold for a small $\delta > 0$.
\begin{lemma}  \label{lemma:targeted_explore_decomp}
Let $(\delta_T)_{T=1}^\infty$ be a sequence with $\lim_{T \rightarrow \infty} \delta_T = 0$ and $\log \log T / \delta^2 = o(\log T)$.
\begin{align} \label{eq:texplore_canassumeclose}
 \bE\bigg[ \sumt \bI(\texplore_t) \bigg] 
 &\leq \bE\bigg[\sumt \bI(\texplore_t, H_t(\delta_T), E_t)\bigg] + o(\log T).
\end{align}
\end{lemma}



If $\{\pull_t^g(m, a), \texplore_t\}$ happens, then it must be that $N_t^g(m, a) \leq \hQ^g_t(m, a)f_T/2$.
Let $\delta_\eps$ be defined as the $\delta$ from \cref{prop:eps_close_delta_linear} using $\eps$ as input.
Suppose $T$ is large enough that $\delta_T \leq \delta_\eps$.
Then, if $H_t(\delta_T)$ is true, by \cref{prop:eps_close_delta_linear}, $N_t^g(m, a) \leq (Q^g(m, a) + \eps) f_T/2$.
Therefore, 
\begin{align} \label{eq:texplore_ub}
\sumt \bI(\texplore_t, H_t(\delta_T), E_t, \pull_t^g(m, a))
\leq (Q^g(m, a) + \eps) f_T/2.
\end{align}

Combining \eqref{eq:notstep4_sublog}, \eqref{eq:texplore_canassumeclose} and \eqref{eq:texplore_ub} gives
\begin{align*}
\limsupT \frac{\bE[N^g_T(m, a)]}{\log T} 
&\leq \limsupT \frac{(Q^g(m, a) + \eps) f_T/2 + o(\log T)}{\log T}  \\
&= Q^g(m, a) + \eps,
\end{align*}
where the last line follows from the fact that $\limT \frac{f_T}{2\log T} = 1$.



\subsection{Proofs of Lemmas} \label{sec:linear_lemmas}

Recall that $E_t = \{N_t(a) \geq \frac{\eps_T S(t)}{2} \; \forall a \in \cA\}$ is the event that all arms have been pulled at least $\eps_T S(t)/2$ times.
We first state a lemma stating that during time steps where $D_t^c$ occurs, we can assume that $E_t$ holds.
\begin{lemma} \label{lemma:fe_lb}
$\bE[\sumt \bI(E_t^c, D_t^c)] = O(1)$.
\end{lemma}

\begin{myproof}[Proof of \cref{lemma:fe_lb}]
$\sumt \Pr(E^c_t, \cD_t^c) \leq \sumt \Pr(E^c_t| \cD_t^c)$.
By the union bound, $ \Pr(E^c_t \; | \; \cD_t^c) \leq \sum_{a \in \cA} \Pr(N_t(a) < \frac{\eps_T S(t)}{2} \; | \; \cD_t^c)$.
Fix $a$, and fix $c$ to be a context that has access to arm $a$.
Let $t_S$ be the time step when $\cD_t^c$ occurs for the $S$'th time.
If $\cD_t^c$ happens less than $S$ times in total, then let $t_S = \infty$.
Define the sequence of random variables $X_1, X_2, \dots$ as follows.
For $S \leq S(T)$, let $X_S = \bI(c_{t_S} = c)$.
For $S > S(T)$, let $X_S =p_{\min}$.
Then, $\bE[X_{s+1} | X_1, \dots, X_s] \geq \pmin$.

Fix $S$. 
Consider the $S/2$ arrivals that consist of the latter half of arrivals when $\cD_t^c$ occurs.
\begin{claim} \label{claim:arrivals_S2}
If $\sum_{i=S/2}^S X_i \geq |\cA|\eps_T S$ occurs, then $N_{t_S}(a) \geq \frac{\eps_T S(t_S)}{2}$.
\end{claim}
\begin{myproof}[Proof of \cref{claim:arrivals_S2}]
Suppose $N_{t_{S/2}}(a) \leq \frac{\eps_T S(t_S)}{2}$.
Then, for every arrival of context $c$, there will be forced exploration. 
There can be at most $|\cA| \eps_T S$ forced exploration steps, until all arms satisfy the forced exploration threshold.
\end{myproof}

Therefore, $\Pr(N_{t_S}(a) \leq \frac{\eps_T S}{2}) \leq  \Pr(\sum_{i=S/2}^S X_i \leq |\cA|\eps_T S)$.
If $T$ is large enough that $\eps_T < \frac{p_{\min}}{2 |\cA|}$, Azuma's inequality gives
\begin{align*}
\Pr \left( \sum_{i=S/2}^S X_i \leq |\cA|\eps_T S \right) \leq \exp(-cS),
\end{align*}

for some constant $c > 0$.
Since $E^c_{t_S}$ occurs only when $\sum_{i=S/2}^S X_i < |\cA|\eps_T S$, we can write:
\begin{align*}
\sumt \Pr(E^c_t, D_t^c)
&= \bE\left[  \sumt \sum_{S=1}^T \bI(E^c_t, D_t^c, t = t_S)\right] \\
&\leq \bE\left[  \sumt \sum_{S=1}^T \bI\bigg(\sum_{i=S/2}^S X_i < |\cA|\eps_T S, D_t^c, t = t_S \bigg)\right] \\
&= \bE\left[  \sum_{S=1}^T  \bI\left(\sum_{i=S/2}^S X_i < |\cA|\eps_T S \right) \sumt \bI\bigg(D_t^c, t = t_S \bigg)\right].
\end{align*}
Since $\sumt \bI(D_t^c, t = t_S ) \leq 1$ for any $S$, we have
\begin{align*}
\sumt \Pr(E^c_t, D_t^c)
&\leq  \sum_{S=1}^T  \Pr\bigg(\sum_{i=S/2}^S X_i < |\cA|\eps_T S \bigg)  \\ 
&\leq \sum_{S=1}^T   \exp(-cS)  \\
&= O(1).
\end{align*}
\end{myproof}

\begin{myproof}[Proof of \cref{lemma:targeted_explore_decomp}]

Since $\texplore_t$ only occurs when $D_t^c$ is true, and \cref{lemma:fe_lb} gives $\bE[\sumt \bI(E_t^c, D_t^c)] = O(1)$,
\begin{align*}
  \bE\bigg[ \sumt \bI(\texplore_t) \bigg]  = \bE\bigg[ \sumt \bI(\texplore_t, E_t) \bigg] + O(1).
\end{align*}

Define
\begin{align*}
\tau_\delta = \min\left\{ t : \forall s \geq t \text{ where }E_s, \max_{a \in \cA} |\langle a, \htheta_s - \theta\rangle| \leq \delta \; \right\}.
\end{align*} 

Note that $\sum_{s=1}^t \bI(\texplore_s) \leq S(t)$. Then we can write
\begin{align} \label{eq:targeted_explore_decomp}
 \bE\bigg[ \sumt \bI(\texplore_t, E_t) \bigg] &\leq \bE[S(\tau_{\delta_T})] + \bE\bigg[\sumt \bI(\texplore_t, H_t(\delta_T), E_t)\bigg].
\end{align}
We need to show $\bE[S(\tau_{\delta_T})] = o (\log T)$.
Define 
\begin{align*}
\Lambda =  \min \left\{ \lambda : \forall t \geq d \text{ where } E_t, \; |\langle a, \hat{\theta}_t  \rangle - \langle a, \theta \rangle | \leq ||a||_{G_t^{-1}}f_{T, 1/\lambda} ^{1/2} \;\forall a \in \cA \right\}.
\end{align*}
If $E_t$ holds, we have $N_t(a) \geq \eps_T S(t)/2$, which implies $||a||_{G_t^{-1}} \leq \sqrt{\frac{2}{\eps_T S(t)}}$.
Then, for all $t$ where $E_t$ holds, 
\begin{align*}
\max_{a \in \cA} |\langle a, \hat{\theta}_t  - \theta \rangle | \leq \sqrt{\frac{2 f_{T, 1/\Lambda}}{\eps_T S(t)}}.
\end{align*}
For any $t'$ such that $\sqrt{\frac{2 f_{T, 1/\Lambda}}{\eps_T S(t')}} \leq \delta_T$, it must be that $\tau_{\delta_T} < t'$.
Therefore, 
\begin{align*}
S(\tau_{\delta_T}) \leq \frac{2 f_{T, 1/\Lambda}}{\eps_T \delta_T^2} + 1
= \frac{2(2(1+1/\log T) \log \Lambda  + cd \log (d \log T))}{\eps_T \delta_T^2}+1.
\end{align*}
By \cref{lemma:HaoA.2}, $\Pr(\Lambda \geq 1/\delta) \leq \delta$, which implies $\bE[\log \Lambda] \leq 1$.
By assumption, $\log \log /\Delta_T^2 = o(\log T)$.
Therefore, $\bE[S(\tau_{\delta_T})] = o(\log T)$ as desired.
\end{myproof}

\begin{myproof}[Proof of \cref{lemma:logarithmic_exploration}]

Define $\Nt(t) = \sum_{i=1}^t \bI(\texplore_i)$, $\Nf(t) = \sum_{i=1}^t \bI(\fexplore_i)$, $\Nb(t) = \sum_{i=1}^t \bI(\bexploit_i)$.
Therefore $S(t) = \Nt(t) + \Nf(t) + \Nb(t)$.
This lemma follows from combining the following three claims:
\begin{claim} \label{lemma:texploration_is_logarithmic}
$\bE[ \Nt(t)] = O(\log T)$. 
\end{claim}

\begin{claim} \label{lemma:s_prime_expectation}
$\bE[\Nb(t)] \leq \frac{1-\pmin}{\pmin} \bE[\Nt(t) + \Nf(t)]$.
\end{claim}

\begin{claim} \label{claim:fe_ub}
$\bE[\Nf(T)] \leq \bE[\Nt(T)]$.
\end{claim}


\begin{myproof}[Proof of \cref{lemma:texploration_is_logarithmic}]

From \cref{lemma:targeted_explore_decomp}, we have
\begin{align*}
 \bE\bigg[ \sumt \bI(\texplore_t, E_t) \bigg] 
 &\leq o(\log T) + \bE\bigg[\sumt \bI(\texplore_t, H_t(\delta_T), E_t)\bigg]
\end{align*}
Now we must bound $\bE\bigg[\sumt \bI(\texplore_t, H_t(\delta_T), E_t) \bigg]$.
First, write
\begin{align*}
\sum_{t} \bI(\texplore_t, H_t) = \sum_{a \in \cA} \sumt \bI(\texplore_t, A_t = a, H_t(\delta_T)).
\end{align*}

Let $T$ be large enough that $\delta_T$ is smaller than the $\delta$ needed by \cref{prop:eps_close_delta_linear} using $\eps = 1$.
Then, \cref{prop:eps_close_delta_linear} states that when $H_t(\delta_T)$ holds, $\hat{Q}^g_t(m, a) \leq Q^g(m, a) + 1$ for all $g \in \cG, m \in \cM, a \in \Asub(m)$.
Therefore, if an action $a$ is chosen for exploration at time $t$ where $H_t(\delta_T)$ holds, 
it must be that $N^g_t(m, a) \leq C  \log T$, for an instance-dependent constant $C > 0$. 
Therefore, given that action $a$ was chosen for exploration, $\sumt \bI(\texplore_t, A_t = a, H_t(\delta_T)) \leq C M \log T$, where $M$ is the total number of contexts.
Summing over actions gives $\sumt \bI(\texplore_t, H_t(\delta_T)) = O(\log T)$, and we are done.
\end{myproof}

The proof of \cref{lemma:s_prime_expectation} requires the following lemma, which says that $\cD_t$ holds when all arms meet the condition for targeted exploration.
\begin{lemma} \label{lemma:exploration_condition_is_sufficient}
Suppose for all $g \in \cG, m \in \cM^g, a \in \cA(m)$,
\begin{align*}
N_t^g(m, a) \geq \hat{Q}_t^g(m, a)f_T/2.
\end{align*}
Then, $\cD_t$ holds. That is, for all $m \in \cM, a \in \cA(m)$ such that $\hDelta_t(m, a) > 0$,
\begin{align*}
||a||^2_{G_t^{-1}} \leq  \frac{\hat{\Delta}_t(m, a)^2}{f_T}.
\end{align*}
Equivalently, if $\cD_t^c$ is true, then there exists a group $g \in \cG$, context $m \in \cM^g$, action $a \in \cA(m)$ where \eqref{eq:texplore_condition} is true.
\end{lemma}

\begin{myproof}[Proof of \cref{lemma:exploration_condition_is_sufficient}]
Let $H_{Q} = \mathsmaller{\sum_{a \in \cA}} Q(a) a a^{\top}$, for $Q(a) = \mathsmaller{\sum}_{g \in \cG} \mathsmaller{\sum}_{m \in \cM^g: a \in \cA(m)} \hQ_t^g(m, a) $.
Since $N_t^g(m, a) \geq \hat{Q}_t^g(m, a)f_T/2$ for all $g, m, a$, this implies that $G_t = (f_T / 2)H_Q  + E$, for a positive semidefinite matrix $E$.
Since $G_t - (f_T / 2)H_Q$ is positive semidefinite, $(2/f_T)H_Q^{-1} - G_t^{-1}$ is positive semidefinite.
Therefore, $||a||^2_{G_t^{-1}} \leq (2/f_T) ||a||^2_{H_Q^{-1}}$ for any $a \in \cA$.
Since $(\hQ_t^g(m, a))$ is feasible in $L(\hDelta)$, we have that for all $m \in \cM, a \in \cA(m)$ such that $\hDelta_t(m, a) > 0$,
\begin{align*}
||a||_{H_{Q}^{-1}}^2 \leq \hDelta_t(m, a)^2/2.
\end{align*}
Therefore, for all $m \in \cM, a \in \cA(m)$ such that $\hDelta_t(m, a) > 0$,
\begin{align*}
||a||_{G_{t}^{-1}}^2 \leq \hDelta_t(m, a)^2/f_T,
\end{align*}
as desired.

Suppose $N_t^g(m, a) \geq \hat{Q}_t^g(m, a)f_T/2$ for all $g, m, a$.
By definition of the optimization problem, for all $m, a$, $(\hat{Q}_t^g(m, a))$ satisfies
\begin{align*}
||a||_{H_{Q}^{-1}}^2 \leq \max\{\hDelta_t(m, a)^2, (\hDelta_t^{\min})^2\}/2,
\end{align*}
where $H_{Q} = \mathsmaller{\sum_{a \in \cA}} Q(a) a a^{\top}$, for $Q(a) = \mathsmaller{\sum}_{g \in \cG} \mathsmaller{\sum}_{m \in \cM^g: a \in \cA(m)} \hQ_t^g(m, a)$.
\end{myproof}

\begin{myproof}[Proof of \cref{lemma:s_prime_expectation}]
For any $t$, suppose $\cD_t^c$ is true.
Then, \cref{lemma:exploration_condition_is_sufficient} states that there exists a group $g \in \cG$, context $m \in \cM^g$, action $a \in \cA(m)$ where \eqref{eq:texplore_condition} is true.
Therefore, there is at least one group, context pair such that if that pair arrives, the policy would explore (either targeted or forced).
That is, $\Pr(\texplore_t, \fexplore_t | \cD_t^c) \geq \pmin$. We can write
\begin{align*}
\bE[\Nt(t) + \Nf(t)] 
&= \sum_{i=1}^t \Pr(\texplore_t, \fexplore_t) \\
&= \sum_{i=1}^t \Pr(\texplore_t, \fexplore_t \; | \; \cD_t^c ) \Pr(\cD_t^c) \\
&\geq \pmin \sum_{i=1}^t \Pr(\cD_t^c) \\
&\geq \pmin \bE[S(t)].
\end{align*}
The result follows from substituting $S(t) = \Nt(t) + \Nf(t) + \Nb(t)$ and rearranging.
\end{myproof}

\begin{myproof}[Proof of \cref{claim:fe_ub}]
Suppose to the contrary, that $\bE[\Nf(T)] >\bE[\Nt(T)]$.
Since we only force explore when an arm was pulled less than $\eps_T S(t)$ times, $\Nf(t) \leq |\cA| \eps_T S(t)$ surely.
Let $c = \frac{1-\pmin}{\pmin}$.
Let $T$ be large enough that if $\delta = |\cA| \eps_T$, $\frac{1-2\delta}{\delta} > 2c$.
Then, $\bE[\Nf(t)] \leq \delta (\bE[\Nf(t)] + \bE[\Nt(t)] + \bE[\Nb(t)]) \leq 2 \delta \bE[\Nf(t)] + \delta \bE[\Nb(t)]$, which implies 
\begin{align*}
\bE[\Nb(T)] \geq \frac{1-2\delta}{\delta} \bE[\Nf(T)].
\end{align*}
However, we also have $\bE[\Nb(T)] \leq c \bE[\Nf(T) + \Nt(T)] \leq 2c \bE[\Nf(T)]$ from \cref{lemma:s_prime_expectation}.
This is a contradiction as $\frac{1-2\delta}{\delta} > 2c$ by construction.
\end{myproof}

\begin{myproof}[Proof of \cref{lemma:step5}]
Note that $\bexploit_t$ can only occur when $\cD^c_t$ is true.
\cref{lemma:fe_lb} proves that $\sumt \Pr(E^c_t, \cD_t^c) = O(1)$, and therefore we can assume $E_t$ is true.
Let $\beta_T = 1/\log \log T$, and define
\begin{align*}
\zeta = \min\{ S : \forall t \text{ where } S(t) \geq S, E_t,  |\langle a, \hat{\theta}_t  \rangle - \langle a, \theta \rangle | \leq \beta_T \; \forall a \in \cA \}.
\end{align*}
We condition on whether $S(t)$ is smaller or larger than $\zeta$ and bound the regret separately.

\textbf{Case $S(t) \leq \zeta$.}
Note that $S(t)$ increments by 1 every time $\mathcal{D}_t^c$ occurs.
Therefore, the number of times that the event $\{S(t) \leq \zeta, \mathcal{B}_t^c, \mathcal{D}_t^c, E_t\}$ can occur is at most $\zeta$.
Hence,
\begin{align}
\bE\left[ \sum_{t=1}^T \bI(S(t) \leq \zeta, \mathcal{B}_t^c, \mathcal{D}_t^c, E_t) \right]	
\leq \bE[\zeta].
\end{align}
We need to show $\bE[\zeta] = o(\log T)$.
Define
\begin{align*}
\Lambda' =  \min \left\{ \lambda : \forall t \geq d \text{ where } E_t, \; |\langle a, \hat{\theta}_t  \rangle - \langle a, \theta \rangle | \leq ||a||_{G_t^{-1}}f_{n, 1/\lambda} ^{1/2} \;\forall a \in \cA \right\},
\end{align*}
where $f_{n, \delta} = 2(1 + 1/\log n)\log (1/\delta) + cd\log(d \log n))$.
If $E_t$ holds, we have
\begin{align*}
||a||^2_{G_t^{-1}} \leq \frac{1}{N_t(a)} \leq \frac{2}{\eps_T S(t)}.
\end{align*}

Therefore, if $S(t)$ is large enough that $\left(\frac{2}{\eps_T S(t)} f_{T, 1/\Lambda'}\right) ^{1/2} \leq \beta_T$, then $|\langle a, \hat{\theta}_t  \rangle - \langle a, \theta \rangle | \leq ||a||_{G_t^{-1}}f_{T, 1/\lambda} ^{1/2}$ implies $|\langle a, \hat{\theta}_t  \rangle - \langle a, \theta \rangle | \leq \beta_T$ (for $t$ where $E_t$ holds).
By definition, $\zeta$ must be no larger than any $S(t)$ where this holds.
This implies
\begin{align*}
\zeta \leq \frac{2 f_{T, 1/\Lambda'}}{\eps_T \beta_T^2 }+1.
\end{align*}
By \cref{lemma:HaoA.2}, $\Pr(\Lambda' \geq 1/\delta) \leq \delta$, which implies $\bE[\log \Lambda'] \leq 1$.
Combining, 
\begin{align*}
\bE[\zeta]	\leq \frac{2(1 + 1/ \log T) \bE[\log \Lambda'] + cd \log (\log(d \log T)) }{\eps_T \beta_T^2} = o(\log T).
\end{align*}

\textbf{Case $S(t) > \zeta$.}
Next, we can assume $S(t) > \zeta$.
By definition of $\zeta$, $|\langle \hat{\theta}_t, a  \rangle - \langle \theta, a \rangle | \leq \beta_T$ for all arms $a$.
Let $\hat{a}^*_t(m) = \argmax_{a \in \cA(m)} \langle \htheta_t, a \rangle$ be the estimated best arm for context $m$.
Let $\hDelta_t(m, a) = \langle \htheta_t, \hat{a}^*_t(m) - a \rangle$ be the estimated regret from arm $a$ under context $m$.
We claim that the estimated optimal arm is the true optimal arm ($a^*_t(m) = \hat{a}^*_t(m)$).
Indeed, for any $m \in \cM$,
\begin{align*}
\hDelta_t(m, a^*_t(m)) 
&= \langle \htheta_t, \hat{a}^*_t(m) - a^*_t(m) \rangle \\
&= \langle \htheta_t, \hat{a}^*_t(m) \rangle -  \langle \theta, \hat{a}^*_t(m) \rangle   - \langle \htheta_t, a^*_t(m) \rangle + \langle \theta, a^*_t(m)\rangle - \langle \theta, a^*_t(m)\rangle + \langle \theta, \hat{a}^*_t(m) \rangle \\
&= \langle \htheta_t -\theta, \hat{a}^*_t(m) \rangle 
+ \langle  \theta - \htheta_t, a^*_t(m) \rangle - \langle \theta, a^*_t(m) - \hat{a}^*_t(m)\rangle  \\
&\leq 2 \beta_T - \langle \theta, a^*_t(m) - \hat{a}^*_t(m)\rangle.
\end{align*}
Suppose $a^*_t(m) \neq \hat{a}^*_t(m)$, in which case $\langle \theta, a^*_t(m) - \hat{a}^*_t(m)\rangle \geq \Delta_{\min}$.
When $T$ is sufficiently large, $\beta_T < \Delta_{\min}/2$ --- this implies that $\hDelta_t(m, a^*_t(m)) < 0$, which is a contradiction. 
Then, it must be that $a^*_t(m) = \hat{a}^*_t(m)$, which means that the estimated optimal arm is indeed the optimal arm.
Therefore, when $S(t) > \zeta$, the regret from backup exploitation is 0.
\end{myproof}

\end{myproof}

\subsection{Proof of \cref{thm:pfoam_nash_solution}} \label{sec:pf_main_result_linear}
Finally, we prove \cref{thm:pfoam_nash_solution} using \cref{prop:num_pull_ub_linear}.
\begin{myproof}[Proof of \cref{thm:pfoam_nash_solution}]
From \cref{prop:num_pull_ub_linear}, we have that for all $g \in \cG$, $m \in \cM^g$, $a \in \Asub(m)$,
\begin{align*}
\limsupT
\frac{\bE[ N^g_T(m, a)]}{\log T} 
\leq Q^g(m, a).
\end{align*}
This implies an upper bound the group regret for any group $g$:
\begin{align} \label{eq:group_regret_ub_linear}
\limsupT
\frac{\bE[ \Reg^g_T(\pi^{\PFUCB}, \cI)]}{\log T} 
\leq \sum_{m \in \cM^g} \sum_{a \in \cA(m)} \Delta(m, a) Q^g(m, a).
\end{align}

We know from \cite{hao2020adaptive} that for any group $g$,
\begin{align} \label{eq:hao_regret_tight}
\limT  \frac{\tR^g_T(\cI)}{\log T} = \cC(\cM^g, \Delta).
\end{align}
We use \eqref{eq:group_regret_ub_linear} and \eqref{eq:hao_regret_tight} to bound the utility gain:
\begin{align*}
\ugain^g(\pi^{\PFUCB}, \cI) 
&= \liminfT  \frac{\tR^g_T(\cI) - \ER^g_T(\pi^{\PFUCB}, \cI)}{\log T}  \\
&= \cC(\cM^g, \Delta) - \limsupT \frac{\ER^g_T(\pi^{\PFUCB}, \cI)}{\log T}  \\
&\geq \cC(\cM^g, \Delta) - \sum_{m \in \cM^g} \sum_{a \in \cA(m)} \Delta(m, a) Q^g(m, a).
\end{align*}
We can use this to lower bound the Nash SW:
\begin{align*}
\welfare(\pi^{\PFUCB}, \cI) 
&= \sum_{g \in \cG} \log(\ugain^g(\pi^{\PFUCB}, \cI) ) \\
&\geq \sum_{g \in \cG} \log\bigg( \cC(\cM^g, \Delta) - \sum_{m \in \cM^g} \sum_{a \in \cA(m)} \Delta(m, a) Q^g(m, a) \bigg) \\
&= Z^*(\Delta).
\end{align*}
\end{myproof}

\section{Price of Fairness Proofs} \label{app:pof_proofs}
\subsection{Proof of \cref{thm:general_pof}} \label{app:proof:general_pof}

\begin{myproof}

Consider the optimization problem \eqref{eq:fair_opt_prob}.
For $g \in \cG$, define the variable $u^g = \max\{0, \sum_{a \in \Asub^g} \Delta^g(a)(J^g(a) - q^g(a) J(a))\}$.
Consider the polytope $P = \{(u^g)_{g \in \cG}: \text{$u$ is feasible in \eqref{eq:fair_opt_prob}}\}$ induced by the feasible region of the optimization problem --- refer to $P$ as the ``utility set'' in the language of \cite{bertsimas2011price}.
$P$ is compact and convex.

We would like to apply Theorem 2 of \cite{bertsimas2011price} to $P$, which would give us the desired result.
To do this, we need to show that the utilities induced by KL-UCB and PF-UCB respectively are the points in $P$ that maximize the sum and the sum of logs of the $u^g$ variables respectively. 
This is indeed the case for PF-UCB (since PF-UCB is the Nash solution).
For KL-UCB, we need to show that the induced utilities are \textit{non-negative} for each group.

Let $g$ be a group and let $a \in \Asub^g$ be a suboptimal arm for that group.
\cref{thm:main_theorem} shows that if $g \notin \gmin(a)$, then $\limsupT \frac{\bE[N_T^g(a)]}{\log T} = 0$.
Otherwise if $g \in \gmin(a)$, we have $J(a) = J^g(a)$.
\cref{theorem:number_pulls} shows that the total number of pulls of arm $a$ is less than $J(a)$.
Therefore,
\begin{align*}
\limsupT \frac{\bE[N_T^g(a)]}{\log T} \leq J^g(a).
\end{align*}
This implies 
\begin{align*}
\limsupT \frac{\ER_T^g(\pi^{\KLUCB}, \cI)}{\log T} \leq \sum_{a \in \Asub^g} \Delta^g(a) J^g(a) = \limsupT \frac{\tR_T^g(\cI)}{\log T}.
\end{align*}
Therefore, $\ugain^g(\pi^{\KLUCB}, \cI) \geq 0$ for all groups $g \in \cG$.

\end{myproof}

\subsection{Proof of \texorpdfstring{\cref{theorem:pof_specialized}}{Theorem \ref{theorem:pof_specialized}}} \label{app:proof:specialized_pof}


We first provide intuition on this result by proving this result for the 2-group 3-arm instance from \cref{ex:3arm}.
For this simple instance, the optimization problem $(P(\theta))$ is very simple --- there is essentially only one parameter $q^B(1) \in [0, 1]$, which represents the percentage of pulls of arm 1 assigned to group B.
Then the optimization problem reduces to:
\begin{align*}
q_*^B(1) &= \argmax_{q \in [0, 1]} \left( q (\OPT(A) - \theta_1) J^A(1) \right)  \left( (\OPT(B) - \theta_1) (J^B(1) -q J^A(1))) \right)  \\
&=\argmax_{q \in [0, 1]} q \left(\frac{J^B(1)}{J^A(1)} - q \right) \\
&= \frac{J^B(1)}{2 J^A(1)}
\end{align*}

Plugging in $q_*^B(1)= \frac{J^B(1)}{2 J^A(1)}$, the utility gain for group B is exactly $(\OPT(B) - \theta_1) J^B(1)/2$.
Note that under a regret-optimal solution, all pulls will be assigned to group A; hence the utility gain for group A will be 0, but group B will incur no regret, and hence the total utility gain will be $(\OPT(B) - \theta_1) J^B(1)$.
Therefore, the total utility gain under the Nash solution is at least half of the total regret under the regret-optimal solution.

\cref{theorem:pof_specialized} is a generalization of the above example.
We cannot derive the solution to $P(\theta)$ in closed form like we did above, 
but we use the intuition from the example to prove structural properties of the optimal solution, that allow us to derive the 1/2 bound.

\begin{myproof}[Proof of \cref{theorem:pof_specialized}]
In this proof, for convenience, we use subscripts instead of superscript to refer to groups $g$ since we do not need to refer to time steps. 

Let $\{1, \dots, M\}$ be the set of shared arms, where $\theta_1 \leq \dots \leq \theta_M$.
Let $\cG =[G]$ be the set of groups, where $\OPT(1) \leq \dots \leq \OPT(G)$.
We assume that $\theta_M < \OPT(1)$.
(If there is a shared arm whose reward is as large as $\OPT(1)$, then neither policy will incur any regret from this arm, and hence this arm is irrelevant.)
In this case, all of the regret in the regret-optimal solution goes to group 1, and the other groups incur no regret.
Therefore, the total utility gain of the regret-optimal solution is the sum of the regret at the disagreement point for groups 2 to G. Specifically, $\limT \mathrm{SYSTEM}_T(\cI) = \limT \sum_{g=2}^G \frac{\tR^g_T(\pi^{\KLUCB})}{\log T}$.

We will show that for each group $g \geq 2$, the regret incurred from $\PFUCB$ is less than half of the regret at the disagreement point --- i.e. $R^g_T(\pi^{\PFUCB}, \cI) \leq \frac{1}{2} \tR^g_T(\cI)$.
Then, the utility gain for the group reduces by at most a half from the regret-optimal solution, which is our desired result.

Let $R_g = \limT \frac{R^g_T(\pi^{\PFUCB}, \cI)}{\log T}$ and $\tR_g = \limT \frac{\tR^g_T(\cI)}{\log T}$ for all $g \in \cG$.
Recall that the proportionally fair solution comes out of the optimal solution to the following optimization problem:
\begin{equation*}
\begin{aligned} 
 \max_{q \geq 0} \quad & \sum_{g \in \cG} \log \bigg( \sum_{a \in \Asubg} \Delta^g(a) \left( \Ngroupa - q^g(a) \Nopta \right)\bigg)^+  \\
\text{s.t. } 
\quad
&\sum_{g \in \cG} q^g(a) = 1 \quad \forall a \in \Asub \\
& q^g(a) = 0 \quad \forall g \in G, a \notin \Asub \cap \cA_g.
\end{aligned} 
\tag{$P(\theta)$}
\end{equation*}

We first show a structural result of the optimal solution.
Let $s_g = \Delta^g(a) \left( \Ngroupa - q^g(a) \Nopta \right)$ be the utility gain for group $g$.
Note that in terms of minimizing total regret, it is optimal for group 1 to pull all suboptimal arms.
Therefore, if $q_g(a) > 0$ for some $g > 1$, we think of this as ``transferring'' pulls of arm $a$ from group 1 to group $g$.
This transfer increases the regret by a factor of $\frac{\Delta_g(a)}{\Delta_1(a)}$.
We prove the following property that these transfers must satisfy:
\begin{claim}[Structure of Optimal Solution] \label{claim:property_convex_opt}
For $g \in [M]$, let $b = \max\{a : q_g(a) > 0\}$.
If $h < g$, then $q_h(a) = 0$ for all $a < b$.
\end{claim}

Writing out the KKT conditions of the optimization problem gives us the following result.
\begin{claim}[KKT conditions] \label{claim:kkt_cond_utility}
Let $g, h \in \cG$, $a \in \cA$ such that $q_g(a) > 0$ and $h < g$.
Then,
$s_g \geq s_h \frac{\Delta_g(a)}{\Delta_h(a)}$.
Moreover, if $q_1(a) > 0$, $s_g \leq \frac{\Delta_2(a)}{\Delta_1(a)} s_1$ for any $g > 1$.
\end{claim}

The next claim is immediate from \cref{claim:kkt_cond_utility}.
\begin{claim} \label{claim:utility_gain_higher}
If $h < g$ and there exists an arm $a$ such that $q_g(a) > 0$, then $s_g \leq s_h$.
\end{claim}

Regret is minimized if $q_1(a) = 1$ for all $a$, in which case $s_1 = 0$.
If $s_1 \neq 0$, then we think of this as pulls from group 1 that are re-allocated to other groups $g \neq 1$.
This re-allocation increases total regret, since other groups incur more regret from pulling any arm compared to group 1.

Let $a_0 = \max \{ a : q_1(a) \neq 1 \}$. 
All pulls for any action $a > a_0$ come from group 1.
We claim that $q_2(a_0) > 0$.
Suppose not. 
Let $a' > 2$ such that $q_2(a_0) > 0$.
Then, by \cref{claim:property_convex_opt}, $q_2(a) = 0$ for all $a$.
This implies that $s_2 = r_2 > r_{a'} \geq s_{a'}$, which contradicts \cref{claim:utility_gain_higher}.
Then, by \cref{claim:kkt_cond_utility}, $s_2 = s_1 \frac{\Delta_2(a_0)}{\Delta_1(a_0)}$.

Next, we claim that $s_2 \geq \frac{\tR_2}{2}$, which proves the desired result for $g=2$.
Note that $s_1$ represents the amount of regret that was ``transferred'' from group 1 to other groups, which increases the total regret.
If \ii{all} of this was transferred to group 2, the total regret from group 2 would be at most $s_1 \frac{\Delta_2(a_2)}{\Delta_1(a_2)} \leq s_2$. 
Therefore, $R_2 \leq s_2$.
Since $R_2 + s_2 = \tR_2$, $s_2 \geq \frac{\tR_2}{2}$.

For $g > 2$, \cref{claim:kkt_cond_utility} shows $s_g \geq s_2$.
Moreover, since $\OPT(g) \geq \OPT(2)$, $\tR_g \leq \tR_2$.
Therefore, $s_g \geq s_2 \geq \frac{\tR_2}{2} \geq \frac{\tR_g}{2}$ as desired.

\end{myproof}

\subsection{Proof of Claims}

\begin{myproof}[Proof of \cref{claim:property_convex_opt}]
Suppose not.
Let $g \in \cG$ and $b = \max\{a : q_g(a) > 0\}$.
Let $a < b$ such that $q_h(a) > 0$.
Then, since $\sum_{g'} q_{g'}(a) = 1$, $q_g(a) < 1$.
By the ordering of arms and groups, we have 
\begin{align} \label{eq:ordering}
\frac{\Delta_h(a)}{\Delta_g(a)}  > \frac{\Delta_h(b)}{\Delta_g(b)}.
\end{align}
We essentially show, using this inequality, that if we want to ``transfer'' pulls from group $h$ to $g$, it is more efficient to do so using arm $a$ rather than arm $b$, and hence it is a contradiction that $q_h(b)$ is positive.

We construct a ``swap'' that will strictly increase the objective function.
Let $\eps = \min\{q_h(a), q_g(b), 1-q_g(a), 1-q_h(b)\}$.
\begin{itemize}
	\item Decrease $q_h(a)$ by $\eps$, and increase $q_h(b)$ by $\frac{\Delta_h(a) J(a)}{\Delta_h(b) J(b)} \eps \leq \eps$, where the last inequality follows from the convexity of $\KL(\theta_b, \cdot)$.
	By construction, $s_h$ does not change.
	\item Increase $q_g(a)$ by $\eps$, and decrease $q_g(b)$ by $\frac{\Delta_h(a) J(a)}{\Delta_h(b) J(b)} \eps$.
	The first operation decreases $s_g$ by $\Delta_g(a) J(a) \eps$, while the second operation increases $s_g$ by $\frac{\Delta_h(a) J(a) \Delta_g(b)}{\Delta_h(b)} \eps$. By \eqref{eq:ordering}, this strictly increases $s_g$ overall.
\end{itemize}
This is a contradiction.
\end{myproof}

\begin{myproof}[Proof of \cref{claim:kkt_cond_utility}]
From the stationarity KKT condition, we have that
\begin{align*}
\frac{\Delta_g(a) J(a)}{s_g} + \lambda(a) - \mu_g(a) &= 0, \\
\frac{\Delta_h(a) J(a)}{s_h} + \lambda(a) - \mu_h(a) &= 0,
\end{align*}
for some $\lambda_a \in \bR$ and $\mu_g(a), \mu_h(a) \geq 0$.
From complementary slackness, $\mu_g(a) q_g(a) = 0$. 
Since $q_g(a) > 0$, it must be that $\mu_g(a) = 0$.
Since $\mu_h(a) \geq 0$, $\frac{\Delta_g(a) J(a)}{s_g} \leq \frac{\Delta_h(a) J(a)}{s_h}$.
\end{myproof}

\section{Other Proofs} \label{app:other_propositions}

\subsection{Proof that the Nash Solution is Unique Under Grouped Bandit Model} \label{app:nash_sol_unique}
The uniqueness of the Nash bargaining solution in the general bargaining problem requires that the set $U$ is convex.
In the grouped bandit model, it is not clear that the set $U(\cI) = \{(\ugain^g(\pi, \cI))_{g \in \cG} :  \pi \in \conspolicies \}$ is convex.
In this section, we show that the uniqueness theorem still holds in the grouped bandit setting. 
The proof is essentially the same as the original proof of \cite{nash1950bargaining}; we simply show that the potential non-convexity due to the $\liminf$s creates inequalities in our favor.

Let $G$ be the number of groups.
Let $\welfare(u) = \sum_{g \in \cG} \log u_g$, and let
$f(U) = \argmax_{u \in U} \welfare(u)$ for $U \subseteq \bR^G$.
Fix a grouped bandit instance $\cI$, and let $\ssu = f(U(\cI))$.
We first show that $\ssu$ is unique (i.e. $\argmax_{u \in U(\cI)} \welfare(u)$ is unique).
Suppose there was another $u' \in U(\cI)$ with the same welfare.
Then, let $\bar{u} \in U(\cI)$ be the policy that runs $u'$ with probability 50\%, and $\ssu$ with probability 50\%.
Using the fact that $\liminfT (a_T + b_T) \geq \liminfT a_T + \liminf b_T$ implies that $\bar{u}_g \geq \frac{1}{2}(\ssu_g + u'_g)$ for all $g$.
Since $\log$ is strictly concave, $\log \bar{u}_g > \frac{1}{2}(\log \ssu_g + \log u'_g)$.
This implies $\welfare(\bar{u}) > \welfare(\ssu)$, which is a contradiction.

Next, we show that $f$ is the unique solution that satisfies the four axioms.
Let $U = U(\cI)$.
It is easy to see that this solution satisfies the axioms.
We need to show that no other solution satisfies them.
Suppose $g(\cdot)$ satisfies the axioms. 
We need to show $g(U) = f(U)$.
Let $U' = \{(\alpha_g u_g)_{g \in \cG} : u \in U; \alpha_g \ssu_g = 1, \alpha_g > 0\}$. $U'$ is the translated utility set so that $\ssu$ becomes the 1 vector. 
Then, the optimal welfare is $\welfare(\bI) = 0$.
We need to show $g(U') = \bI$.
We claim that there is no $v \in U'$ such that $\sum_{g \in \cG} v_g > G$.
Assume that such a $v$ exists.
For $\lambda \in (0, 1)$, let $t$ be the utilities from the policy that runs the policy induced by $v$ with probability $\lambda$, and the policy induced by $\bI$ with probability $1-\lambda$.
Then, by the same argument with $\liminf$ to prove uniqueness, $t_g \geq \lambda v_g + (1-\lambda) 1$.
If $\lambda$ is small enough, then $\sum_{g \in \cG} \log t_g > 0$.
This is a contradiction to $\bI$ maximizing $\welfare(\cdot)$.

Consider the symmetric set $U'' = \{ u \in \bR^G: u \geq 0, \sum_{g} u_g \leq G\}$.
We have shown that $U' \subseteq U''$.
By Pareto efficiency and symmetry, it must be that $g(U'') = \bI$.
By independence of irrelevant alternatives, $g(U') = \bI$, and we are done.

\subsection{Proof of \cref{prop:klucb_disagreement_pt}} \label{sec:pf_klucb_disagreement}

\begin{myproof}[Proof of \cref{prop:klucb_disagreement_pt}]
Let $\cI$ be a grouped $K$-armed bandit instance, and let $g \in \cG$.
Let $\cI_g$ be the single-group instance derived from $\cI$.
First, we show a lower bound: we show that for any consistent  policy $\pi \in \conspolicies$,
\begin{align} \label{eq:dis_lb}
\liminfT
\frac{\ER_T^g(\pi, \cI_g)}{\log T} 
\geq \sum_{a \in \cA^g} \Delta^g(a) \Ngroupa.
\end{align}	
Applying \cref{prop:lb_group_num_pulls} to $\cI_g$ on any consistent policy $\pi$ gives 
\begin{align*}
\liminfT \frac{\bE[N_T(a)]}{\log T} \geq \Ngroupa.
\end{align*}
\eqref{eq:dis_lb} follows from summing over actions.

Now, we show that $\KLUCB$ matches this lower bound.
Let $T(g)$ be the number of arrivals of group $g$ by time $T$.
Let $t(n)$ be the time step in which the $n$'th arrival of $g$ occurs.
Let $\IReg_t(\pi, \cI) = \mu(\ssA_t) - \mu(A_t)$ be the instantaneous regret at time $t$ under a policy $\pi$.
Note that $\IReg_t(\pi, \cI_g) = 0$ for any time $t$ where $g_t \neq g$.
Therefore,
\begin{align*}
\ER_T(\pi^{\KLUCB}, \cI_g)
&= \bE\bigg[\sumt \IReg_t(\pi^{\KLUCB}, \cI_g)\bigg] \\
&= \bE\bigg[\sum_{n=1}^{T(g)} \IReg_{t(n)}(\pi^{\KLUCB}, \cI_g)\bigg]
\end{align*}

By restricting to only time steps of $g_t = g$, $\cI_g$ can be interpreted as the non-grouped $K$-armed bandit instance with arms $\cA^g$.
Then, the results of \cite{garivier2011kl} show that
\begin{align*}
\lim_{N \rightarrow \infty} \frac{\bE\big[\sum_{n=1}^{N} \IReg_{t(n)}(\pi^{\KLUCB}, \cI_g)\big]}{\log N} = \sum_{a \in \cA^g} \Delta^g(a) \Ngroupa.
\end{align*}
Denote by $L = \sum_{a \in \cA^g} \Delta^g(a) \Ngroupa$ the RHS.
Our goal is to show
\begin{align*}
\limsupT \frac{\bE\bigg[\sum_{n=1}^{T(g)} \IReg_{t(n)}(\pi^{\KLUCB}, \cI_g)\bigg]}{\log T}  \leq L.
\end{align*}
Let $\eps > 0$.
Let $N'$ be large enough that if $N > N'$,
\begin{align} \label{eq:limit_N_eps}
\frac{\bE\big[\sum_{n=1}^{N} \IReg_{t(n)}(\pi^{\KLUCB}, \cI_g)\big]}{\log N} \leq L + \eps.
\end{align}

Let $T'$ be large enough that if $T > T'$, $\Pr(T(g) < \frac{p_g T}{2}) \leq \frac{1}{T}$ and $\frac{p_g T}{2} > N'$.
Fix $T > T'$.
Then,
\begin{align*}
\bE\bigg[\sum_{n=1}^{T(g)} \IReg_{t(n)}(\pi^{\KLUCB}, \cI_g)\bigg]
\leq \bE\bigg[\sum_{n=1}^{T(g)} \IReg_{t(n)}(\pi^{\KLUCB}, \cI_g) \; \big| \; T(g) \geq \frac{p_g T}{2} \bigg] +  T \Pr\left(T(g) < \frac{p_g T}{2}\right)
\end{align*}

Then we have
\begin{align*}
\frac{\bE\bigg[\sum_{n=1}^{T(g)} \IReg_{t(n)}(\pi^{\KLUCB}, \cI_g)\bigg]}{\log T} 
&\leq \bE\bigg[\frac{\sum_{n=1}^{T(g)} \IReg_{t(n)}(\pi^{\KLUCB}, \cI_g) }{\log T} \bigg| \; T(g) > \frac{p_g T}{2}\bigg]\\
&\leq \bE\left[\frac{\sum_{n=1}^{T(g)} \IReg_{t(n)}(\pi^{\KLUCB}, \cI_g)}{\log T(g)} \; \bigg| \; T(g) > \frac{p_g T}{2} \right] \\
&\leq L+\eps,
\end{align*}
where the last step follows from \eqref{eq:limit_N_eps} and the fact that for any $n$, $\IReg_{t(n)}(\pi^{\KLUCB}, \cI_g)$ is independent of $T(g)$.

\end{myproof}

\subsection{Omitted Details of \cref{prop:nash_sw_ub}} \label{sec:app_f_pi_upper_bound_proof}

We provide details on the two steps in \cref{sec:ub_nash_sw} starting from \eqref{eq:regret_param_N_t_log}.
\eqref{eq:lower_bound_pulls} implies that for every $\eps > 0$, there exists a $T_\eps$ such that if $T \geq T_\eps$, then
\begin{align*}
\frac{\bE[N_T(a)]}{\log T} \geq (1-\eps)\Nopta.
\end{align*}
Therefore, for large enough $T$, plugging into \eqref{eq:regret_param_N_t_log}, we get
\begin{align*}
  \frac{\ER^g_T(\pi, \cI)}{\log T} \geq  \sum_{a \in \Asub} \Delta^g(a) q_T^g(a, \pi) \Nopta (1-\eps).
\end{align*}
This implies that
\begin{align*}
 \limsupT  \frac{\ER^g_T(\pi, \cI)}{\log T} \geq  \limsupT  (1-\eps)\sum_{a \in \Asub} \Delta^g(a) q_T^g(a, \pi) \Nopta.
\end{align*}
Since this holds for every $\eps > 0$ and the RHS is continuous in $\eps$,
\begin{align} \label{eq:limsup_regret}
 \limsupT  \frac{\ER^g_T(\pi, \cI)}{\log T} \geq  \limsupT  \sum_{a \in \Asub} \Delta^g(a) q_T^g(a, \pi) \Nopta.
\end{align}
Plugging in \eqref{eq:limsup_regret} into the definition of $\ugain^{g}(\pi, \cI)$ gives
\[
\ugain^{g}(\pi, \cI) 
\leq
\liminfT 
\sum_{a \in \Asubg} \Delta^g(a) \left( \Ngroupa - q_T^g(a, \pi) \Nopta \mathbf{1}\{a \in \Asub\} \right).
\]
Using the definition of $\welfare(\pi, \cI)$ and taking the $\liminf$ outside of the sum gives
\[
\welfare(\pi, \cI) \leq \liminfT  \sum_{g \in \cG} \log \bigg( \sum_{a \in \Asubg} \Delta^g(a) \left( \Ngroupa - q_T^g(a, \pi) \Nopta \mathbf{1}\{a \in \Asub\}  \right)\bigg)^+. 
\]

\subsection{Proof of \cref{thm:sw_ub_linear}} \label{sec:pf:sw_ub_linear}

%
%
Fix an instance $\cI$ such that $G(\cI) = \cG$ and a consistent policy $\pi$.
Let $N_T^g(m,a)$ be the number of pulls of arm $a$ from a context $m$ from group $g$.
\begin{align*}
\welfare(\pi, \cI) 
&= \sum_{g \in \cG} \log\left(\liminfT  \frac{\tR^g_T(\cI) - \ER^g_T(\pi, \cI)}{\log T} \right) \\
&\leq  \liminfT \sum_{g \in \cG} \log\left(\cC(\cM^g, \Delta) - \frac{\ER^g_T(\pi, \cI)}{\log T} \right) \\
&\leq  \liminfT \sum_{g \in \cG} \log\left(\cC(\cM^g, \Delta) - \frac{ \sum_{m \in \mathcal{M}^g} \sum_{a \in \cA_{\text{sub}}^m} \Delta(m, a)\bE[ N_T^g(m, a)] }{\log T} \right) \\
\end{align*}
Then, $\alpha_T^g(m, a) = \frac{\bE[ N_T^g(m, a)]}{\log T}$ is asymptotically feasible for $\fairopt(\Delta)$.
Indeed, since $\pi$ is consistent, Lemma 3.2 of \cite{hao2020adaptive} shows that for any arm $a \in \cA$,
\begin{align*}
\frac{\Delta(m, a)^2}{2} 
\geq \limsup_{T \rightarrow \infty} \log T ||a||^2_{\bar{G}_T^{-1}}
= \limsup_{T \rightarrow \infty} ||a||^2_{H^{-1}(\alpha_T)}.
\end{align*}
Therefore for any $\eps > 0$, if $T$ is large enough, $||a||^2_{H^{-1}(\alpha_T)} \leq \frac{\Delta(m, a)^2}{2}  + \eps$, and so
\begin{align*}
\sum_{g \in \cG} \log\left(\cC(\cM^g, \Delta) - \frac{ \sum_{m \in \mathcal{M}^g} \sum_{a \in \cA_{\text{sub}}^m} \Delta(m, a)\bE[ N_T^g(m, a)] }{\log T} \right) \leq Z_{\eps}(\Delta), 
\end{align*}
where $Z_\eps(\Delta)$ is the objective value to the optimization problem $\fairopt(\Delta)$ with $\frac{(\Delta^m(a))^2}{2}$  replaced by $\frac{(\Delta^m(a))^2}{2} + \eps$ in the RHS of the constraints.
Hence, $\welfare(\pi, \cI) \leq Z_\eps(\cI)$.
Since we assumed that the objective value $Z$ is continuous in $\Delta$, $\inf_{\eps > 0} Z_\eps(\Delta) = Z(\Delta)$.
Therefore, since $\eps$ is arbitrary, we get the desired result.

\end{APPENDICES}

%
%




\end{document}